\documentclass{article}

\usepackage[final]{neurips_2023}
\setcitestyle{numbers,square,citesep={,}}

\usepackage[utf8]{inputenc} 
\usepackage[T1]{fontenc}    
\usepackage{booktabs}       
\usepackage{amsfonts}       
\usepackage{bm}
\usepackage{nicefrac}       
\usepackage{microtype}      
\PassOptionsToPackage{dvipsnames,table}{xcolor}
\usepackage{graphicx}
\usepackage{subfigure}
\usepackage{makecell}
\usepackage{hyperref}       
\usepackage{wrapfig}
\usepackage{float}
\usepackage{tcolorbox}
\usepackage{adjustbox}
\usepackage{caption}
\usepackage{multirow}
\usepackage{enumitem}
\usepackage{amsmath}
\usepackage{xspace}
\usepackage[normalem]{ulem}
\useunder{\uline}{\ul}{}
\usepackage[misc]{ifsym}

\hypersetup{
  colorlinks   = true,    
  urlcolor     = blue,    
  linkcolor    = blue,    
  citecolor    = blue      
}
\newcommand{\huien}[1]{\textcolor{black}{#1}}

\newcommand{\Name}{\texttt{RoboSMPLX}\xspace}

\title{Towards Robust and Expressive Whole-body Human Pose and Shape Estimation}

\author{
    Hui En Pang\textsuperscript{1}, \quad
    Zhongang Cai\textsuperscript{1,2}, \quad
    Lei Yang\textsuperscript{2}, \quad
    Qingyi Tao\textsuperscript{2}, \quad
    Zhonghua Wu\textsuperscript{2}, \quad
    \\
    \textbf{Tianwei Zhang\textsuperscript{1 \Letter}}, \quad
    \textbf{Ziwei Liu\textsuperscript{1}} \\
    \textsuperscript{1}S-Lab, Nanyang Technological University \quad
    \textsuperscript{2}SenseTime Research \quad \\
    \texttt{\{huien001, tianwei.zhang, ziwei.liu\}@ntu.edu.sg} \quad \\ \texttt{\{caizhongang, yanglei, taoqingyi, wuzhonghua\}@sensetime.com}
}

\begin{document}

\maketitle

\begin{abstract}

Whole-body pose and shape estimation aims to jointly predict different behaviors (e.g., pose, hand gesture, facial expression) of the entire human body from a monocular image. Existing methods often exhibit degraded performance under the complexity of in-the-wild scenarios. We argue that the accuracy and reliability of these models are significantly affected by the quality of the predicted \textit{bounding box}, e.g., the scale and alignment of body parts. The natural discrepancy between the ideal bounding box annotations and model detection results is particularly detrimental to the performance of whole-body pose and shape estimation.
In this paper, we propose a novel framework \Name to enhance the robustness of whole-body pose and shape estimation. \Name incorporates three new modules to address the above challenges from three perspectives: \textbf{1) Localization Module} enhances the model's awareness of the subject's location and semantics within the image space. \textbf{2) Contrastive Feature Extraction Module} encourages the model to be invariant to robust augmentations by incorporating contrastive loss with dedicated positive samples. \textbf{3) Pixel Alignment Module} ensures the reprojected mesh from the predicted camera and body model parameters are accurate and pixel-aligned. We perform comprehensive experiments to demonstrate the effectiveness of \Name on body, hands, face and whole-body benchmarks. Codebase is available at \url{https://github.com/robosmplx/robosmplx}.

    
\end{abstract}

\section{Introduction}
\label{sec:intro}

Human pose and shape estimation tries to build human body models from monocular RGB images or videos. It has gained widespread attention owing to its extensive applications in various fields, including robotics, computer graphics, and augmented/virtual reality. Early works use various statistical models (e.g., SMPL \cite{SMPL2015}, MANO \cite{MANO2017}, FLAME \cite{FLAME2017}) to individually reconstruct different parts, including human body \cite{Kanazawa2017, Kolotouros2019, Choi2020, Joo2021, Kolotouros2021, Dwivedi2022, Kocabas2022, Kocabas2021}, face \cite{Deng2019, Danecek2022, Feng2021}, and hand \cite{Lin2021, Boukhayma2019, Zimmermann2021}.
Recently, there is a growing interest in whole-body estimation \cite{Feng2019, Choutas2020, Zhou2021, Rong2021, Zhang2022}, which jointly estimates the pose, hand gestures and facial expressions of the entire human body from the input. Commonly these methods first employ separate sub-networks to extract the features of body, hands and face. These features are then used to predict whole-body 3D joint rotations and other parameters (e.g., body shape, facial expression), which are further combined to generate the whole-body 3D mesh. This is a crucial step towards modeling human behaviors in an efficient and practical manner. 

However, achieving accurate and robust whole-body estimation is particularly challenging as it requires precise estimation of each body part and the correct connectivity between them. 
In particular, due to the smaller sizes of hand and face images, they are typically localized, cropped and resized to higher resolutions before being processed by the relevant sub-network. To tackle the absence of ground-truth bounding boxes in the real-world scenarios, existing whole-body methods utilize various detection techniques to obtain the crops. The accuracy of the whole-body estimation is highly sensitive to the quality of input crops. Our experiment results in Section \ref{sec:motivation} show that even minor fluctuations in the scale and alignment of input crops can significantly affect the model performance, indicating a limited ability to localize and extract meaningful features about the subject in the image.

The lack of robustness in existing whole-body pose and shape estimation methods highlights three critical aspects that can be improved upon: 1) accurate localization of the subject and its parts, 2) accurate extraction of useful features, and 3) accurate pixel alignment of outputs. Inspired by these findings, we propose three novel modules, each specifically designed to address a particular goal: 

\begin{itemize}[leftmargin=*,topsep=0pt]
    \item \textbf{Localization Module}. 
    This module implements sparse and dense prediction branches to ensure the model is aware of the location and semantics of the subject's parts in the image. The learned location of the joint positions are helpful in recovering the relative rotations.  

    \item \textbf{Contrastive Feature Extraction Module}.
    This module incorporates a pose- and shape-aware contrastive loss, along with positive samples, to promote better feature extraction under robust augmentations. By minimizing the contrastive loss, the model can produce consistent representations for the same subject, even when presented with different augmentations, making it robust to various transformations and capable of extracting meaningful invariant features.

    \item \textbf{Pixel Alignment Module}.
    This module applies differentiable rendering to ensure \huien{a more} precise pixel alignment of the projected mesh, and learn more accurate pose, shape and camera parameters.

\end{itemize}

By integrating these three modules, we build a more robust and reliable whole-body pose and shape estimation framework, \Name. Comprehensive evaluations demonstrate its effectiveness on body, face, hands and whole-body benchmarks. 

\section{Related Works}


\textbf{Whole-body Mesh Recovery.} Despite significant progress in 3D body-specific \cite{Kolotouros2019b, Kolotouros2019, Choi2020, Joo2021, Kolotouros2021, Dwivedi2022, Kocabas2022, Kocabas2021}, hand-specific \cite{Lin2021, Boukhayma2019}, and face-specific \cite{Deng2019} mesh recovery methods, there have been limited attempts to simultaneously recover all those parts. Early studies on whole-body pose and shape estimation primarily fit a 3D human model to 2D or 3D evidence \cite{Joo2018,Xiang2019,Pavlakos2019,Xu2020}, which can be slow and susceptible to noise. Recent studies utilized neural networks to regress the SMPL-X parameters for a whole-body 3D human mesh. The model is composed of separate sub-networks to process body, hand and face, respectively. \textit{One-stage} methods, e.g., OS-X \cite{Lin2023}, have the benefit of reduced computational costs and improved communication within part modules for more natural mesh articulation. However, the omission of hand and face experts makes it difficult for the model to leverage the widely available part-specific datasets, thus decreasing the hand and face performance. \textit{Multi-stage} methods, e.g., ExPose \cite{Choutas2020}, FrankMocap \cite{Rong2021}, PIXIE \cite{Feng2019} and Hand4Whole \cite{Moon2022}, use different techniques to localize part crops. 

Expose \cite{Pavlakos2019} and PIXIE \cite{Feng2019} localize hand and part crops from the body mesh, making them dependent on the accuracy of body poses. Minor rotation errors accumulated along the kinematic chain may result in deviations in joint locations and thus inaccurate part crops. In contrast, Hand4Whole \cite{Moon2022} predicts hand and face bounding boxes using a network leveraging image features and 3D joint heatmaps, but the resulting crops have low resolution. PyMAF-X \cite{Feng2019} relies on an off-the-shelf whole-body pose estimation model to obtain crops, which, while more accurate, incurs extra computation. More detailed comparison with PyMAF-X are in Appendix \ref{sec:pymafx_comparison}.


\textbf{Robustness in vision tasks.}
Efforts to tackle robustness in vision tasks have utilized diverse strategies such as data augmentation, architectural innovations, and training methodologies \cite{kong2023robo3d,yang2020learn,rempe2021humor,liu2023poseexaminer,Wang2021,Zhang2022b,Bai2023}. AdvMix \cite{Wang2021} employs adversarial augmentation and knowledge distillation, challenging models with corrupted images to foster learning from complex samples. Architectural modifications, such as novel heatmap regression \cite{Zhang2022b}, have been introduced to mitigate the impact of minor perturbations. HuMoR \cite{rempe2021humor} utilizes a conditional variational autoencoder to capture the dynamics of human movement, thereby achieving generalization across diverse motions and body shapes. Additionally, PoseExaminer \cite{liu2023poseexaminer} employs a multi-agent reinforcement learning system to uncover failure modes inherent in human pose estimation models, highlighting model limitations in real-world scenarios. Complementing these efforts, Robo3D \cite{kong2023robo3d} provides a comprehensive benchmark for assessing the robustness of 3D detectors and segmentors in out-of-distribution scenarios. Furthermore, \cite{yang2020learn} utilize a confidence-guided framework to improve the accuracies of propagated labels. Contrastive learning, as demonstrated by CoKe \cite{Bai2023}, has also been employed to enhance robustness in keypoint detection, especially in occlusion-prone scenarios.

\textbf{Contrastive Learning.} 
Recently contrastive learning has demonstrated state-of-the-art performance among self-supervised learning (SSL) approaches. 
This strategy has been applied to 3D hand pose and shape estimation \cite{Spurr2021,Zimmermann2021}. 
\huien{Sanyal et al. \cite{sanyal2019} incorporate a novel shape consistency loss for 3D face shape and pose estimation that encourages the face shape parameters to be similar when the identity is the same and different for different people.}
Choi et al. \cite{Choi2023} were the first to apply contrastive learning for 3D human pose and shape estimation. They found that SSL is not useful for this task, as the learned representations could be challenging to embed with high-level human-related information. 
Khosla et al. \cite{Khosla2020} proposed supervised contrastive learning for image classification tasks, which incorporates label information during training. 
\huien{Currently there is not attempt to apply this strategy to human pose and shape estimation, where the definition of positive samples is unclear, and data lie in a continuous space. We are the first to overcome these challenges and integrate supervised contrastive learning with whole-body pose and shape estimation. }

\textbf{Pixel Alignment in Pose and Shape Estimation.}
Many studies have been done to learn the subject's location in an image. Some works implicitly supervise the location. They primarily utilize projected meshes by supervising 2D joints regressed from the mesh \cite{Kanazawa2017, Kolotouros2019b, Kolotouros2019, Choi2020, Joo2021, Kolotouros2021, Dwivedi2022, Kocabas2022, Kocabas2021}. 
Further supervision, such as dense body landmarks, silhouettes, and body part segmentation, is also employed to better align the predictions with the image \cite{Xu2019, Omran2018, Pavlakos2018,Tung2017,Zhang2020,Zanfir2020,Dwivedi2022}.
Some other works explicitly learn the subject's location. Moon et al. \cite{Moon2022} explicitly predict the keypoint locations in the image. Semantic body part segmentation is used as an explicit intermediate representation \cite{Pavlakos2019,Omran2018}. PARE \cite{Pavlakos2019} employs a renderer to project the ground-truth mesh to the image space, and supervise the predicted part silhouette mask. However, dense \huien{part segmentation} and differentiable rendering have not been employed in whole-body pose and shape estimation, which will be achieved in our framework.

\section{Motivation}
\label{sec:motivation}

\begin{figure}[t]
    \subfigure{
    \includegraphics[width=0.98\linewidth ,keepaspectratio]{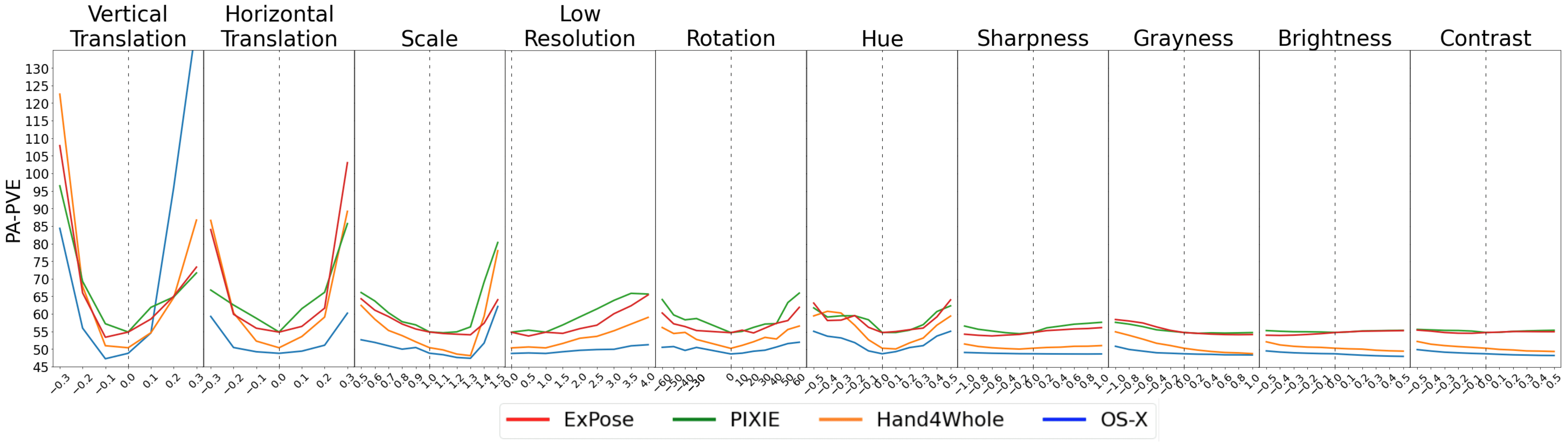}
    }
    \vspace{-10pt}
    \caption{\small \textbf{Wholebody PA-PVE errors under different augmentations (sorted in descending order).} The dashed line indicates baseline performance without augmentation.}
    \vspace{-15pt}
    \label{figure:wholebody_errors_papve_compile}
\end{figure}

As discussed in Section \ref{sec:intro}, existing whole-body pose and shape estimation approaches suffer from the robustness issue, due to the models' sensitivity to the quality of input crops. To investigate the reasons and disclose the influence factors, we conduct a comprehensive evaluation of four state-of-the-art methods: ExPose \cite{Choutas2020}, PIXIE \cite{Feng2019}, Hand4Whole \cite{Moon2022} and OS-X \cite{Lin2023}. We opt for a set of ten commonly encountered augmentations and vary their scales within a realistic range (see Appendix \ref{sec:augmentation-robustess} for more details). The augmentations can be classed into three categories (1) \textit{image-variant} augmentations: they affect the image without altering the objects' 3D poses or positions, such as color jittering; (2) \textit{location-variant} augmentations: they modify the subject's location without changing its pose, involving operations like translation and scaling; (3) \textit{pose-variant} augmentations: they simultaneously alter both the 3D pose and location, including rotation.

\textbf{Impact of subject localization}. 
We first reveal that existing models demonstrate high sensitivity to the subject's position, indicating potential difficulties in subject localization. Figure \ref{figure:wholebody_errors_papve_compile} reports the PA-PVE errors of the whole body under different augmentations. We observe that image-variant augmentations (contrast, sharpness, brightness, hue and grayscale) lead to an acceptable range of error rates (approximately in the 50s) and minimal fluctuation (around $\pm2$). In contrast, location-variant augmentations altering the subject's position within the frame, such as rotation, scaling, and horizontal or vertical translation, result in substantially higher error magnitudes. This demonstrates the heightened sensitivity of existing models to changes in the subject's position. In Appendix, we provide the results of other metrics and benchmarks in Figures \ref{figure:wholebody_errors_compile1} -- \ref{figure:wholebody_errors_compile2}, and visualizations of whole-body estimation under different settings in Figures \ref{figure:wb_vis_translatex} -- \ref{figure:wb_vis_translatey}.




\begin{wrapfigure}{r}{0.47\textwidth} 
\vspace{-15pt}

    \centering
    \subfigure{
    \includegraphics[width=\linewidth,keepaspectratio]{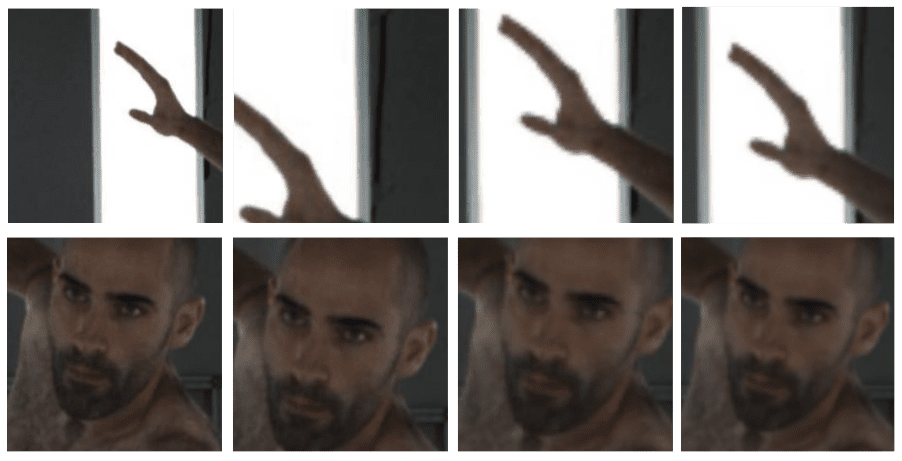}
    }
    \vspace{-17pt}
    \caption{\small \textbf{Crops from (a) ExPose \cite{Choutas2020} (b) PIXIE \cite{Feng2019}, (c) Hand4Whole \cite{Moon2022} (d) \Name.}}
    \label{figure:crops}
\vspace{-10pt}
\end{wrapfigure}

Such position-altering augmentations are common in real-world scenarios, where the subject in the image is often localized using external detection models and control over the quality of crops is less feasible. In practice, to guarantee the visibility of the subject, crops are often made broader, This can lead to significant performance degradation, as errors increase with smaller augmentation scale factors (<1.0) (Figure \ref{figure:wholebody_errors_papve_compile}). Besides, horizontal and vertical translations, which correspond to scenarios where the subject is not perfectly centralized or entirely visible within the frame, can further decrease the performance. 
Similarly, the alignment and scale of these crops also influence the pose and shape estimation systems targeting body, face and hands (Figure \ref{figure:scale_jitter}, more quantitative and qualitative evidence in Appendix \ref{sec:qualitative_evaluation_aug}). Whole-body methods bear the additional responsibility of accurately localizing body parts such as hands and face. Inaccurate part crops (Figure \ref{figure:crops}) can adversely affect the performance of part subnetworks, and further the whole-body estimation.

\begin{figure}[t]
    \centering
    \subfigure{
    \includegraphics[width=0.98\linewidth ,keepaspectratio]{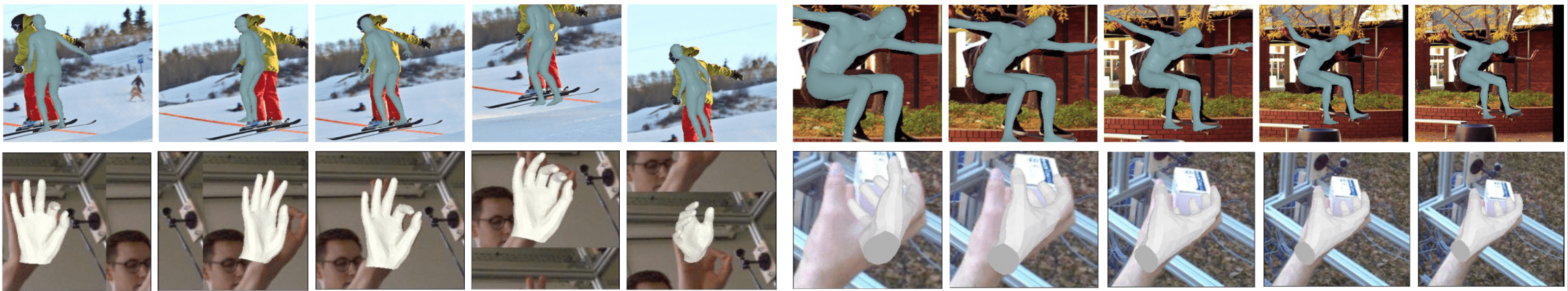}
    }
    \vspace{-10pt}
    \caption{\small \textbf{Sensitivity of existing body and hand models to different alignments (left) and scales (right).}}
    \vspace{-12pt}
    \label{figure:scale_jitter}
\end{figure}



\textbf{Impact of feature extraction}.
The deterioration of performance in the face of such variations suggests that the model struggles to extract meaningful features. Under alterations in translation or scale, the subject remains within the image frame, though the proportion of background content may vary. It is difficult for existing methods to effectively disregard irrelevant background elements and extract relevant features related to the subject of interest. To enhance the model's robustness, it is critical to produce consistent features irrespective of various augmentations applied to the image. 


\textbf{Impact of output pixel alignment}.
Pixel alignment is a critical aspect of high model performance. In certain instances, despite having precise subject localization, the model fails to produce properly aligned results (Figure \ref{figure:wb_vis_translatex} in Appendix). This is often caused by the suboptimal camera parameter estimation. To address this issue, we need to accurately estimate the camera parameters, ensuring the projected mesh is precisely aligned with the ground-truth at the pixel level. Such precision would enhance the effectiveness of the model in producing accurate pose, shape and camera parameter predictions, improving the overall accuracy and reliability of the estimation process.

\section{\Name Framework}

We design \Name to enhance the robustness of whole-body pose and shape estimation. 
It provides three specialized modules to address each challenge in Section \ref{sec:motivation}: 1) \textbf{Localization Module} (Section \ref{localization_module}): explicitly learning the location information of the subject and incorporating it into model estimations for pose, shape and camera ; 2) \textbf{Contrastive Feature Extraction Module} (Section \ref{contrastive_module}): reliably extracting pertinent features under various augmentations, thereby improving the model's generalization ability and robustness to a broader range of real-world scenarios; 3) \textbf{Pixel Alignment Module} (Section \ref{pixel_alignment_module}): ensuring that the outputs are pixel aligned. 

\begin{figure*}[t]
    \centering
    \subfigure{
    \includegraphics[width=0.98\linewidth ,keepaspectratio]{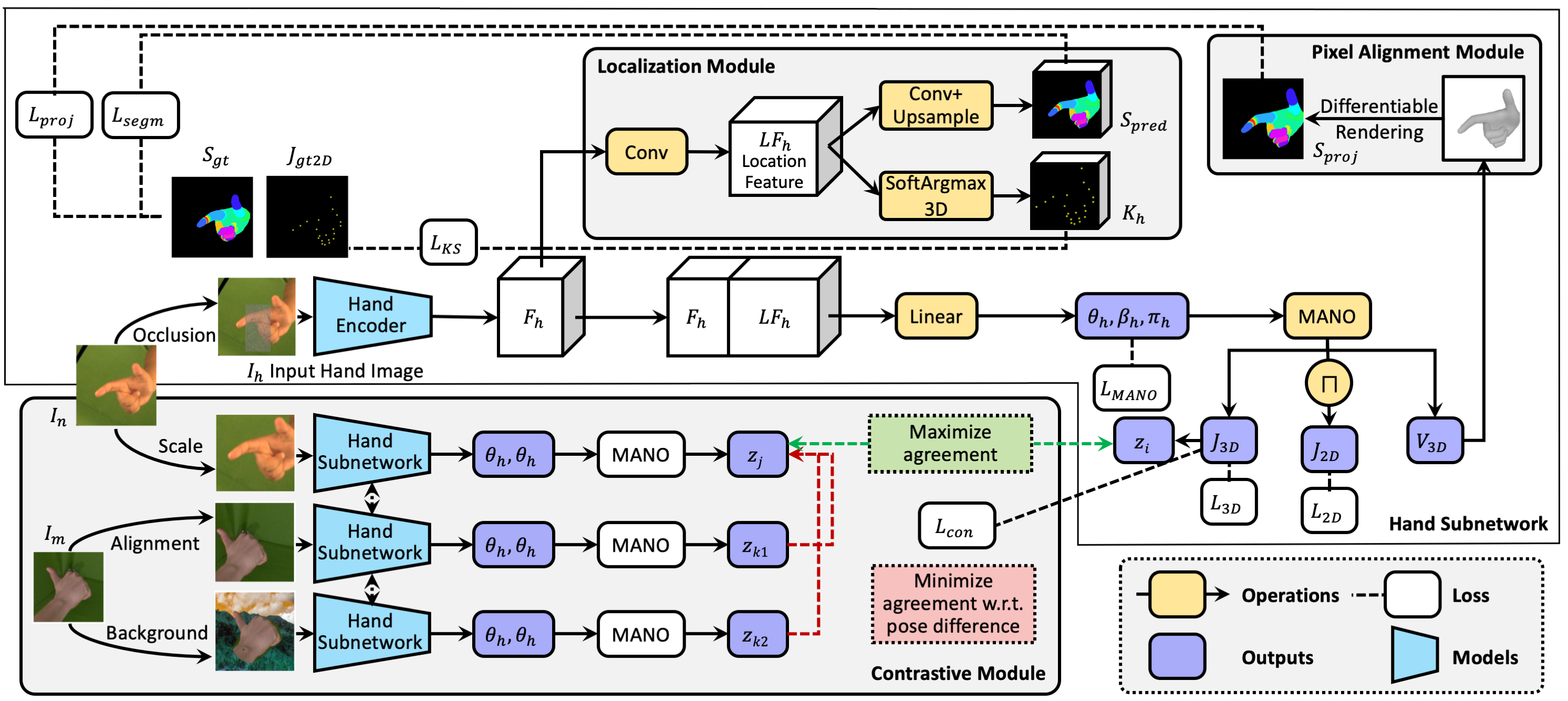}
    }
    \vspace{-10pt}
    \caption{\small \textbf{Subnetwork Architecture with three modules.} We use the Hand subnetwork as an example. $z$ represents normalized $J_{3D}$ while $p$ corresponds to the ground-truth of $z$. \textcolor{ForestGreen}{Green} and \textcolor{red}{red} dashed lines refers to contrastive loss for positive and negative samples respectively. }
    \vspace{-10pt}
    \label{figure:modules_combine}
\end{figure*}

We start with the description of \Name architecture with Body, Hand and Face subnetworks (Section \ref{wholebody_network}). Each subnetwork is integrated with the \textbf{Localization Module} and \textbf{Pixel Alignment Module}, and applies the \textbf{Contrastive Feature Extraction Module} for learning more robust features. Figure \ref{figure:modules_combine} shows the Hand subnetwork architecture. The other two subnetworks have the same designs.

\subsection{Architecture and Training Details}
\label{wholebody_network}

\begin{figure*}[t]
    \centering
    \subfigure{
    \includegraphics[width=0.95\linewidth ,keepaspectratio]{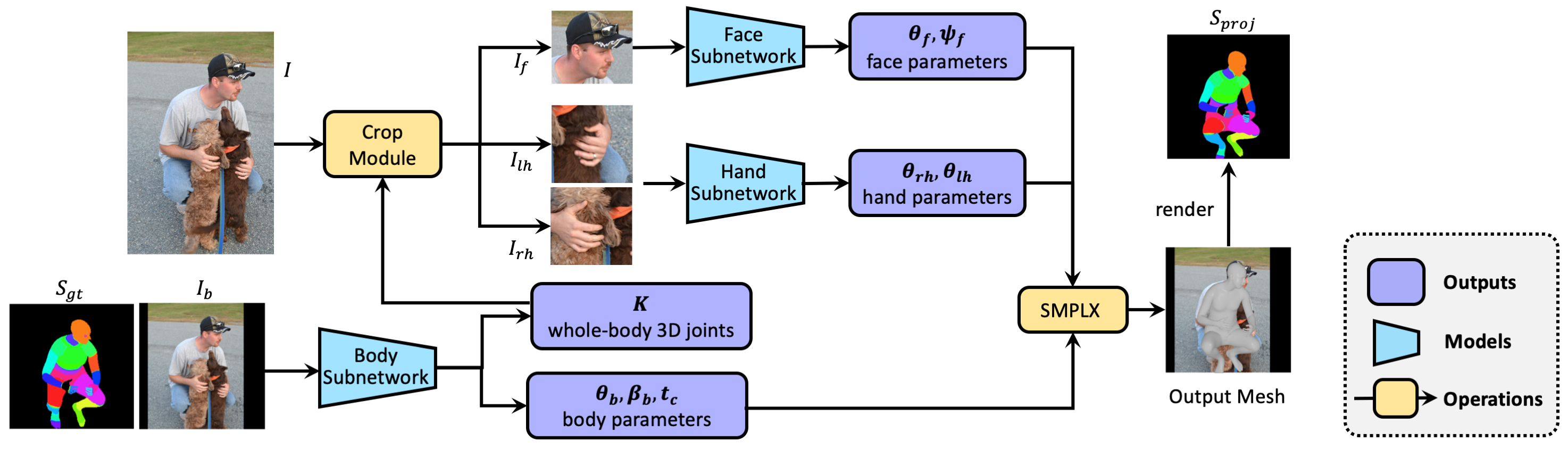}
    }
    \vspace{-10pt}
    \caption{\small \textbf{Pipeline of our \Name framework consisting of Body, Hand and Face subnetworks. }}
    \label{figure:wholebody_archi}
\end{figure*}

Figure \ref{figure:wholebody_archi} shows the overall pipeline of \Name for whole-body 3D human pose and mesh estimation. The Body subnetwork outputs 3D body joint rotations $\theta_{b}$ \( \in \mathbb{R}^{21\times3}\), global orientation $\theta_{bg}$ \( \in \mathbb{R}^{3}\), shape parameters $\beta_{b}$ \( \in \mathbb{R}^{10}\), camera parameters $\pi_{b}$ \( \in \mathbb{R}^{3}\), and whole-body joints \( K \in \mathbb{R}^{137 \times 3}\).  Joints corresponding to the hand and face are used to derive bounding boxes. Subsequently, hand and face images are cropped from a high-resolution image to preserve details. The Hand subnetwork predicts left and right hand 3D finger rotations $\theta_{h}$ \( \in \mathbb{R}^{15\times3}\). Simultaneously, the Face subnetwork generates 3D jaw rotation $\theta_{f}$ \( \in \mathbb{R}^{3}\) and expression $\psi_{f}$ \( \in \mathbb{R}^{10}\). When training Hand and Face subnetworks with part-specific datasets, additional parameters such as global orientation $\theta_{fg} \in \mathbb{R}^{3}$, shape $\beta_{f} \in \mathbb{R}^{50}$, and camera $\pi_{f} \in \mathbb{R}^{3}$ are estimated. These branches are discarded during whole-body estimation and training. 
\huien{Additional information concerning each subnetwork can be found in Appendix \ref{sec:experiment-setup-extras}. Further details regarding the training and inference durations are elaborated upon in Appendix \ref{sec:runtime}.} 

Subnetworks are trained separately, then integrated in a multi-stage manner. Initial whole-body training runs for 20 epochs. The hand and face modules are substituted with the trained Hand and Face subnetworks, followed by 20 epochs of fine-tuning to better unify the knowledge from the Hand and Face subnetworks into the whole-body understanding. Each subnetwork is trained by minimizing the following loss function \emph{L}:
\begin{equation}
L = \lambda_{3D}L_{3D}+\lambda_{2D}L_{2D}+\lambda_{BM}L_{BM}+\lambda_{proj}L_{proj}+\lambda_{segm}L_{segm}+\lambda_{con}L_{con}
\end{equation}
Here $L_{BM}$ is the L1 distance between the predicted and ground-truth body model parameters. $L_{3D}$ denotes the L1 distance between 3D keypoints and joints regressed from the body model. $L_{2D}$ signifies the L1 distance of the ground-truth 2D keypoints to predicted and projected 2D joints. The latter are obtained by projecting the regressed 3D coordinates from the 3D mesh to the image space using the perspective projection \cite{Kanazawa2017}. The part segmentation loss $L_{segm}$ is the cross-entropy loss between $P_{h, w}$ after softmax and $P_{h, w}$ averaged over H×W elements, following \cite{Kocabas2021}. $L_{proj}$ refers to the projected segmentation loss, which is the sigmoid loss between the projected mesh and the ground-truth segmentation map. $L_{con}$ is the contrastive loss described in Section \ref{contrastive_module}. For wholebody training, $L_{box}$ is added to measure the L1 distance between the predicted and actual center and scale of the hands' and face's boxes.


\begin{figure}[t]

\vspace{-5pt}
    \centering
    \subfigure{
    \includegraphics[width=0.98\linewidth ,keepaspectratio]{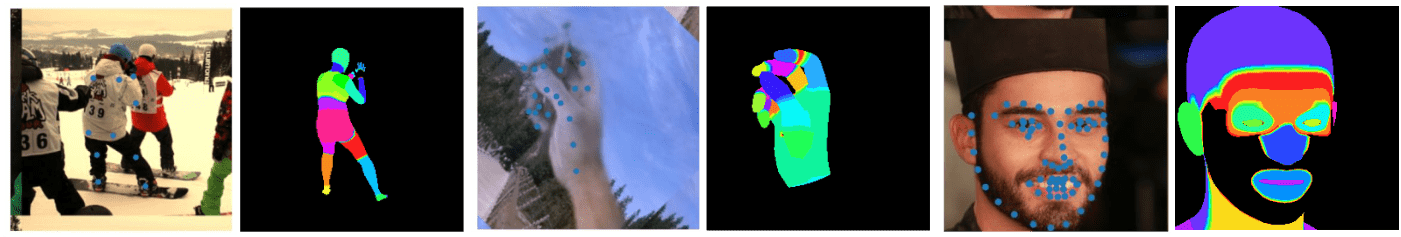}
    }
    \vspace{-10pt}
    \caption{\small \textbf{Examples of keypoint and part segmentation supervision for Body, Hand and Face subnetworks.}}
    \label{figure:smpl_mano_flame_supervision}
\vspace{-15pt}
\end{figure}




\subsection{Localization Module} 
\label{localization_module}

This module focuses on subject localization by explicitly learning both sparse and dense predictions of the subject within the image. Figure \ref{figure:smpl_mano_flame_supervision} shows an example of the supervision used for each subnetwork. In contrast to prior methods that directly output pose rotations from backbone features, this module aims to make the model explicitly conscious of the subject's location and semantics while predicting pose, shape and camera parameters. It can reduce the model's sensitivity to the variations of the subject's position, caused by minor shifts in the scale and alignment of the bounding box.

As shown in Figure \ref{figure:modules_combine}, given an image, a convolutional backbone is utilized to extract its feature map \emph{F} \( \in \mathbb{R}^{512 \times 32 \times 32}\). Following \cite{Moon2022}, a 1$\times$1 convolutional layer is then used to predict 3D feature maps \emph{LF} \( \in \mathbb{R}^{32J \times 32 \times 32}\) from \emph{F}, where \emph{J} represents the number of predicted joints with a feature map depth of 32. \emph{LF} contains valuable information about the mesh's position in the image and semantics of various parts. It is concatenated with the backbone feature map \emph{F} to predict pose \( \theta \in \mathbb{R}^{P}\), shape \( \beta \in \mathbb{R}^{10}\) and camera translation \(\pi \in \mathbb{R}^{3}\), where $P$ is the number of body parts. Meanwhile, \emph{LF} is also used to obtain extra information with two branches: (1) 3D joint coordinates \emph{K}\( \in \mathbb{R}^{J \times 3}\) are obtained from \emph{LF} using the soft-argmax operation \cite{Sun2018} in a differentiable manner. (2) 2D part segmentation maps \emph{S} \( \in \mathbb{R}^{P+1 \times 64 \times 64}\) are extracted from \emph{LF} with several convolution layers, which model \emph{P} part segmentation and 1 background mask. Here, 64 represents the height and width of the feature volume, and each pixel $(h,w)$ stores the likelihood of belonging to a body part \emph{P}.

Note that learning part segmentation maps and 3D joint coordinates is complementary, as 3D joint coordinates encode depth information that may inform part ordering in segmentation maps. Additionally, joints often reside at the boundaries of part segmentation maps, serving as separators for distinct parts. The Body subnetwork utilizes 24 parts \emph{P} and 137 joints \emph{J}, the Hand subnetwork employs 16 parts \emph{P} and 21 joints \emph{J}, while the Face subnetwork employs 15 parts \emph{P} and 73 joints \emph{J}.

\begin{figure*}[t]
    \centering
    \subfigure{
    \includegraphics[width=0.98\linewidth ,keepaspectratio]{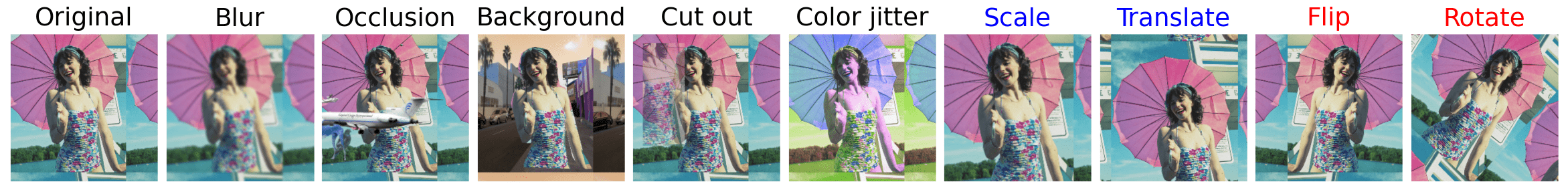}
    }
    \vspace{-10pt}
    \caption{\small \textbf{Augmentations for the Body subnetwork.} Black, \textcolor{blue}{blue} and \textcolor{red}{red} labels represent image-variant, location-variant and pose-variant augmentations, respectively.}
    \label{figure:simclr_body_aug}
    \vspace{-20pt}
\end{figure*}

\subsection{Contrastive Feature Extraction Module} 
\label{contrastive_module}

This module incorporates a pose- and shape-aware contrastive loss, along with positive samples.
By minimizing this loss, the model can produce consistent representations for the same subject, even when presented with different augmentations, thus fostering the extraction of meaningful features.

Conventional contrastive learning methods based on SSL (e.g., SimCLR) face challenges in unifying similar pose embeddings and distancing dissimilar ones in human pose and shape estimation tasks. Without labels for guidance, images with similar poses could be misidentified as negative samples and contrasted away, complicating the self-organization of the embeddings in pose space. 
Figures \ref{figure:embedding1} to \ref{figure:embedding4} in Appendix show their ineffectiveness for the 3D human pose and shape task \cite{Choi2023} by visualizing the retrieved samples from the embeddings.
The supervised contrastive learning approach by Khosla et al. \cite{Khosla2020}, though effective for image classification, might not extend well to human pose and shape estimation, which is a high-dimensional regression problem and poses exist in a continuous space rather than well-defined classes. 

Our module overcomes the aforementioned issues with two innovations. First, we experiment with three human pose representations $\boldsymbol{z}$ and the corresponding distance functions: (1) A concatenated form of the global orientation and rotational pose; (2) global orientation and rotational pose as separate entities (3) 3D root-aligned joints regressed from the body model, derived from pose and shape inputs. For (1) and (2), we explore relative rotations in two forms: 6D vector and rotation matrix representation. For (3), L1, Smooth L1, and Mean Squared Error (MSE) was used (Table \ref{table:hand_cl_representation}).

Second, we investigate ten data augmentations, and classify them into three categories (see Figure \ref{figure:simclr_body_aug} for the Body subnetwork, and Figure \ref{figure:simclr_hand_aug} in Appendix for Hand subnetwork): (1) \textit{image-variant} augmentations such as color jittering, blur, occlusion and background swapping; (2) \textit{location-variant} augmentations involving translation and scaling; (3) \textit{pose-variant} augmentations including rotation and horizontal flipping. Our ablation study in Table \ref{table:hand_cl_augmentation} shows that augmentations with varied global orientation are detrimental to the model performance. Consequently, we exclude such modifications when constructing positive pairs. Instead, each positive sample is constructed utilizing a random combination of location-variant and color-variant augmentations.



Formally, for a batch of $N$ images, we construct another $N$ images by applying augmentation to each sample. For each anchor $i$, let $j$ be the corresponding augmented sample. Then $i$ is contrasted against $2N - 1$ terms (1 positive and $2N-2$ negatives). The loss takes the following form:
\vspace{-5pt}
\begin{equation}
\mathcal{L}_{con} = \sum_{i=1}^{N} \left(\tau_{pos} \left(\left|\operatorname{d}\left(\boldsymbol{p}_i, \boldsymbol{p}_j\right)  - \operatorname{d}\left(\boldsymbol{z}_i, \boldsymbol{z}_j\right)\right|\right) + \tau_{neg}\sum_{k=1}^{2 N} \mathbb{1}_{[k \neq i, j]}\left(\left|\operatorname{d}\left(\boldsymbol{p}_i, \boldsymbol{p}_k\right)  - \operatorname{d}\left(\boldsymbol{z}_i, \boldsymbol{z}_k\right)\right|\right)\right)
\end{equation}
where $\boldsymbol{z}_i$, $\boldsymbol{z}_j$ and $\boldsymbol{z}_k$ denote the predicted pose representations, and $\boldsymbol{p}_i$, $\boldsymbol{p}_j$ and $\boldsymbol{p}_k$ denote the ground-truth pose representations for the anchor, positive and negative samples in the batch. The objective of this loss function is to minimize the distance between the positive pairs and maximize the distance between the negative pairs, in alignment with the pose similarity. Note that unlike traditional approaches where the distance is the same for all negative samples, the pairwise distance  $\operatorname{d}(\boldsymbol{p}_i$,  $\boldsymbol{p}_k)$ varies depending on the pose similarity.

\subsection{Pixel Alignment Module}
\label{pixel_alignment_module}

This module employs differentiable rendering to ensure that the projected mesh aligns precisely at the pixel level. The alignment is supervised by the projected mask loss.  Attaining a proper alignment between the ground-truth part segmentation and rendered mesh requires the accurate prediction of pose, shape, and camera parameters, which subsequently leads to a more precise estimation process.

\vspace{-5pt}
\section{Experiments}

%


\textbf{Datasets.}
For whole-body training, we employ Human3.6M (H36M) \cite{Ionescu2014}, COCO-Wholebody \cite{Jin2020} (the whole-body version of MSCOCO \cite{Lin2014}) and MPII \cite{Andriluka2014}. The 3D pseudo-ground truths for training are acquired using NeuralAnnot \cite{Moon2022b}. For hand-specific training, we use FreiHAND \cite{Zimmermann2019}, Interhand \cite{MoonInterhand2020} and COCO-Wholebody Hands \cite{Jin2020}. For face-specific training, we use FFHQ \cite{Karras2021}, BUPT \cite{Wang2020} and AffectNet \cite{Mollahosseini2019}. For evaluations specific to 3D body, 3D hand, and 3D face, we utilize 3DPW \cite{VonMarcard2018}, FreiHAND \cite{Zimmermann2019}, and Stirling \cite{Feng2019}, respectively. For the 3D whole-body evaluation, we use EHF \cite{Pavlakos2019} and AGORA \cite{Patel2021}. Additionally, we present qualitative results on the MSCOCO validation set. 


\textbf{Metrics.}
\label{evaluation_Metrics}
Mean Per Joint Position Error (MPJPE) and Mean Per-Vertex Position Error (MPVPE) are employed to evaluate the positions of 3D joint and mesh vertices, respectively. Each metric calculates the average 3D joint distance (in \emph{mm}) and 3D mesh vertex distance (in \emph{mm}) between the predicted and ground-truth values after aligning the root joint translation. The pelvis serves as the root joint for whole-body and body, whereas the wrists and neck are utilized as root joints for hands and face. Procrustes Aligned (PA) variants of these metrics, PA-MPJPE and PA-MPVPE, further align with rotation and scale. We report the average errors for the left and right hands as the 3D hand error.


\begin{table}
\vspace{-15pt}
\parbox{.53\linewidth}{
\small
\caption{\small \textbf{Evaluation of the Hand subnetwork}.}%
\resizebox{.53\textwidth}{!}{
\begin{tabular}{@{}lrrl@{}}
\toprule

\textbf{Method}                       & \multicolumn{1}{l}{PA-PVE $\downarrow$} & \multicolumn{1}{l}{PA-MPJPE $\downarrow$} & F-Scores $\uparrow$ \\ \midrule
\textbf{* Hand-only} & \multicolumn{1}{l}{}         & \multicolumn{1}{l}{}           &                       \\
FreiHAND \cite{Zimmermann2019}             & 10.7                         & \multicolumn{1}{l}{-}          & 0.529/0.935           \\
Pose2Mesh \cite{Choi2020}           & 7.8                          & 7.7                            & 0.674/0.969           \\
I2L-MeshNet \cite{Moon2020}          & 7.6                          & 7.4                            & 0.681/0.973           \\
METRO (HR64) \cite{Lin2021}               & \textbf{6.7}                          & \textbf{6.8}                            & 0.717/0.981           \\
\midrule
\textbf{* Whole-body} & \multicolumn{1}{l}{}         & \multicolumn{1}{l}{}           &                       \\
ExPose \cite{Pavlakos2019}              & 11.8                         & 12.2                           & 0.484/0.918           \\
Zhou et al. \cite{Zhou2021}        & -        & 15.7                           & -/-                   \\
FrankMocap \cite{Rong2021}          & 11.6                         & 9.2                            & 0.553/0.951           \\
PIXIE \cite{Feng2019}              & 12.1                         & 12                             & 0.468/0.919           \\
Hand4Whole $\dagger$ \cite{Moon2022}          & 7.7                          & 7.7                            & 0.664/0.971           \\
HMR \huien{(Baseline)} \cite{Kanazawa2017}                   & 8.6                          & 8.9                            & 0.605/0.963           \\
PyMAF  \cite{Zhang2022}                       & 8.1                          & 8.4                            & 0.638/0.969           \\
PyMAF $\dagger$ \cite{Zhang2022}                       & 7.5                          & 7.7                            & 0.671/0.974           \\ 
\textbf{\Name}                        &  \textbf{7.3}                          & \textbf{7.5}                            & 0.683/0.976          \\ 
\textbf{\Name $\dagger$}                       & \textbf{7.1}                          & \textbf{7.4}                            & 0.688/0.978          \\ 
\textbf{\Name (HR64)}                       & \textbf{6.7}                          & \textbf{6.9}                            & 0.715/0.981            \\ 
\bottomrule
\label{table:freihand}
\end{tabular}}
}
\hfill
\parbox{.44\linewidth}{
\vspace{-9pt}
\small
\caption{\small \textbf{Evaluation of the Body subnetwork}. }%
\resizebox{.43\textwidth}{!}{
\begin{tabular}{@{}llll@{}}
\toprule
\textbf{Method}                     & PA-MPJPE $\downarrow$  & MPJPE $\downarrow$  & PVE $\downarrow$ \\ \midrule
HMR (Res50) \cite{Kanazawa2017}      & 76.7  & 130  & -            \\
GraphCMR (Res50) \cite{Kolotouros2021} & 70.2  & -     & -          \\
SPIN (Res50) \cite{Kolotouros2019}    & 59.2  & 96.9  & 116.4      \\
HMR-EFT (Res50) \cite{Joo2021} & 54.3  & -     & -         \\
ROMP (Res50)    & 53.5  & 89.3  & 105.6    \\
PARE (Res50) \cite{Kocabas2021}     & 52.3  & 82.9 & 99.7      \\
PARE (HR32) \cite{Kocabas2021}    & 50.9  & 82    & 97.9       \\
PyMAF (Res50)\cite{Zhang2022}           & 49.0   & 79.7  & 94.4       \\
PyMAF (HR48) \cite{Zhang2022}          & \textbf{47.1}   & \textbf{78.0}  & 91.3        \\
Baseline (Res50)        & 52.4 & 85.2  & 103.6                   \\
\Name (Res50)            &  49.8 & 80.8   & 96.7                 \\
Baseline (HR48)        & 50.3   & 84.5   & 101.5                \\
\Name (HR48)             & 48.5  & 80.1    & 95.2               \\ \bottomrule
\label{table:pw3d}
\end{tabular}}

\small
\vspace{-10pt}
\centering
\caption{\small \textbf{Evaluation of the Face subnetwork}.}%
\resizebox{.40\textwidth}{!}{
\begin{tabular}{@{}l|ll@{}}
\Xhline{1pt}
\textbf{Method} & LQ Mean(mm) $\downarrow$ & HQ Mean(mm) $\downarrow$ \\ \Xhline{1pt}
ExPose \cite{Choutas2020}   &  2.27 & 2.42  \\
ExPose $\dagger$        &  2.46 & 2.38  \\
\huien{HMR} 	& 2.18 	& 2.11 \\
\huien{HMR $\dagger$} 	& 2.31 	& 2.27 \\
\huien{HMR *} & 2.02 & 2.04 \\
\huien{PyMAF *} & 1.97 & 1.92 \\
\textbf{\Name}        &  \textbf{2.12} & \textbf{2.08}  \\
\Name $\dagger$        &  \textbf{2.12} & \textbf{2.10}  \\  \bottomrule
\end{tabular}
\label{table:stirling}
}}
\vspace{-15pt}
\end{table}

\subsection{Benchmarking Results}

\textbf{Hand Subnetwork}.
Table \ref{table:freihand} compares the performance of the Hand subnetwork with different hand-only and whole-body methods. Our method outperforms that of our whole-body counterparts when trained with only the FreiHAND dataset (i.e. PIXIE, Hand4Whole, PyMAF)  or under mixed datasets (i.e. Hand4Whole $\dagger$, PyMAF $\dagger$)\footnote{$\dagger$ denotes training with extra datasets in the following evaluation and tables.} using an identical backbone. Prior research \cite{Moon2020, Tian2022} demonstrated that whole-body methods generally employ a parametric representation of the hand mesh, and are numerically inferior to the non-parametric representation used in recent hand-only methods \cite{Moon2020, Lin2021}. Despite such reported gap,
\Name manages to outperform mesh-based techniques, and achieve comparable results as the state-of-the-art METRO when using the same backbone (HRNet-64). 
Table \ref{table:hand_jitter_scale} compares the estimation errors of the Hand subnetwork in Hand4Whole (current whole-body method with SOTA on hands) and \Name under different positional augmentations on the FreiHAND test set. It is clear that \Name exhibits much better robustness than Hand4Whole. More visualizations are provided in Figure \ref{figure:hand_aug_vis} in Appendix.

\begin{table}[t]
\small
\vspace{-5pt}
\caption{\small \textbf{PA-PVE/PVE errors of the Hand subnetwork under different positional augmentations.}}%
\resizebox{0.95\textwidth}{!}{
\begin{tabular}{@{}llllllll@{}}
\toprule
       & Normal               & Transx +0.2x         & Transx -0.2x         & Transy +0.2y         & Transy -0.2y         & Scale 1.3x           & Scale 0.7x           \\ \midrule
Hand4Whole \cite{Moon2022}    & 7.47/ 15.70          & 8.51/ 21.58          & 8.38/ 20.36          & 8.74/ 22.51          & 8.48/ 19.85          & 7.73/ 16.44          & 7.78/ 17.00          \\
\Name   & \textbf{7.24/ 15.23} & \textbf{7.27/ 15.62} & \textbf{7.36/ 15.59} & \textbf{7.28/ 15.50} & \textbf{7.34/ 15.50} & \textbf{7.49/ 15.90} & \textbf{7.45/ 16.51} \\ \bottomrule

\label{table:hand_jitter_scale}
\end{tabular}}
\vspace{-15pt}
\end{table}

\textbf{Body Subnetwork.}
Table \ref{table:pw3d} compares the performance of the Body subnetwork across different methods on the 3DPW test set. We observe the competitiveness of \Name in relation to other SMPL-based approaches. Besides, since the performance of various methods may significantly differ based on their backbone initialization, datasets and training strategies \cite{Pang2022}, we establish a baseline to evaluate the effectiveness of our added modules in Table \ref{table:body_ablation} in Appendix. \Name achieves a substantial improvement compared to the baseline. 

\begin{figure}[!ht]
    \subfigure{
    \includegraphics[width=0.9\linewidth ,keepaspectratio]{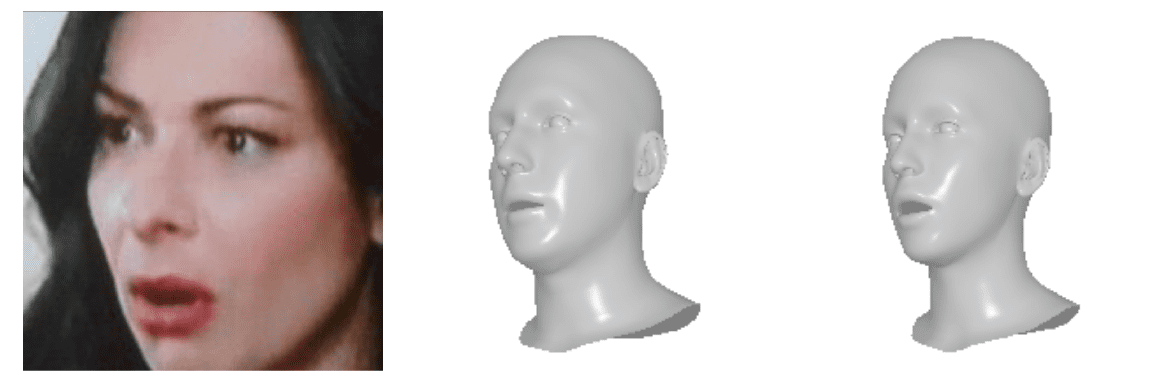}
    }
    \subfigure{
    \includegraphics[width=0.9\linewidth ,keepaspectratio]{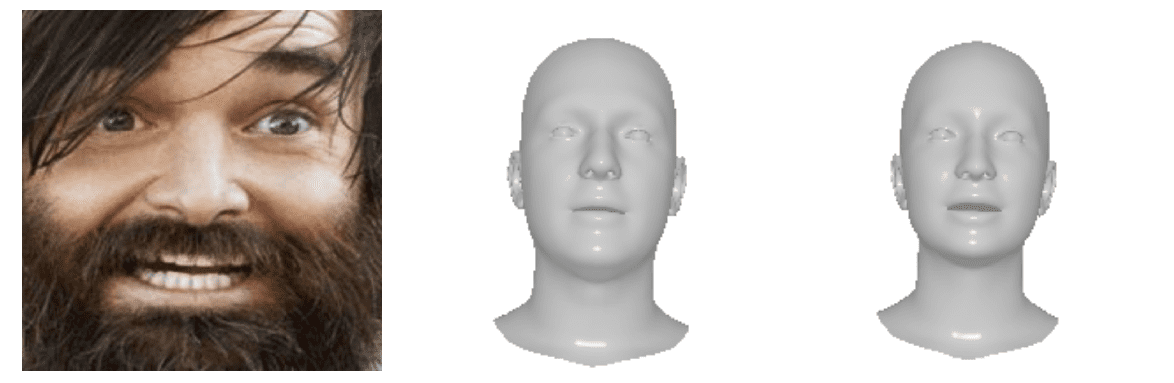}
    }
    \subfigure{
    \includegraphics[width=0.9\linewidth ,keepaspectratio]{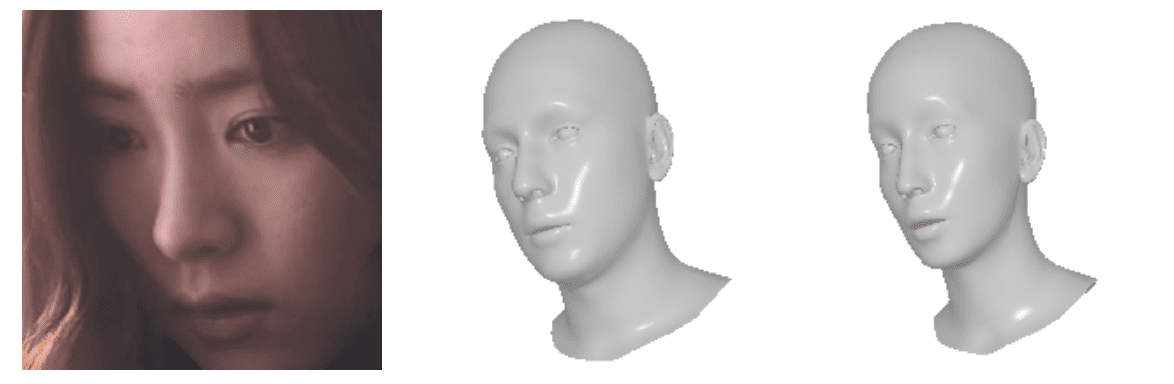}
    }
    \subfigure{
    \includegraphics[width=0.9\linewidth ,keepaspectratio]{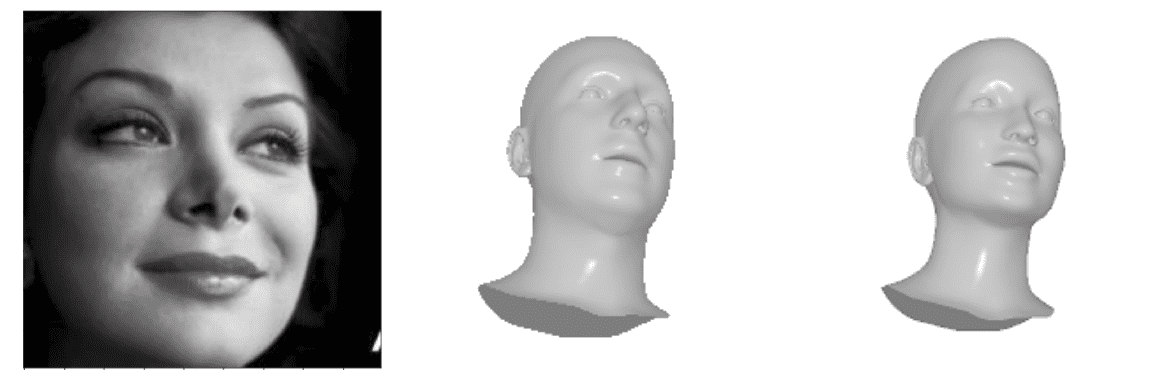}
    }
    \subfigure{
    \includegraphics[width=0.9\linewidth ,keepaspectratio]{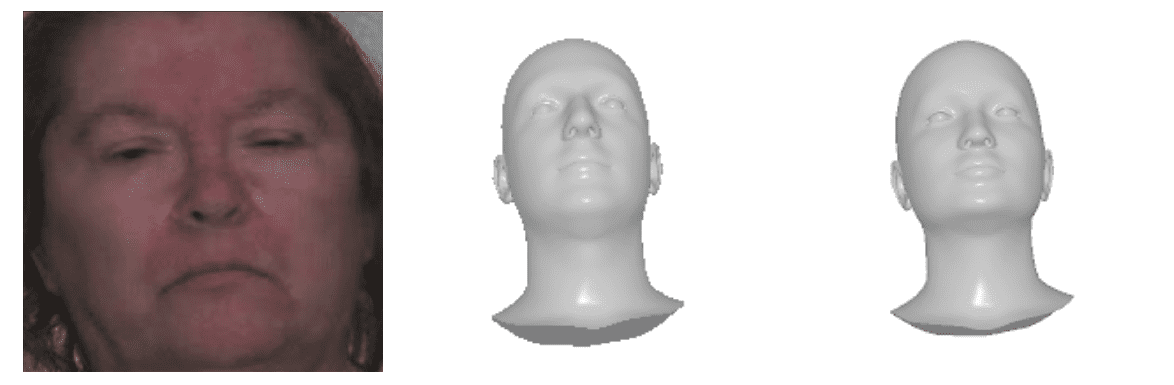}
    }
    \caption{\small \textbf{Inference on AffectNet validation images using Expose \cite{Choutas2020} and \Name's Face subnetwork.}}
    \label{figure:face}
\end{figure}

\textbf{Face Subnetwork.}
Table \ref{table:stirling} compares the performance of the Face subnetwork for different methods on the Stirling3D test set. When training with the same dataset, \Name outperforms ExPose. The performance of ExPose declines when training on multiple datasets, while \Name can still keep low and consistent errors. Figure \ref{figure:face} in Appendix shows some qualitative results for the in-the-wild scenarios, which demonstrates the high generalization of \Name. Table \ref{table:face_jitter_scale} compares the robustness of ExPose and \Name under different positional augmentations. We also observe that \Name has lower errors with different translation and scaling operations. More visualizations are provided in Figure \ref{figure:face_aug_vis} in Appendix.



%
\begin{table}[t]
\small
\centering
\vspace{-10pt}
\caption{\small \textbf{3DRMSE errors of the Face subnetwork under different positional augmentations.}}%
\resizebox{0.85\textwidth}{!}{

\begin{tabular}{@{}llllllll@{}}
\toprule
       & Normal               & Transx +0.2x         & Transx -0.2x         & Transy +0.2y         & Transy -0.2y         & Scale 1.3x           & Scale 0.7x           \\ \midrule
ExPose \cite{Choutas2020} & 2.27                  &       2.38               &               2.29       &         2.46          &       2.30                &           2.46             &          2.27            \\
\Name   & \textbf{2.12}                  &         \textbf{2.20}             &          \textbf{2.17}            &             \textbf{2.13}       &              \textbf{2.18}          &            \textbf{2.24}      &         \textbf{2.10}             \\ \bottomrule

\label{table:face_jitter_scale}
\end{tabular}}
\vspace{-20pt}
\end{table}

\textbf{Whole-body Network.}
We further provide results of the whole-body network on two benchmarks: EHF val set and AGORA test set in Table \ref{table:ehf_agora}. On EHF, \Name outperforms other full-body approaches, particularly in hand and face performance evaluations, and under different positional augmentations (Table \ref{table:wholebody_jitter_scale}). It gives subpar performance on AGORA as the predominant source of error is the misidentification of individuals under intense person-person occlusion. We give detailed investigation in Appendix \ref{sec:failure_analysis_agora}.

\vspace{-5pt}
\begin{table}[t]
\centering
\small
\vspace{-5pt}
\caption{\small \textbf{Evaluation of wholebody network on EHF and AGORA test set.}}%
\resizebox{0.9\textwidth}{!}{
\begin{tabular}{@{}l|rrrrrr|rrrlrr@{}}
\toprule
\multirow{3}{*}{\textbf{Method}} & \multicolumn{6}{c|}{\textbf{EHF}}                                                                                                                                                                              & \multicolumn{6}{c}{\textbf{AGORA}}                                                                                                                                                                                 \\ \cmidrule(l){2-13} 
                                 & \multicolumn{3}{c|}{PVE $\downarrow$}                                                                       & \multicolumn{3}{c|}{PA-PVE $\downarrow$}                                                                 & \multicolumn{4}{c|}{PVE $\downarrow$}                                                                                                              & \multicolumn{2}{c}{N-PVE $\downarrow$}                                \\ \cmidrule(l){2-13} 
                                 & \multicolumn{1}{c|}{WB} & \multicolumn{1}{c|}{H} & \multicolumn{1}{c|}{F}    & \multicolumn{1}{c|}{WB} & \multicolumn{1}{c|}{H} & \multicolumn{1}{c|}{F} & \multicolumn{1}{c|}{WB} & \multicolumn{1}{c|}{B} & \multicolumn{1}{c|}{F} & \multicolumn{1}{c|}{LH/RH}     & \multicolumn{1}{c|}{WB} & \multicolumn{1}{c}{B} \\ \midrule
ExPose \cite{Choutas2020}                          & 77.1                             & 51.6                            & \multicolumn{1}{r|}{35}            & 54.5                             & 12.8                            & 5.8                             & 217.3                            & 151.5                           & 51.1                            & \multicolumn{1}{l|}{74.9/71.3}          & 265                              & 184.8                          \\
PIXIE  \cite{Feng2019}                          & 89.2                             & 42.8                            & \multicolumn{1}{r|}{32.7}          & 55                               & 11.1                            & \textbf{4.6}                    & 191.8                            & 142.2                           & 50.2                            & \multicolumn{1}{l|}{49.5/49.0}          & 233.9                            & 173.4                          \\
Hand4Whole \cite{Moon2022}                      & 76.8                             & 39.8                      & \multicolumn{1}{r|}{26.1}    & 50.3                             & 10.8                      & 5.8                             & 135.5                            & 90.2                            & 41.6                            & \multicolumn{1}{l|}{46.3/48.1}          & 144.1                            & 96.0                             \\
OSX (ViT-L)             & {\ul 70.8}             & 53.7                  & \multicolumn{1}{r|}{26.4}                  & \textbf{48.7}          & 15.9                  & 6.0                     & \textbf{122.8}         & \textbf{80.2}         & \textbf{36.2}         & \multicolumn{1}{l|}{45.4/46.1}                          & \textbf{130.6}         & \textbf{85.3}         \\

\huien{PyMAF-X (HR48)}                 & \textbf{64.9}          & \textbf{29.7}         & \multicolumn{1}{r|}{{\ul 19.7}}            & 50.2                   & {\ul 10.2}            & 5.5                   & {\ul 125.7}            & {\ul 84}              & {\ul 35}              & \multicolumn{1}{l|}{\textbf{44.6/45.6}}                 & 141.2                  & 94.4                  \\

Ours                    & 73.7                   & {\ul 34.9}            & \multicolumn{1}{r|}{\textbf{17.8}}         & {\ul 49.7}             & \textbf{10.0}           & \textbf{4.6}          & 132.3                  & 85                    & 39.4                  & \multicolumn{1}{l|}{{\ul 45.3/46.1}}                    & {\ul 138.2}            & {\ul 91.5}   
\\
\bottomrule
\end{tabular}}
\label{table:ehf_agora}
\vspace{-10pt}
\end{table}

\begin{table}[t]
\small
\vspace{-15pt}
\caption{\small \textbf{Wholebody, Hand and Face PA-PVE errors under different positional augmentations.}}%
\resizebox{\textwidth}{!}{
\begin{tabular}{@{}llrrrrrrr@{}}
\toprule
                           & Method     & \multicolumn{1}{l}{Normal} & \multicolumn{1}{l}{Transx +0.2x} & \multicolumn{1}{l}{Transx -0.2x} & \multicolumn{1}{l}{Transy +0.2y} & \multicolumn{1}{l}{Transy -0.2y} & \multicolumn{1}{l}{Scale 1.3x} & \multicolumn{1}{l}{Scale 0.7x} \\ \midrule
\multirow{5}{*}{Hands}     & ExPose \cite{Choutas2020}     & 14.39                      & 17.36                            & 17.86                            & 14.93                            & 17.21                            & 14.15                          & 14.56                          \\
                           & PIXIE \cite{Feng2019}     & 14.68                      & 15.05                            & 16.11                            & 15.32                            & 15.85                            & 14.52                          & 14.79                          \\
                           & Hand4Whole \cite{Moon2022} & 10.83                      & 11.15                            & 11.34                            & 10.50                            & 13.70                            & 10.77                          & 11.25                          \\
                           & OSX \cite{Lin2023}       & 15.97                      & 16.42                            & 16.55                            & 16.94                            & 17.86                            & 15.91                          & 17.24                          \\
                           & \Name       & \textbf{10.00}             & \textbf{10.37}                   & \textbf{10.21}                   & \textbf{10.16}                   & \textbf{12.49}                   & \textbf{9.98}                  & \textbf{10.19}                 \\ \midrule
\multirow{5}{*}{Face}      & ExPose \cite{Choutas2020}    & 6.34                       & 10.28                            & 6.71                             & 8.17                             & 6.43                             & 6.24                           & 6.24                           \\
                           & PIXIE \cite{Feng2019}      & 5.63                       & 6.67                             & 6.94                             & 6.53                             & 6.94                             & 5.84                           & 5.84                           \\
                           & Hand4Whole \cite{Moon2022} & 5.81                       & 5.88                             & 5.91                             & 5.74                             & 5.93                             & 5.76                           & 5.76                           \\
                           & OSX \cite{Lin2023}       & 6.09                       & 6.03                             & 6.09                             & 5.83                             & 5.96                             & 5.92                           & 5.92                           \\
                           & \Name       & \textbf{4.65}              & \textbf{5.10}                    & \textbf{5.38}                    & \textbf{4.75}                    & \textbf{5.30}                    & \textbf{4.77}                  & \textbf{5.22}                  \\ \midrule
\multirow{5}{*}{Wholebody} & ExPose \cite{Choutas2020}    & 54.82                      & 61.64                            & 65.98                            & 65.03                            & 65.98                            & 54.03                          & 59.23                          \\
                           & PIXIE \cite{Feng2019}      & 54.85                      & 66.16                            & 69.26                            & 64.83                            & 69.26                            & 56.28                          & 60.31                          \\
                           & Hand4Whole \cite{Moon2022} & 50.37                      & 59.10                            & 67.85                            & {\huien{\ul 64.64}}                            & 67.85                            & 48.10                          & 55.28                          \\
                           & OSX \cite{Lin2023}       & \textbf{48.79}             & \textbf{51.09}                   & {\ul 55.96}                      & 95.97                      & \textbf{55.96}                   & \textbf{47.35}                 & \textbf{50.89}                 \\
                           & \Name       & {\ul 49.79}                & {\ul 52.46}                      & \textbf{53.62}                   & \textbf{61.65}                   & {\ul 63.99}                      & {\ul 47.90}                    & {\ul 51.39}                    \\ \bottomrule
\label{table:wholebody_jitter_scale}
\end{tabular}}
\vspace{-20pt}
\end{table}
\begin{table}[t]
\vspace{-5pt}
\parbox{.30\linewidth}{
\small
\caption{\small \textbf{Ablation of contrastive learning methods and loss.}}%
\resizebox{.30\textwidth}{!}{
\begin{tabular}{@{}lrrrr@{}}
\toprule
Scale factor  & \multicolumn{1}{l}{Mean $\downarrow$} & \multicolumn{1}{l}{Std $\downarrow$} 
\\ \midrule
SimCLR         & 0.227                         & 0.0915  \\
SimCLR (+ pose-variant aug.) & 0.230                         & 0.0911     \\ 
SimCLR (+ background aug.) & 0.222                         & 0.0959     \\ 
SimCLR (+ $L_{con}$) & 0.164                         & 0.0772     \\
HMR & 0.140                         & 0.0823    \\ 
HMR (+ $L_{con}$) & 0.124                         & \textbf{0.0624}     \\
HMR (+ $L_{con}$, +ve samples)  & \textbf{0.119}                        & 0.0679     \\
\bottomrule
\label{table:embedding}
\end{tabular}}
}
\hfill
\parbox{.34\linewidth}{
\small
\caption{\small \textbf{Ablation of different representation for contrastive loss.}}%
\resizebox{.34\textwidth}{!}{
\begin{tabular}{@{}lrrrr@{}}
\toprule
Representation  & \multicolumn{1}{l}{PA-$\downarrow$} & \multicolumn{1}{l}{MPJPE$\downarrow$} & \multicolumn{1}{l}{PA $\downarrow$}  & \multicolumn{1}{l}{PVE$\downarrow$} 
\\ \midrule
baseline & 7.49          &15.51              & 7.46 & 15.59 \\
pose  & 8.11       &      15.81            & 7.67    & 16.08 \\ 
go + pose  & 7.71     & 14.98                    & 7.54  & 14.91  \\ 
keypoint  & 7.48          & 15.01               & 7.32   & 15.29  \\
pose, +ve  & 7.45      &    14.94                & 7.20    & \textbf{14.77} \\ 
keypoint, +ve  & \textbf{7.31}                &\textbf{14.62}            & \textbf{7.18}     & 15.01    \\   
\bottomrule
\label{table:hand_cl_representation}
\end{tabular}}
}
\hfill
\parbox{.33\linewidth}{
\small
\caption{\small \textbf{Ablation of augmentation +ve samples, using pose rotation as representation. }}%
\resizebox{.33\textwidth}{!}{
\begin{tabular}{@{}lrrrr@{}}
\toprule
Augmentation  & \multicolumn{1}{l}{PA-$\downarrow$} & \multicolumn{1}{l}{MPJPE$\downarrow$} & \multicolumn{1}{l}{PA- $\downarrow$}  & \multicolumn{1}{l}{PVE$\downarrow$} 
\\ \midrule
baseline (no +ve) & 8.11       &      15.81            & 7.67    & 16.08 \\
color  & \textbf{\huien{7.42}}                         & 15.01   & \textbf{\huien{7.18}} & 14.94  \\ 
pose  & 8.59      &    16.96                & 8.15    & 17.21 \\ 
location  & 7.80     & 15.98    & 7.46   & 15.56 \\   
color + location  & 7.45    & \textbf{14.94}   & 7.20 & \textbf{14.77} \\ 
\bottomrule
\label{table:hand_cl_augmentation}
\end{tabular}}
}
\vspace{-10pt}
\end{table}

\subsection{Ablation Studies}
\label{ablation_contrastive}

\textbf{Contrastive loss}.
We validate prior contrastive SSL methods \cite{Zimmermann2021,Spurr2021,Choi2023} are not particularly adept at learning useful embeddings for human pose and shape estimation. Figures \ref{figure:embedding1} -- \ref{figure:embedding4} in Appendix visualize the retrieved images based on the top-5 embedding similarity. They show that without labels, the model primarily extracts features based on background information instead of pose information. Table \ref{table:embedding} shows the estimation errors of top-1 retrieved pose (COCO-train) and query pose (COCO-test) with different methods and contrastive loss functions. We observe that SimCLR has higher mean errors than the supervised training method HMR. These results are aligned with \cite{Choi2023} that the representations learned through SSL are not transferable for human pose and shape estimation tasks. \Name incorporates contrastive loss and positive samples ("HMR + $L_{con}$, +ve"), which can produce similar representations under varied augmentations, enhancing its robustness.

\begin{figure}[!htb]
    \centering
    \subfigure{
    \includegraphics[width=\textwidth ,height=\textheight, keepaspectratio]{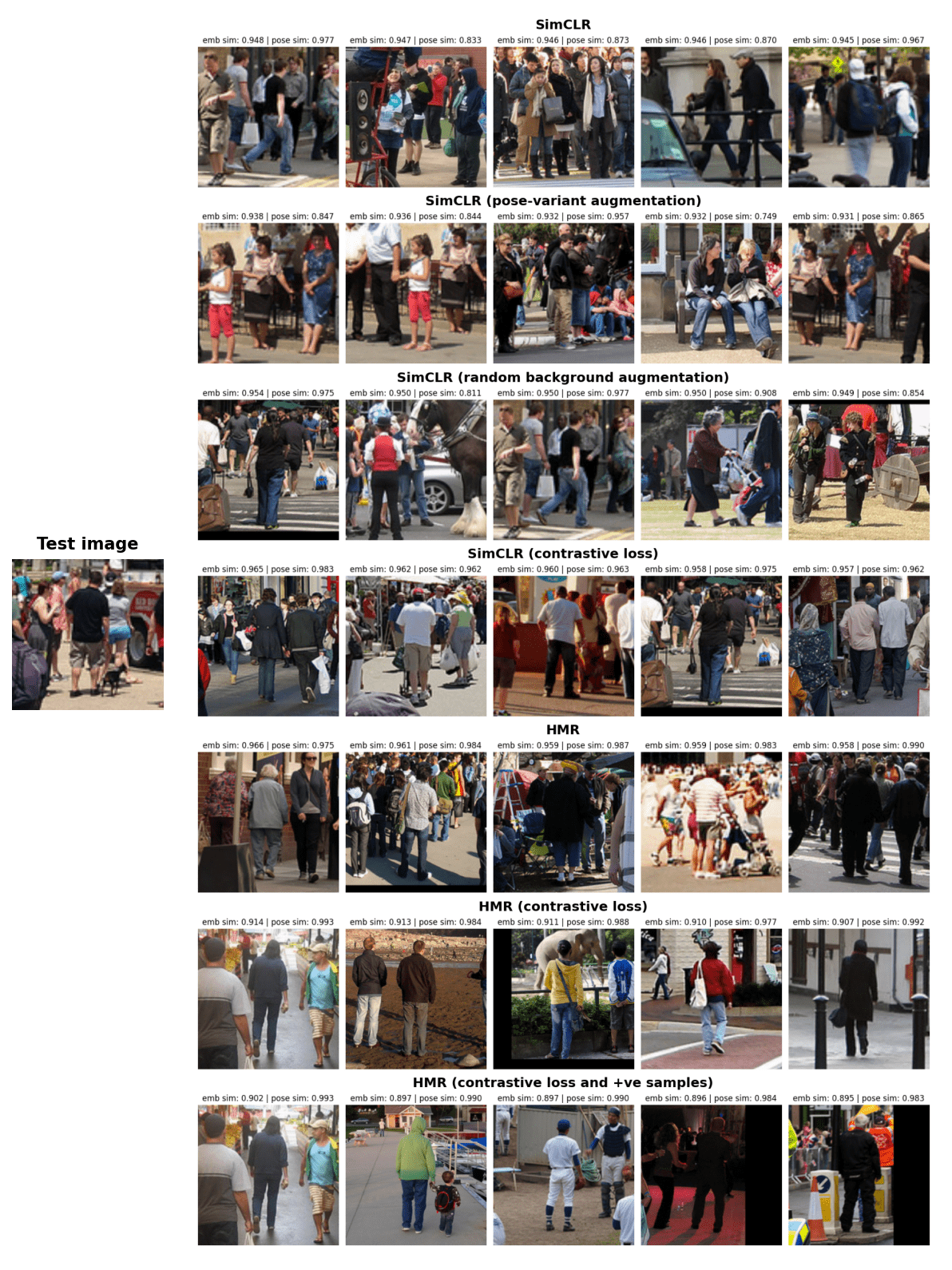}
    }
    \caption{\small \textbf{Left: Query image from the EFT-COCO-Test set, Right: Retrieved image from the EFT-COCO-Train set ordered in descending embedding similarity. }}
    \label{figure:embedding1}
\end{figure}
\begin{figure}[!htb]
    \centering
    \subfigure{
    \includegraphics[width=0.98\linewidth ,keepaspectratio]{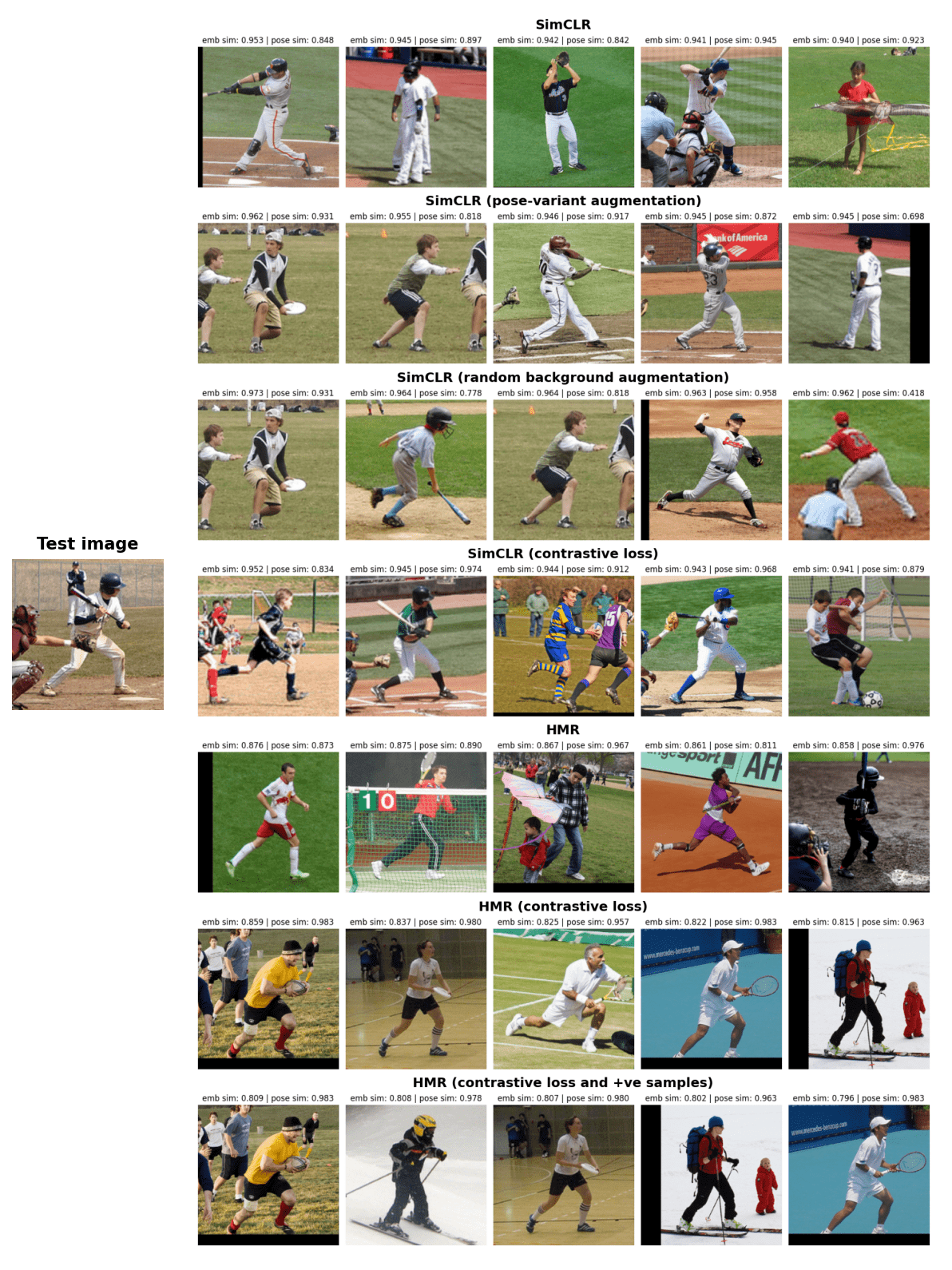}
    }
    \caption{\small \textbf{Left: Query image from the EFT-COCO-Test set, Right: Retrieved image from the EFT-COCO-Train set ordered in descending embedding similarity. }}
    \label{figure:embedding2}
\end{figure}
\begin{figure}[!htb]
    \centering
    \subfigure{
    \includegraphics[width=0.98\linewidth ,keepaspectratio]{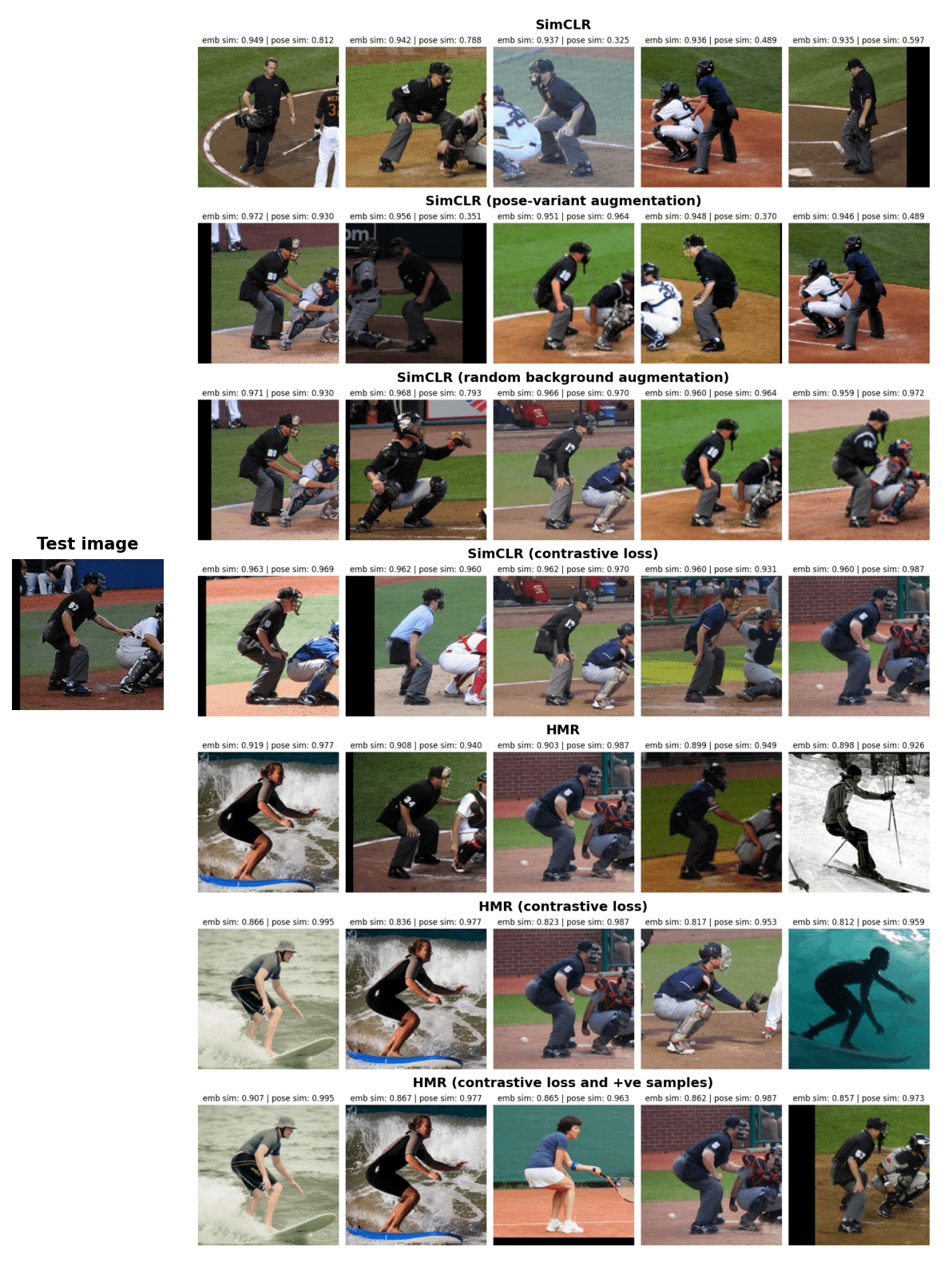}
    }
    \caption{\small \textbf{Left: Query image from the EFT-COCO-Test set, Right: Retrieved image from the EFT-COCO-Train set ordered in descending embedding similarity. }}
    \label{figure:embedding3}
\end{figure}
\begin{figure}[!htb]
    \centering
    \subfigure{
    \includegraphics[width=0.98\linewidth ,keepaspectratio]{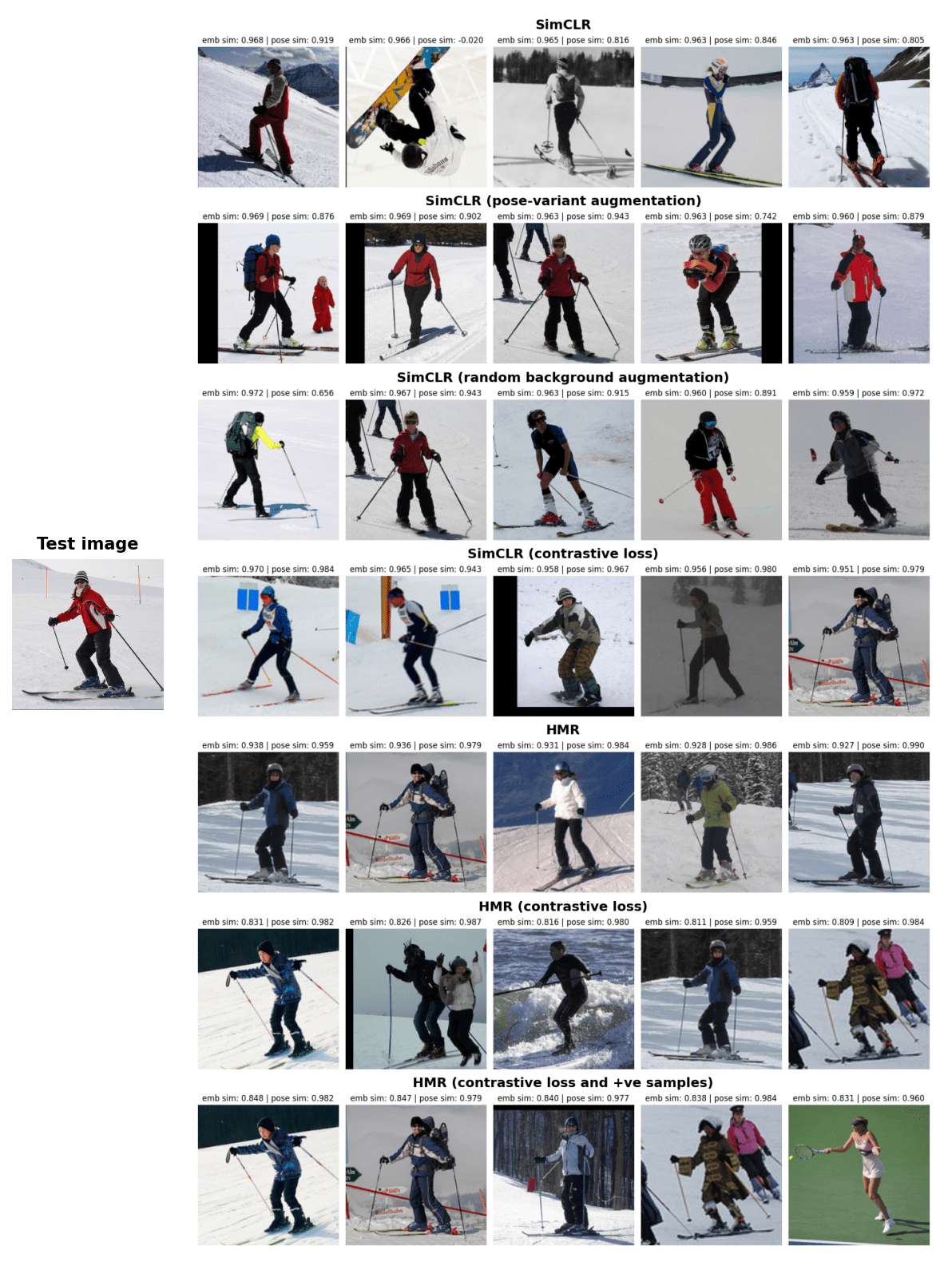}
    }
    \caption{\small \textbf{Left: Query image from the EFT-COCO-Test set, Right: Retrieved image from the EFT-COCO-Train set ordered in descending embedding similarity. }}
    \label{figure:embedding4}
\end{figure}



Table \ref{table:hand_cl_representation} shows the estimation errors when applying contrastive loss with different representations in Section \ref{contrastive_module}: "pose" (a concatenated form of global orientation and rotational pose), "go+pose" (global orientation and rotational pose as separate entities), "keypoint" (3D joints regressed from the body model). We observe that regressed 3D joints are the most effective representation, as they encode both shape and pose information in a normalized space. In contrast, the representation of pose as relative rotation has a detrimental impact on the model performance. 
Incorporating positive samples ("pose, +ve" and "keypoint, +ve") bolsters contrastive learning, encouraging the model to generate similar representations under varied augmentations. Table \ref{table:hand_cl_augmentation} compares the model performance with different augmentations. Prior methods \cite{Zimmermann2021,Spurr2021} employed pose-variant augmentations (e.g., rotation and flipping), which can adversely affect the learning by altering the global orientation, and lead to increased errors ("pose") compared to "baseline". Conversely, color-variant, location-variant and their combination provide an improvement over the baseline, showing these augmentations are helpful.

\begin{wraptable}{r}{0.57\textwidth}
\small
\vspace{-15pt}
\centering
\caption{\small \textbf{Ablation of different modules on Hand subnetwork. Results are trained and evaluated on FreiHAND.}}%
\vspace{-5pt}
\resizebox{.57\textwidth}{!}{
\begin{tabular}{@{}lrrrrr@{}}
\toprule
                                & \multicolumn{1}{l}{Supervision} & \multicolumn{1}{l}{PA-$\downarrow$} & \multicolumn{1}{l}{MPJPE$\downarrow$} & \multicolumn{1}{l}{PA-$\downarrow$} & \multicolumn{1}{l}{PVE$\downarrow$} \\ \midrule
Base (R50)         &       & 8.06                         & 16.78                     & 7.85                       & 16.71                   \\
Base (R50) + Strongaug        &       & 8.47                         & 17.01                     & 8.11                       & 16.17                   \\
Base (DR54)        &         & 7.8                          & 15.57                     & 7.67                       & 15.72                   \\

Base (DR54) & $L_{KS}$ & 7.68                         & 15.8                      & 7.62                       & 16.29                   \\
PPP \cite{Moon2022} & $L_{KS}$     & 7.65                         & 15.93                     & 7.56                       & 16.37                   \\
LF & $L_{KS}$   & 7.52                & 15.84            & 7.56              & 16.15          \\
joints & $L_{KS}$           & 7.86                         & 15.92                     & 7.75                       & 16.24                   \\
\midrule
LF (all)  & $L_{KS}$ & 7.49                &15.51            & 7.46
              & 15.59          \\
\midrule
LF (all) + $L_{con}$  & $L_{KS}$ & 7.48                &15.01            & 7.32
              & 15.29          \\
LF (all) + $L_{con}$, +ve  & $L_{KS}$ & 7.42                &14.88            & 7.16
              & 14.57          \\
\midrule

LF (all)  & $L_{KS}$, $L_{segm}$  & 7.44                & 14.92            & 7.58             & 15.30         \\
LF (all)  & $L_{KS}$, $L_{segm}$, $L_{proj}$ & 7.36                & \textbf{14.38}            & 7.53              & 15.05          \\
\textbf{LF (all)+ $L_{con}$, +ve}  & \textbf{$L_{KS}$, $L_{segm}$, $L_{proj}$} & \textbf{7.33}                & 14.59            & \textbf{7.02}              & \textbf{14.11}          \\
\bottomrule
\label{table:hand_ablation}
\end{tabular}}
\vspace{-20pt}
\end{wraptable}



\textbf{Location features.}
Table \ref{table:hand_ablation} shows the ablation of different modules on the Hand subnetwork (ablation for the Body subnetwork is in Table \ref{table:body_ablation} in Appendix).
The baseline model is trained that randomly augments images with a scale factor of 0.2 and bounding box jitter of 0.2. We observe that training using \emph{strongaug} with a larger scale and jitter factor harms the baseline performance. This is likely due to a domain shift.
Hand4Whole \cite{Moon2022} employs sampled features from positional pose-guided pooling (PPP) to predict pose parameters while shape and camera parameters only utilize backbone features. Our method focuses on explicitly learning the location and part silhouettes, utilizing sparse and dense supervision methods. This proves advantageous as the location information ("LF") improve the performance of pose and shape estimations, with the reduced joint and vertex errors of the regressed mesh. Moreover, we find that using location features "LF (all)" for predicting shape and camera parameters is also beneficial.

\textbf{Pixel alignment.}
Tables \ref{table:hand_ablation} also shows that incorporating differential rendering and using projected segmentation loss ($L_{proj}$) for the mesh in \Name helps to achieve lower PVE and MPJPE errors.
It facilitates the learning of more precise body model and camera parameters to improve the alignment between the rendered 3D model and 2D image. Notably, metrics such as PVE and MPJPE errors is calculated after root alignment and may not sufficiently reflect the quality of mesh projection onto the image space. \huien{To offer a more precise analysis, we evaluate the discrepancies between the projected 2D vertices of the ground-truth and projected meshes. More quantitative and qualitative comparisons can be found in Appendix \ref{sec:qualitative_pixel_alignment}. }

\vspace{-10pt}
\section{Conclusion}

In this paper, we introduce a new framework \Name to advance the field of whole-body pose and shape estimation. It enhances the whole-body pipeline by learning more precise localization for part crops while ensuring that part subnetworks are robust enough to handle suboptimal part crops and produce reliable outputs. It achieves this goal with three innovations: accurate subject localization by explicitly learning both sparse and dense predictions of the subject, robust feature extraction with supervised contrastive learning, and accurate pixel alignment of outputs with differentiable rendering. \huien{ Nevertheless, it is important to acknowledge that there are instances in which our framework exhibits limitations, such as (1) inaccurate beta estimation due to out-of-distribution data (children), (2) challenges posed by severe object-occlusion, (3) difficulties arising from person-person occlusion, and (4) the potential for prediction errors in multi-person scenarios, as exemplified by the cases detailed in Appendix \ref{sec:failure_cases}. These challenges represent important avenues for future refinement of our approach.}

\huien{
There are several potential avenues for future research. First, the current approach does not deliberately select negative samples during training. Future work could explore if hard mining by intentionally selecting similar poses in a batch could enhance learning. Second, the careful selection of augmentations is essential. While augmentations that modify the global orientation, such as flipping and rotation, have proven detrimental and are not employed, the effects of individual augmentations and their combinations are not examined. Future research could explore the potential for automatically determining the optimal selection of augmentations to achieve improved performance. Additionally, simplifying the complex framework without sacrificing performance is a beneficial direction for future work. Lastly, considering that videos are a prevalent input format, the integration of video-based estimation can contribute to bolstering model robustness can enhance model robustness, alleviate depth ambiguity, and improve temporal consistency.}

\section*{Acknowledgements} This research/project is supported by the National Research Foundation, Singapore under its AI Singapore Programme (AISG Award No: AISG3-PhD-2023-08-049T). This study is also supported by the Ministry of Education, Singapore, under its MOE AcRF Tier 2 (MOE-T2EP20221-0012), NTU NAP, and under the RIE2020 Industry Alignment Fund – Industry Collaboration Projects (IAF-ICP) Funding Initiative, as well as cash and in-kind contribution from the industry partner(s). We sincerely thank the anonymous reviewers for their valuable comments on this paper.

\bibliography{neurips_2020}
\bibliographystyle{neurips_2020}

\newpage
\appendix
\section*{Overview}

This supplementary material presents more details and additional results not included in the main paper due to the page limitation. The list of items included are:
\begin{itemize}[leftmargin=*,topsep=0pt]
    
    \item Description of augmentation settings for robustness benchmarking in Section \ref{sec:augmentation-robustess}  
    
    \item More experiment setup and details in Section \ref{sec:experiment-setup-extras}  

    \item Comparison with PyMAF-X in Section \ref{sec:pymafx_comparison}

    \item Analysis of the subpar performance on AGORA test set in Section \ref{sec:failure_analysis_agora} 

    \item Ablation of different modules on the Body subnetwork in Section \ref{sec:ablation-body-subnetwork}

    \item \huien{Quantitative and qualitative and comparisons for pixel alignment in Section \ref{sec:qualitative_pixel_alignment}} 

    \item \huien{Examples of failure cases in Section \ref{sec:failure_cases}}

    \item \huien{Analysis of embedding similarity in Section \ref{sec:contrastive_features}}

    \item Discussion on pose (rotation) versus keypoint representation in Section \ref{sec:rotation_vs_joints} 

    \item \huien{Extra comparisons against SOTA body networks in Section \ref{sec:sota_body_comparison}}

    \item \huien{Training and inference time in Section \ref{sec:runtime}}
    
    \item \huien{Accuracy of derived part bounding boxes in Section \ref{sec:pred_bbox_accuracy}}

    \item Qualitative comparisons of \Name's Hand, Face and Body subnetworks under augmentations in Section \ref{sec:qualitative_evaluation_aug} 

    \item Quantitative and qualitative comparisons of \Name's wholebody model in Section \ref{sec:evaluation_wholebody}

\end{itemize}



\section{Augmentation Settings for Robustness Benchmarking}
\label{sec:augmentation-robustess}

In the selection of augmentations, we opted for a set of ten commonly encountered augmentations that could be benchmarked in a controlled setting. We also ensure that the selected values for manipulation fall within a realistic range. We used the following augmentations:
\begin{enumerate}[leftmargin=*]
\item Vertical translation: We shifted the image by factors relative to the image size. For instance, a +0.1 shift corresponds to a 10\% upward movement, while a -0.1 shift represents a 10\% downward movement. Our boundaries were set at $\pm0.3$ to ensure that majority of the subject remains visible within the image frame.
\item Horizontal translation: We manipulated the image by factors relative to the image size. A shift of +0.1 denotes a 10\% move to the right, while -0.1 indicates a 10\% shift to the left. We imposed a $\pm0.3$ limit to keep the majority of the subject within the image.
\item Scale: We adjusted the person's crop using factors relative to the bounding box size. For example, a factor of +0.1 leads to a 10\% size reduction, resulting in a tighter crop, while a -0.1 factor enlarges the crop size by 10\%. A $\pm0.5$ boundary was set to maintain visibility of the majority of the person within the image.
\item Low Resolution: The resolution of the cropped image was modified by factors related to the image size. A 2.0 factor signifies that the image was downsampled to half its original size before being upsampled back, reducing the resolution by a factor of 2.0.
\item Rotation: The image was manipulated by various rotations up to degrees of $\pm60$.
\item Hue: The image hue was altered by converting the image to HSV format, cyclically shifting intensities in the hue channel (H), and converting back to the original image mode. Hue adjustments were limited to $\pm0.5$.
\item Sharpness: Sharpness was controlled by introducing an enhancement factor. A factor of -1.0 leads to a blurred image, while +1.0 results in a sharpened image, with 0.0 leaving the image unaltered. This effect is achieved by blending the source image with the degraded mean image.
\item Grayness: The degree of grayness was adjusted by introducing an enhancement factor. A factor of -1.0 results in a completely grayed image, while +1.0 leads to a whitened image, with 0.0 leaving the image unaltered. This effect is achieved by blending the source image with its gray counterpart. The limit was set to $\pm0.5$, as the subject becomes unidentifiable at extremes of $\pm1.0$. 
\item Contrast: This was controlled by introducing an enhancement factor. A factor of -1.0 leads to a completely grayed image, while +1.0 results in a whitened image, with 0.0 leaving the image unaltered. This effect is achieved by blending the source image with the degraded mean image. The limit was set to $\pm0.5$, as the subject becomes unidentifiable at extremes of $\pm1.0$.
\item Brightness: The brightness of the image was adjusted by introducing an enhancement factor. A factor of -1.0 results in a black image, while +1.0 leads to a white image, with 0.0 leaving the image unaltered. This effect is achieved by blending the source image with the degraded black image. The limit was set to $\pm0.5$, as the subject becomes unidentifiable at extremes of $\pm1.0$.
\end{enumerate}

\section{More Experiment Setup}
\label{sec:experiment-setup-extras}

This section includes extra description of each submodule and implementation details.

\textbf{Body subnetwork.}
The body image is downsampled from the original image to reduce the computational cost, resulting in $I_{b}$ \( \in R^{3\times256\times256}\). The Body subnetwork outputs 3D body joint rotations $\theta_{b}$ \( \in R^{21\times3}\), global orientation $\theta_{bg}$ \( \in R^{3}\), shape parameters $\beta_{b}$ \( \in R^{10}\), camera parameters $\pi_{b}$ \( \in R^{3}\), and whole-body joints \( K \in R^{137 \times 3}\). Hand and face bounding boxes are then derived from the face and hand keypoints. Width and height are determined from the x-y range of the keypoints, and the center is the aggregated mean of the keypoints. High resolution crops are used for hand and face inputs following ExPose and PIXIE. In line with ExPose \cite{Choutas2020} and PIXIE \cite{Feng2019}, hand and face input images are obtained from high resolution crops to utilize the information available from the original image instead of the downsampled image.

\textbf{Hand subnetwork.}
After obtaining the cropped hand images $I_{h}$ \( \in R^{3\times256\times256}\), the left hand images are flipped to match the orientation of the right hands before being input to the Hand subnetwork. After predicting the 3D finger rotations $\theta_{h}$ \( \in R^{15\times3}\), the outputs of the flipped left hands are reverted to their original orientation. The 3D finger rotations of the left and right hands are denoted as $\theta_{rh}$ and $\theta_{lh}$ respectively. When training the full version on hand datasets, we also output the global orientation $\theta_{hg}$ \( \in R^{3}\), shape $\beta_{h}$ \( \in R^{10}\) and camera $\pi_{h}$ \( \in R^{3}\). However, these branches are discarded during whole-body estimation and training.

\textbf{Face subnetwork.}
\label{face_subnetwork}
This subnetwork generates the 3D jaw rotation $\theta_{f}$ \( \in R^{3}\) and expression $\psi_{f}$ \( \in R^{10}\) from the cropped face image $I_{f}$ \( \in R^{3\times256\times256}\). When training the full version on face datasets, additional outputs include the global orientation $\theta_{fg}$ \( \in R^{3}\), shape $\beta_{f}$ \( \in R^{50}\), expression $\psi_{f}$ \( \in R^{50}\) and camera $\pi_{f}$ \( \in R^{3}\). These branches are also discarded during whole-body estimation and training.

\textbf{Implementation details.} 
The training and evaluation of our model builds upon the MMHuman3D framework \cite{mmhuman3d}. For model initialization, we pre-train the ResNet backbone on the MSCOCO 2D whole-body human pose dataset. During training, we use the Adam optimizer with a mini-batch size of 32 and apply data augmentations, e.g., scaling, rotation, random horizontal flip, and color jittering. The initial learning rate is set to $10\mathrm{e}{-4}$, decayed by a factor of 10 at the later epoch. 
We use the SMPL, MANO, FLAME and SMPL-X body models for the training of body, hand, face and wholebody respectively. 
Further details will be provided in our code.


\section{Comparison with PyMAF-X}
\label{sec:pymafx_comparison}
Below we provide detailed discussions and comparisons with PyMAF-X \cite{Zhang2022}. 
\begin{enumerate}
\item Acquisition of part bounding boxes: PyMAF-X relies on an off-the-shelf whole-body pose estimation model (OpenPifpaf) to obtain whole body 2D keypoints of the person in the image, from which part
crops are derived. During the EHF evaluation, PyMAF-X employs hand and face
bounding boxes derived from OpenPose keypoints. In contrast, our method and other works (ExPose \cite{Choutas2020}, PIXIE \cite{Feng2019}, Hand4Whole \cite{Moon2022} and OS-X \cite{Lin2023}.) encompass a self-integrated module designed to extract hand and face bounding boxes directly from the image. 
\item Operational efficiency: Openpifpaf imposes extra computation during inference, making PyMAF-X
less efficient than our method. Please refer to Section \ref{sec:runtime} in Appendix.
\item Network architecture: Due to the diverse backbone and dataset combinations utilized, it is
challenging for us to make whole-body network comparisons. In Table \ref{table:freihand}, we focus on
contrasting RoboSMPLX’s Hand subnetwork with PyMAF’s Hand subnetwork. Both networks are
trained and evaluated on the same backbone and dataset, FreiHAND. In this context, our method
surpasses PyMAF.
\item Performance: On the EHF metrics, our performance lags behind PyMAF-X. This could potentially
arise from variations in the training datasets employed. While the training pipeline of the body
network for PyMAF-X has been disclosed, the training specifics for hands and face and the
methodology to integrate hand, face, and body module PyMAF-X, remains undisclosed. We intend to
replicate with similar training datasets in the future.
\end{enumerate}

\section{Analysis of performance on AGORA test set}
\label{sec:failure_analysis_agora}

\begin{figure}[H]
    \centering
    \subfigure{
    \includegraphics[width=0.8\linewidth ,keepaspectratio]{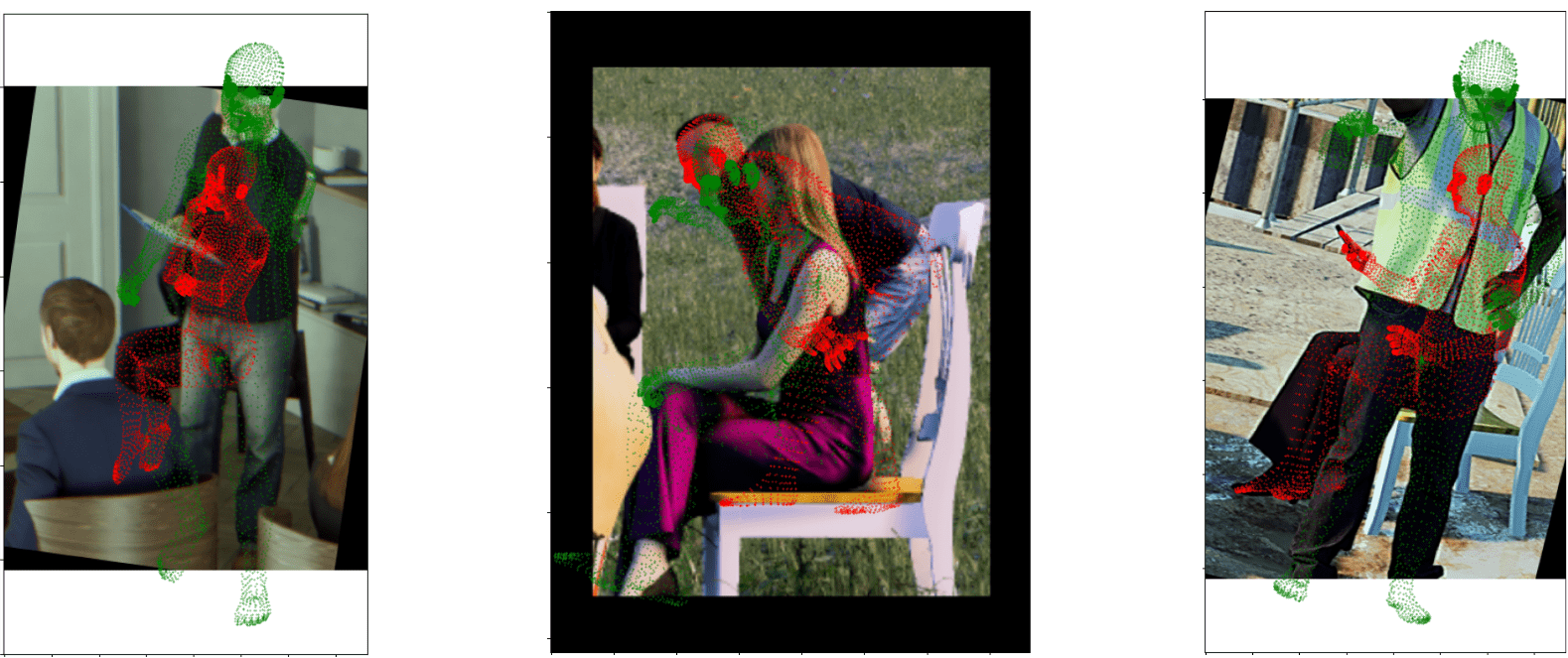}
    }
    \vspace{-10pt}
    \caption{\small \textbf{Visualisation of samples with high errors at train time.} Red vertices indicates the target person while green vertices are the model's predictions.}
    \vspace{-20pt}
    \label{figure:agora_failure_cases}
\end{figure}

Figure \ref{figure:agora_failure_cases} visualise samples with significant errors during training. AGORA contains extensive person-to-person occlusion, frequently leading to substantial overlap between the target individual (marked with red vertices) and another person. In cases that experienced large errors, the model often incorrectly identified the target individual as the person situated in the forefront (model predictions marked with green vertices), thereby introducing instability throughout the training process due to the model's challenge in accurately discerning the intended subject.

We also added qualitative comparisons of \Name under varying scales and alignments as shown in Figures \ref{figure:agora_aug_vis}. We demonstrate that \Name produces better pixel alignment of the body, and more accurate hand and face predictions where the target person has been accurately identified.

\begin{figure}[H]
    \centering
    \subfigure{
    \includegraphics[width=0.9\linewidth,keepaspectratio]{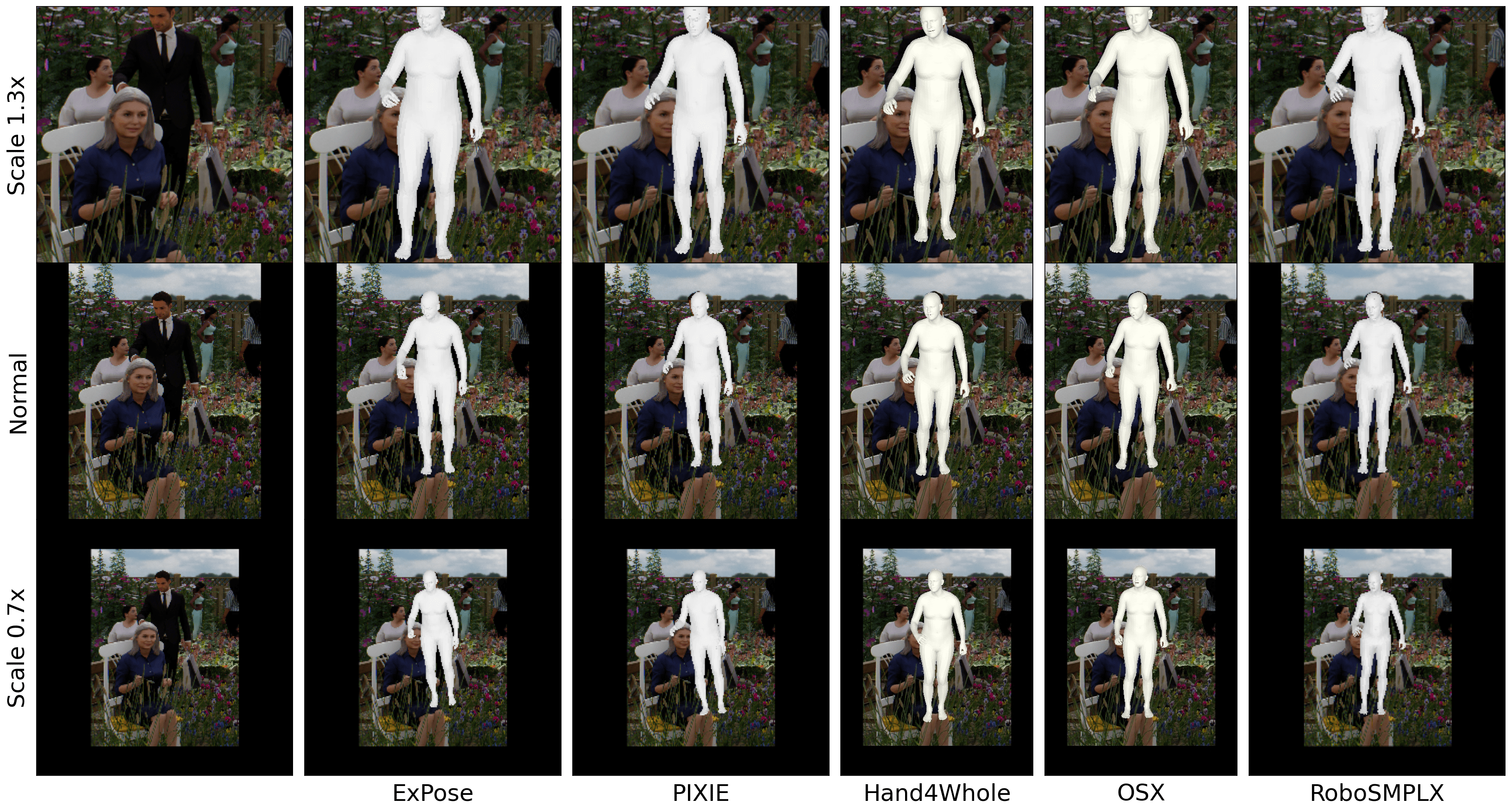}
    }
    \caption{\small \textbf{ Visualisation of Expose \cite{Pavlakos2019}, PIXIE \cite{Feng2019}, Hand4Whole \cite{Moon2022}, OS-X \cite{Lin2023} and \Name under different scales and alignment on AGORA validation set.}}
\end{figure}
\begin{figure}[H]\ContinuedFloat
    \subfigure{
    \includegraphics[width=0.9\linewidth ,keepaspectratio]{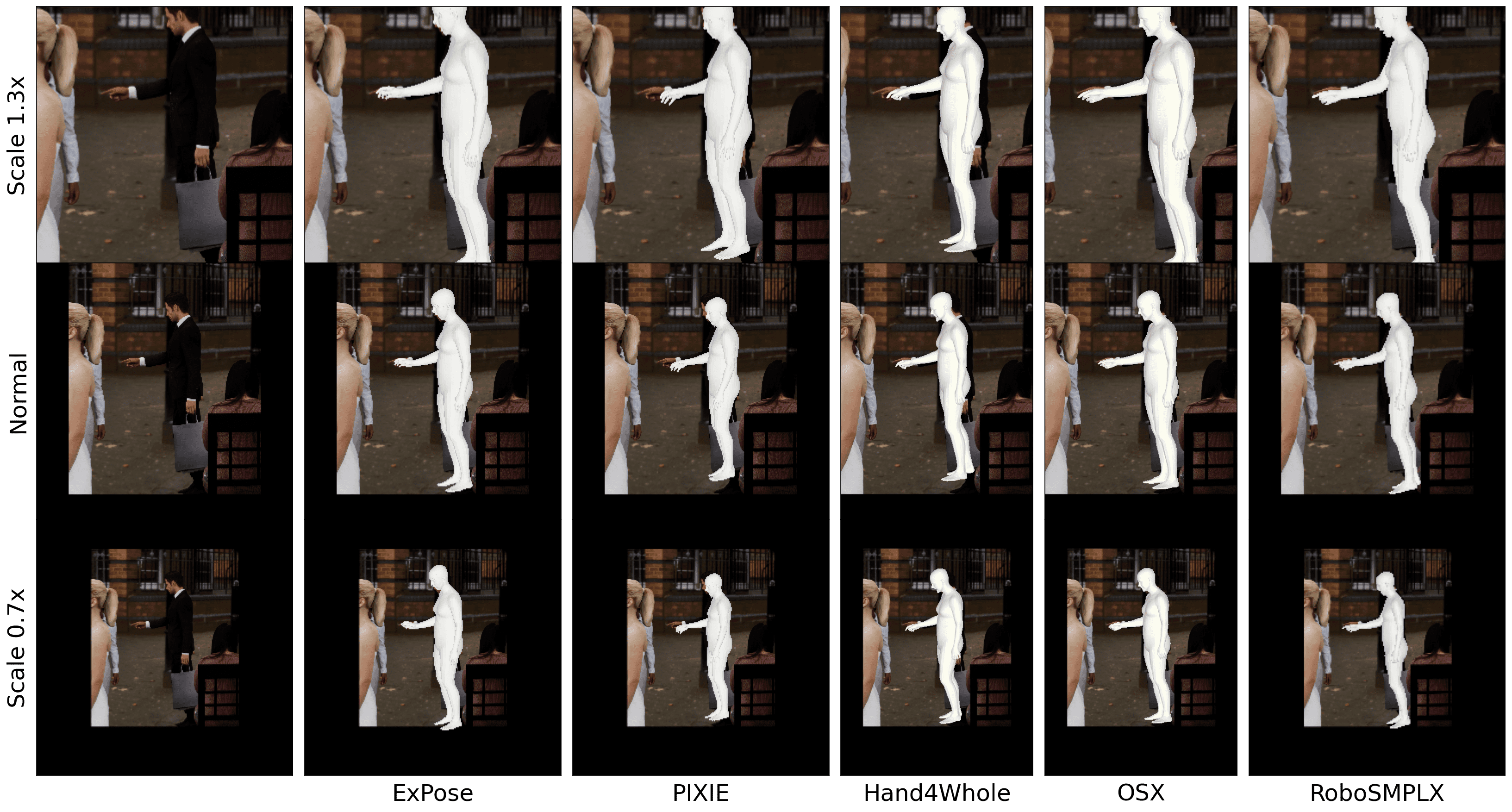}
    }
    \subfigure{
    \includegraphics[width=0.9\linewidth ,keepaspectratio]{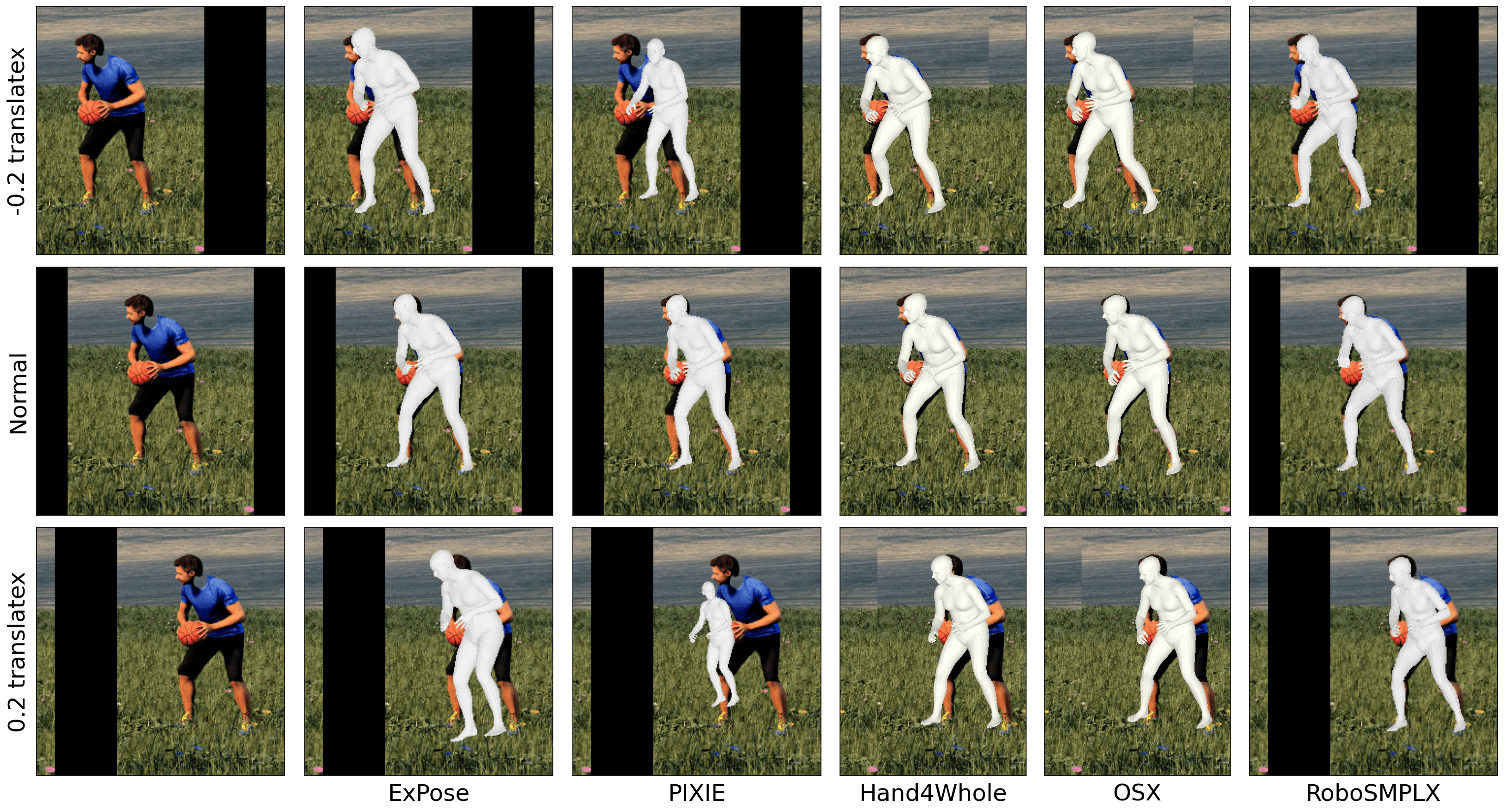}
    }
    \subfigure{
    \includegraphics[width=0.9\linewidth ,keepaspectratio]{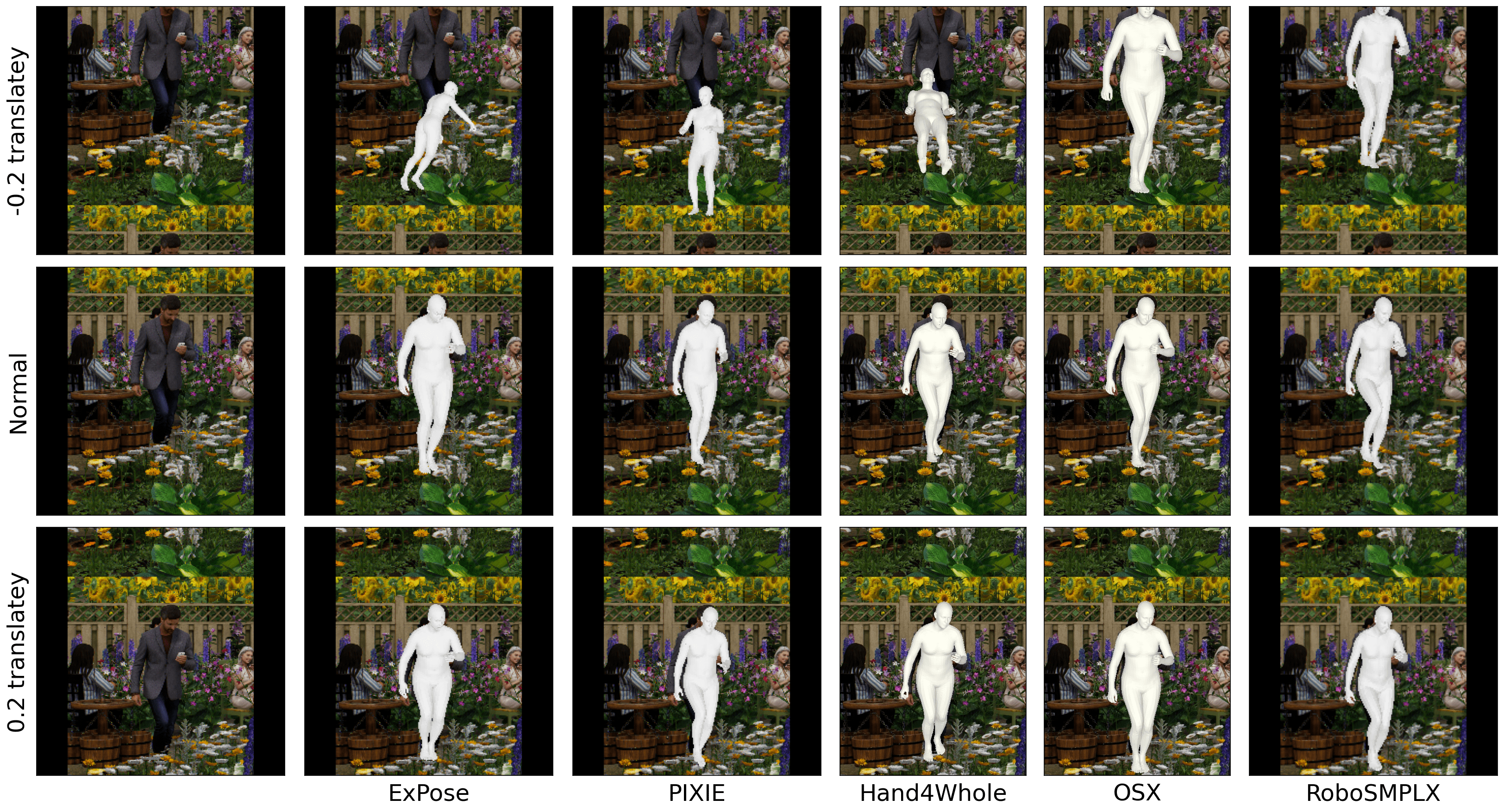}
    }
    \caption{\small \textbf{Visualisation of Expose \cite{Pavlakos2019}, PIXIE \cite{Feng2019}, Hand4Whole \cite{Moon2022}, OS-X \cite{Lin2023} and \Name under different scales and alignment on AGORA validation set (cont.).}}
    \label{figure:agora_aug_vis}
\end{figure}

\section{Ablation on Body Subnetwork}
 Table \ref{table:body_ablation} shows the ablation of different modules on the Body subnetwork. The conclusions derived from the Hand ablation study (Table \ref{table:hand_ablation}) extends to the Body subnetwork as well.

\label{sec:ablation-body-subnetwork}
\begin{table}[ht]
\centering
\vspace{-15pt}
\tiny
\caption{\small \textbf{Ablation of different modules on Body subnetwork. Results are trained on EFT-COCO and tested on 3DPW test set.}}%
\begin{tabular}{@{}lrrrrr@{}}
\toprule 
                                & \multicolumn{1}{l}{loss} & \multicolumn{1}{l}{representation}  
                                & \multicolumn{1}{l}{PA-} & \multicolumn{1}{l}{MPJPE} \\ \midrule
Baseline (HMR)        &-&-    & 60.8                          & 96.2                                        \\
LF (all)        &-&-           & 56.7                         & 105.7                                               \\
LF (all), $L_{con}$  & L1 & pose  & 55.9                        & 90.9                                         \\

LF (all), $L_{con}$ & MSE & pose & 58.5                        & 93.9                                         \\
LF (all), $L_{con}$ & SmoothL1 & pose  & 56.6                        & 92.5                                       \\
LF (all), $L_{con}$  & L1 & pose(rot6d)  & 58.9                        & 95.0                                          \\
LF (all), $L_{con}$  & L1 & pose + go & 76.8                        & 118.9                                         \\

\textbf{LF (all),  $L_{con}$, +ve}   & L1 & keypoints & \textbf{55.4}                        & \textbf{90.56}                                     \\
\bottomrule
\label{table:body_ablation}
\end{tabular}
\vspace{-15pt}
\end{table}


\section{Qualitative and quantitative comparisons for pixel alignment}
\label{sec:qualitative_pixel_alignment}
\begin{figure}[H]
    \centering
    \subfigure{
    \includegraphics[width=0.55\linewidth,keepaspectratio]{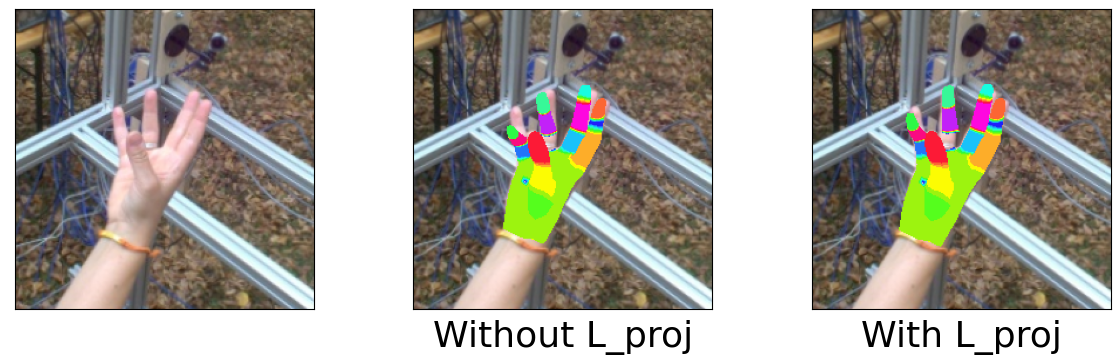}
    }
    \vspace{-11pt}
    \subfigure{
    \includegraphics[width=0.55\linewidth,keepaspectratio]{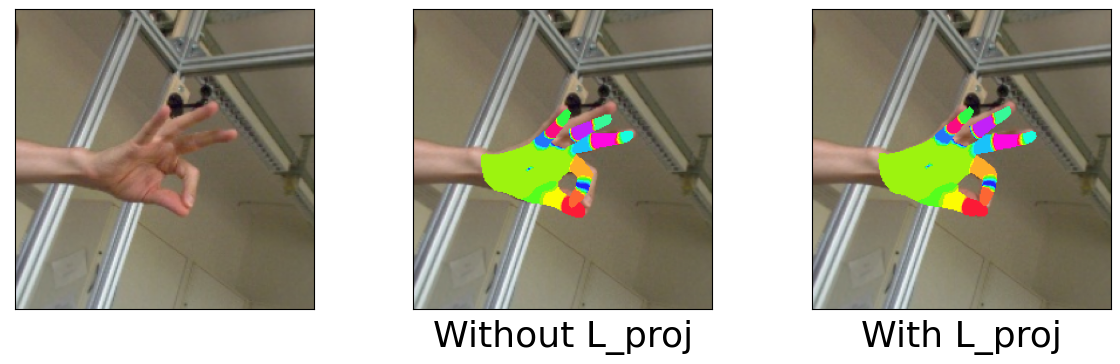}
    }
    \vspace{-11pt}
    \subfigure{
    \includegraphics[width=0.55\linewidth,keepaspectratio]{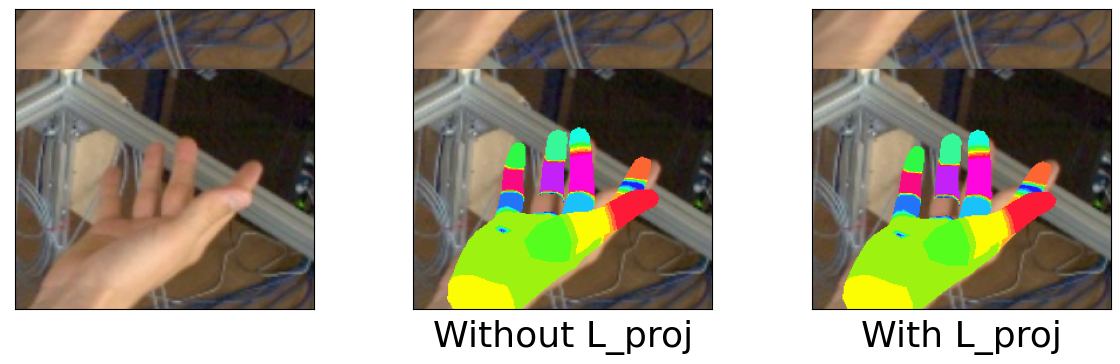}
    }
    \vspace{-11pt}
    \subfigure{
    \includegraphics[width=0.55\linewidth,keepaspectratio]{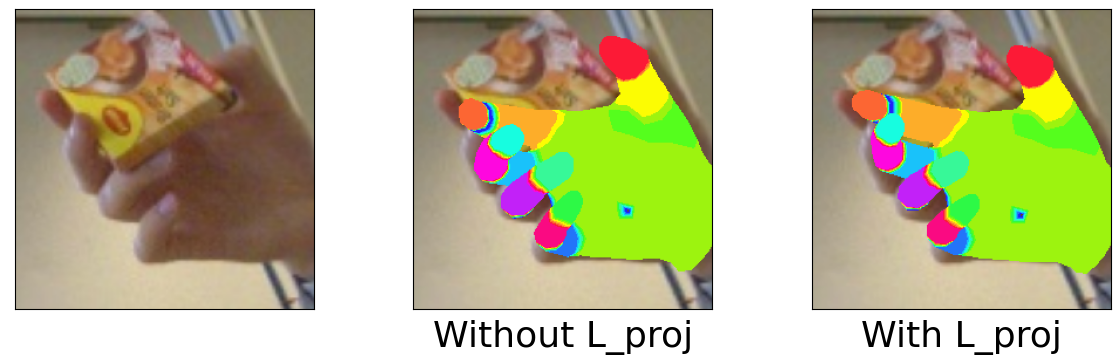}
    }
    \caption{\small \textbf{(C) Visualisation from training with and without $L_{proj}$.}}
    \label{figure:pixel_alignment}
\end{figure}

Prevailing metrics such as Per Vertex Error (PVE) and Mean Per Joint Position Error (MPJPE) do not incorporate alignment measurement in their evaluation. Before these metrics are computed, the mesh undergoes root alignment, but this process does not necessarily reflect the level of alignment accuracy when the mesh is reprojected back into the image space.

Moreover, for pose and shape estimation methods, the absence of ground-truth camera parameters implies that there is no direct supervision for these parameters. Camera parameters are, instead, often weakly supervised through the supervision of projected keypoints (derived from regressed joints of the mesh and predicted camera parameters) and the ground-truth 2D joints by ensuring their alignment. This only provides a sparse supervision. To enhance better learning of camera, pose and shape parameters, pixel alignment strategy is introduced, which ensures denser supervision. 

Presently, there's an absence of a metric tailored to gauge the degree of pixel alignment of a mesh in this context. We included qualitative examples of training with and without $L_{proj}$, and demonstrate that the projection of vertices results in better pixel alignment (Figure \ref{figure:pixel_alignment}). 





\begin{minipage}{\textwidth}
\begin{minipage}[b]{0.49\textwidth}
\centering
\includegraphics[width=0.5\linewidth,keepaspectratio]{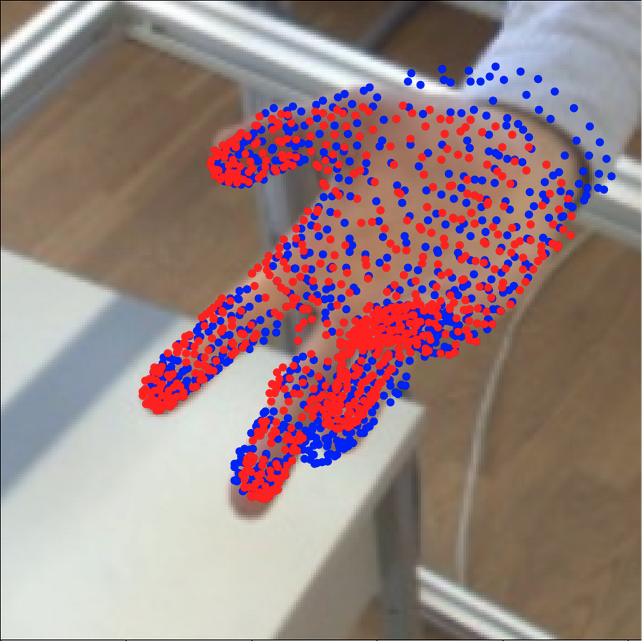}

\captionof{figure}{\small \textbf{Projected Vertex Errors is measured as distance between projected ground-truth (red) and predicted (blue) vertices in image space.}}%
\label{figure:pixel_alignment_quantitative}
\end{minipage}
\hfill
\begin{minipage}[b]{0.49\textwidth}
\centering
\tiny
\begin{tabular}{@{}ll@{}}
\toprule
Method                                                & \multicolumn{1}{c}{Projected Vertex Errors $\downarrow$ } \\ \midrule
HMR (no PA)                       & 11.796                                      \\
HMR  + PA (vertex)   & 11.211                                      \\
HMR + PA (part-seg) & \textbf{10.298}                                      \\ \bottomrule
\end{tabular}

\captionof{table}{\small \textbf{Results of Projected Vertex Errors under different Part Alignment (PA)}}%
\label{table:pixel_alignment_quantitative}
\end{minipage}
\end{minipage}


To provide quantitative analysis, we measure errors between the projected 2D vertices of ground-truth and projected meshes (Figure \ref{figure:pixel_alignment_quantitative}). From Table \ref{table:pixel_alignment_quantitative}, it is evident that omitting the pixel alignment module leads to suboptimal outcomes. In contrast, our pixel alignment strategy, leveraging rendered segmentation maps, showcases better performance than using vertex loss as supervision.

\section{Failure cases}
\label{sec:failure_cases}
\begin{figure}[H]
\vspace{-10pt}
    \centering
    \subfigure{
    \includegraphics[width=0.8\linewidth,keepaspectratio]{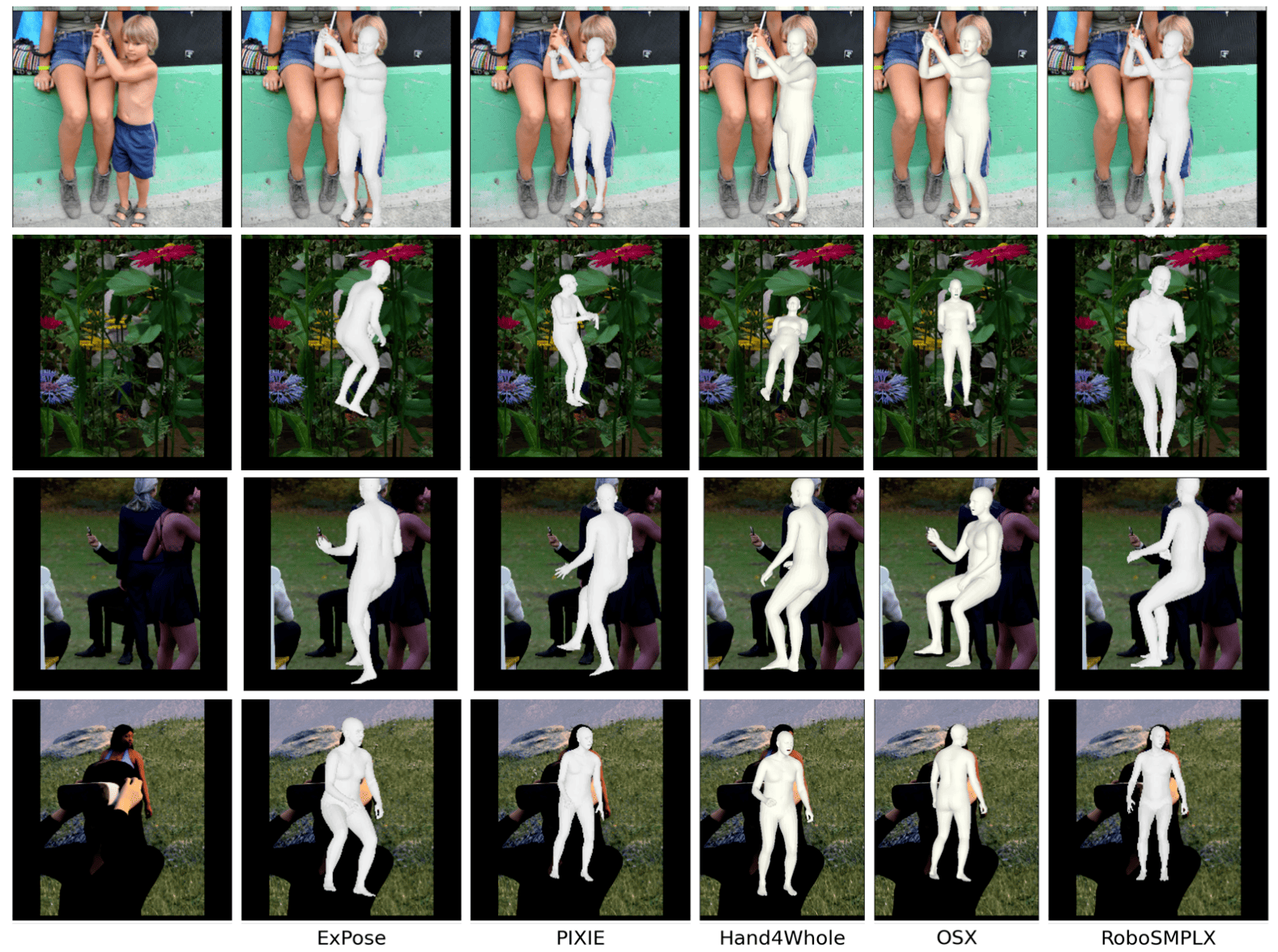}
    }\vspace{-10pt}
    \caption{\small \textbf{Examples of failure cases. (1) Inaccurate beta estimation due to out-of-distribution data (children) (2) Severe object-occlusion (3) Person-person occlusion (4) Prediction for wrong person in multi-person scenarios.}}
    \label{figure:failure_cases}
\end{figure}

\section{Embedding similarity}
\label{sec:contrastive_features}
Our use of the contrastive module is motivated by the need to constrain/maintain the same pose feature for different augmentations, to avoid domain shift caused by strong augmentation alone. The experiments show that the use of strong augmentation alone for training can lead to performance deterioration, while combining it with the contrastive loss consistently results in minimal errors (Table \ref{table:hand_ablation}).

\begin{table}[!hbtp]
\tiny
\centering
\caption{\small \textbf{Ablation of CFE module on Hand Subnetwork. (This excludes Localization and Pixel Alignment Module). Results are trained and evaluated on FreiHAND.}}%
\centering\
\begin{tabular}{@{}llllll@{}}
\toprule
Method                                                                                       & PA-MPJPE $\downarrow$ & MPJPE $\downarrow$ & PA-PVE $\downarrow$ & PVE $\downarrow$    & Pose embedding distance $\downarrow$ \\ \midrule
Model 0: HMR                                                                   & 8.06     & 16.78 & 7.85   & 16.71  & 0.132           \\
Model 1: HMR + Strongaug                                               & 8.47     & 17.01 & 8.11   & 16.17  & 0.138           \\
Model 2: HMR + Strongaug + CL & \textbf{7.79}    & \textbf{15.68} & \textbf{7.41}  & \textbf{15.27} & 0.101           \\ \bottomrule
\end{tabular}
\label{table:embedding_similarity_cfe}
\end{table}

To illustrate this further, we delved into a visualization of the pose similarity for augmented samples. The findings reveal that augmented samples are perceived as dissimilar in both Model 0 and Model 1 (Table \ref{table:embedding_similarity_cfe}). Yet, when examining Model 2, a marked increase in embedding similarity is evident, underscoring the advantage of the contrastive approach.

\section{Discussion of pose versus keypoint representation}

\label{sec:rotation_vs_joints}
\begin{figure}[H]
    \subfigure{
    \includegraphics[width=0.8\linewidth ,keepaspectratio]{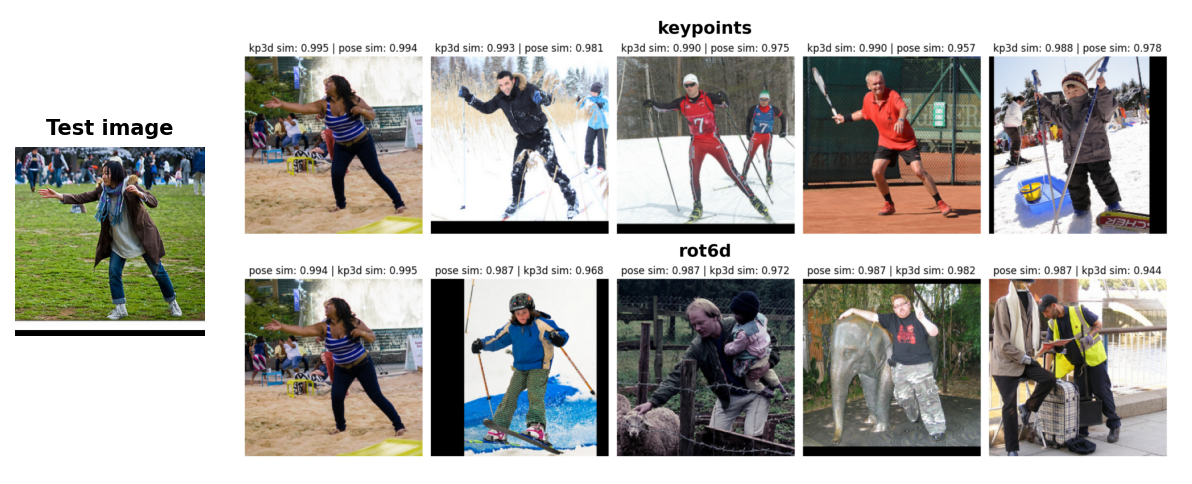}
    }
    \vspace{-15pt}
    \subfigure{
    \includegraphics[width=0.8\linewidth ,keepaspectratio]{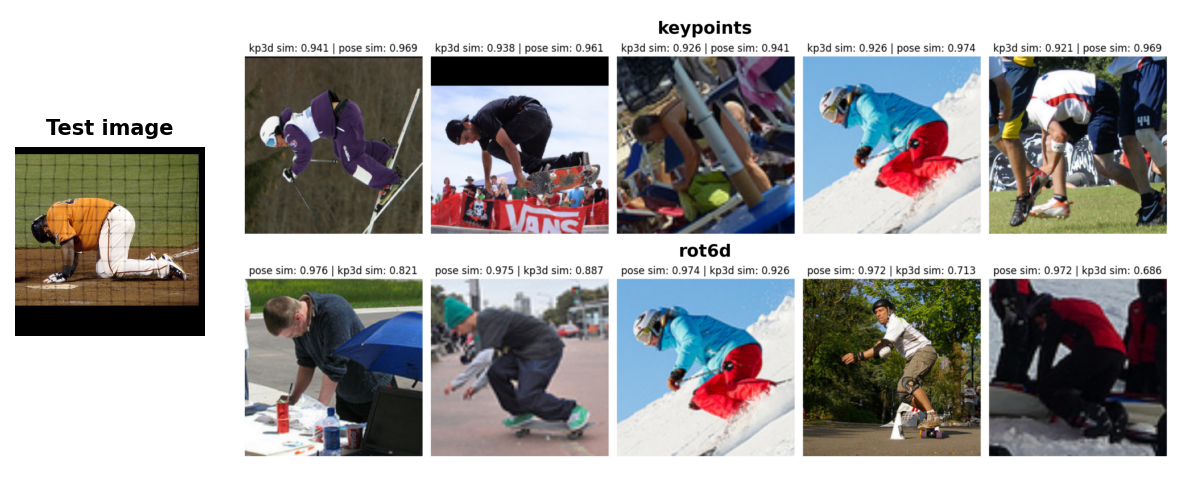}
    }
    \vspace{-15pt}
    \subfigure{
    \includegraphics[width=0.8\linewidth ,keepaspectratio]{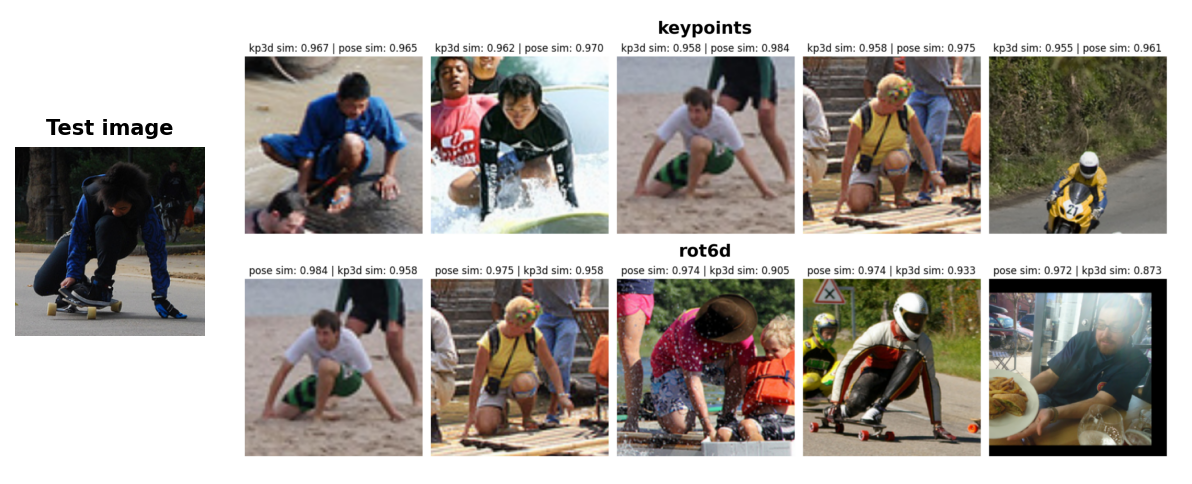}
    }
    \caption{\small \textbf{Comparison of keypoints and pose representations.}}
    \label{figure:pose_vs_keypoints}
\end{figure}%

Figure \ref{figure:pose_vs_keypoints} compares two distinct methods for image retrieval: one based on pose similarity (rot6d representation) and the other based on keypoint similarity. Samples with high keypoint similarity tends to have comparatively high pose similarity.  On the contrary, similar pose representation might have considerably lower joint representation. This could occur due to the accumulation of minor discrepancies in joint rotations which, over time, may result in significant disparities in the keypoints. This results in instances where the overall pose of the retrieved sample is very high to the query sample, but the keypoints may not coincide as accurately. Meanwhile, using keypoint representation would result in samples that have demonstrate improved alignment with the query image, presenting a more accurate correspondence.

This could explain why using regressed keypoints as representation have better performance (Table \ref{table:hand_cl_representation}). Clustering based on keypoint similarity is more effective than pose similarity, as pose representation might be susceptible to minor shifts in joint rotations.

\section{Extra comparisons against SOTA body networks}
\label{sec:sota_body_comparison}

There are many factors affecting training,including, but not limited to, the choice of backbone, datasets employed, and specific protocols executed during evaluation. Specifically, with regards to 3DPW, various protocols—ranging from fine-tuning (3DPW Protocol 1), collective training, to omission during training (3DPW Protocol 2)—have a large influence on 3DPW results in the evaluation process.

In Table \ref{table:body_network_sota_comparison}, we outperform HybrIK when using the same backbone (HRNet-W48) and not fine-tuning on 3DPW (3DPW Protocol 2). Notably, CLIFF incorporated 3DPW within its training datasets. Given that our approach and that of both HybrIK and CLIFF do not utilize identical dataset combinations, a direct comparison becomes inherently challenging.


\begin{table}[!htbp]
\centering
\tiny
\caption{\small Evaluation of HybrIK, CLIFF and our network on 3DPW. Our results are also available in Table 2.}%
\begin{tabular}{@{}lllll@{}}
\toprule
Method & Backbone  & F-T on 3DPW       & PA-MPJPE (3DPW) & MPJPE (3DPW) \\ \midrule
HybrIK & HRNet-W48 & No                & 48.6            & 88.0         \\
HybrIK & HRNet-W48 & Yes               & 41.8            & 71.3         \\
CLIFF  & Res-50    & Trained with 3DPW & 45.7            & 72.0         \\
CLIFF  & HRNet-W48 & Trained with 3DPW & 43.0            & 69.0         \\
Ours   & Resnet-50 & No                & 49.8            & 80.8         \\
Ours   & HRNet-W48 & No                & 48.5            & 80.1         \\ \bottomrule
\label{table:body_network_sota_comparison}
\end{tabular}
\end{table}

We have provided qualitative comparisons of body-only methods under different scale and alignment in Figure \ref{figure:body_aug_vis}. Below, we provide quantitative evaluations of our method with HMR, SPIN and PARE (Table \ref{table:body_network_scale_alignment}). Our method is able to achieve better performance under different scales and alignment.


\begin{table}[!htbp]
\centering
\tiny
\caption{\small  Evaluated on 3DPW (PA-MPJPE/MPJPE) under different scales and alignment. * denote the same dataset combination}%
\begin{tabular}{@{}llllllll@{}}
\toprule
               & Normal             & Transx +0.2x        & Transx -0.2x         & Transy +0.2y       & Transy -0.2y        & Scale 1.3x         & Scale 0.7x          \\ \midrule
HMR            & 67.53/112.34       & 77.31/141.70        & 77.06/ 138.51        & 86.57/ 151.15      & 77.26/148.33        & 68.46/ 117.1       & 75.38/ 124.79       \\
SPIN           & 57.54/94.11        & 70.14/122.56        & 68.67/ 120.04        & 73.08/ 111.33      & 70.64/133.2         & 61.08/ 103.60      & 61.63/ 99.6         \\
PARE  (HR32) * & \textbf{49.3/81.8} & 74.9/139.2          & 77.1/ 141.7          & 59.1/92.3          & 64.2/ 109.7         & 54.7/86.9          & \textbf{50.5/ 83.9} \\
Ours (R50) *   & 49.8/80.8          & \textbf{67.2/117.2} & \textbf{67.72/111.5} & \textbf{56.4/90.0} & \textbf{62.8/105.6} & \textbf{50.2/84.6} & 50.8/ 82.4          \\ \bottomrule
\label{table:body_network_scale_alignment}
\end{tabular}
\end{table}

\section{Training and inference time}
\label{sec:runtime}

Our model was trained utilizing a cluster of 8xTesla V100-SXM2-32GB GPUs. Specific to the training duration, the hand models required approximately one day, whereas the body and face models necessitated two days. The joint training process was completed within a day.

We measure the model size, computation complexity and inference time for different models including ours, as shown in Table \ref{table:runtime}. Although our framework has sophisticated design, it has comparable inference speed as others, validating its efficacy.
\begin{table}[!htbp]
\centering
\tiny
\caption{\small These results are tested on RTX3090. FLOP refers to the total number of floating point operations required for a single forward pass. The higher the FLOPs, the slower the model and hence low throughput. Inference Time is obtained by averaging across 100 runs.}%
\begin{tabular}{@{}llll@{}}
\toprule
                      & Total parameters (M) & GFLOPs         & Inference time (s) \\ \midrule
ExPose                & 26.06                & 21.04          & 0.1330 $\pm$ 0.0050     \\
PIXIE                 & 109.67               & 24.23          & 0.1670 $\pm$ 0.0065     \\
Hand4Whole            & 77.84                & 17.98          & 0.0709 $\pm$ 0.0022     \\
OSX                   & 422.52               & 83.77          & 0.1998 $\pm$ 0.0028     \\
PyMAF-X (gt H/F bbox) & 205.93               & 33.41          & 0.2194 $\pm$ 0.0027     \\
PyMAF-X + OpenPipaf   & 205.93 + 115.0       & 33.41 + 120.52 & 0.2727 $\pm$ 0.0136     \\
RoboSMPLX             & 120.68               & 29.66          & 0.2008 $\pm$ 0.0220     \\ \bottomrule
\label{table:runtime}
\end{tabular}
\end{table}

\section{Quantitative evaluation of predicted bounding box accuracy}
\label{sec:pred_bbox_accuracy}
To assess the precision of predicted part bounding boxes on the EHF test set, we utilized Intersection over Union (IoU) as our evaluation metric (see Figure \ref{figure:iou_calculation}). Our method achieved the highest IoU scores, as demonstrated in Table \ref{table:iou_calculation}. It is important to note that in the OSX implementation, the hand and face features are cropped from the body features rather than directly from the image.

\begin{minipage}{\textwidth}
\begin{minipage}[b]{0.49\textwidth}
\centering
\includegraphics[width=0.4\linewidth,keepaspectratio]{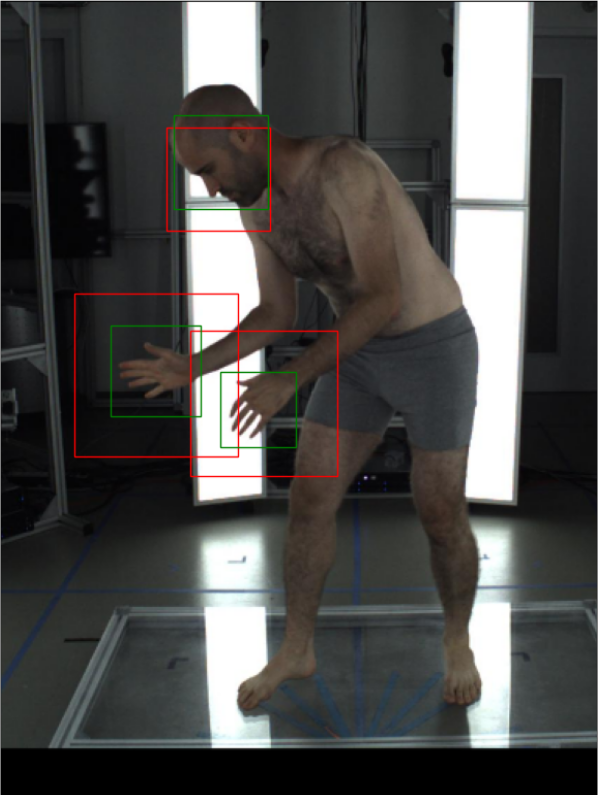}

\captionof{figure}{\small \textbf{Calculation for the Face, LHand and RHand IoU scores for ground-truth (green) and predicted (red) part bounding boxes.}}%
\label{figure:iou_calculation}
\end{minipage}
\hfill
\begin{minipage}[b]{0.49\textwidth}
\centering
\tiny
\begin{tabular}{@{}llll@{}}
\toprule
Method     & Face IoU & LHand IoU & RHand IoU \\ \midrule
ExPose     & 0.61     & 0.23      & 0.31      \\
PIXIE      & 0.66     & 0.34      & 0.36      \\
Hand4Whole & 0.75     & 0.41      & 0.45      \\
OSX        & 0.70     & 0.38      & 0.41      \\
RoboSMPLX  & 0.86     & 0.52      & 0.55      \\ \bottomrule
\end{tabular}

\captionof{table}{\small \textbf{Results for IoU of the predicted part bounding boxes on the EHF test set.}}%
\label{table:iou_calculation}
\end{minipage}
\end{minipage}

\clearpage
\section{Qualitative comparisons for different models under augmentation}
\label{sec:qualitative_evaluation_aug}

\begin{figure}[H]
    \subfigure{
    \includegraphics[width=0.9\linewidth ,keepaspectratio]{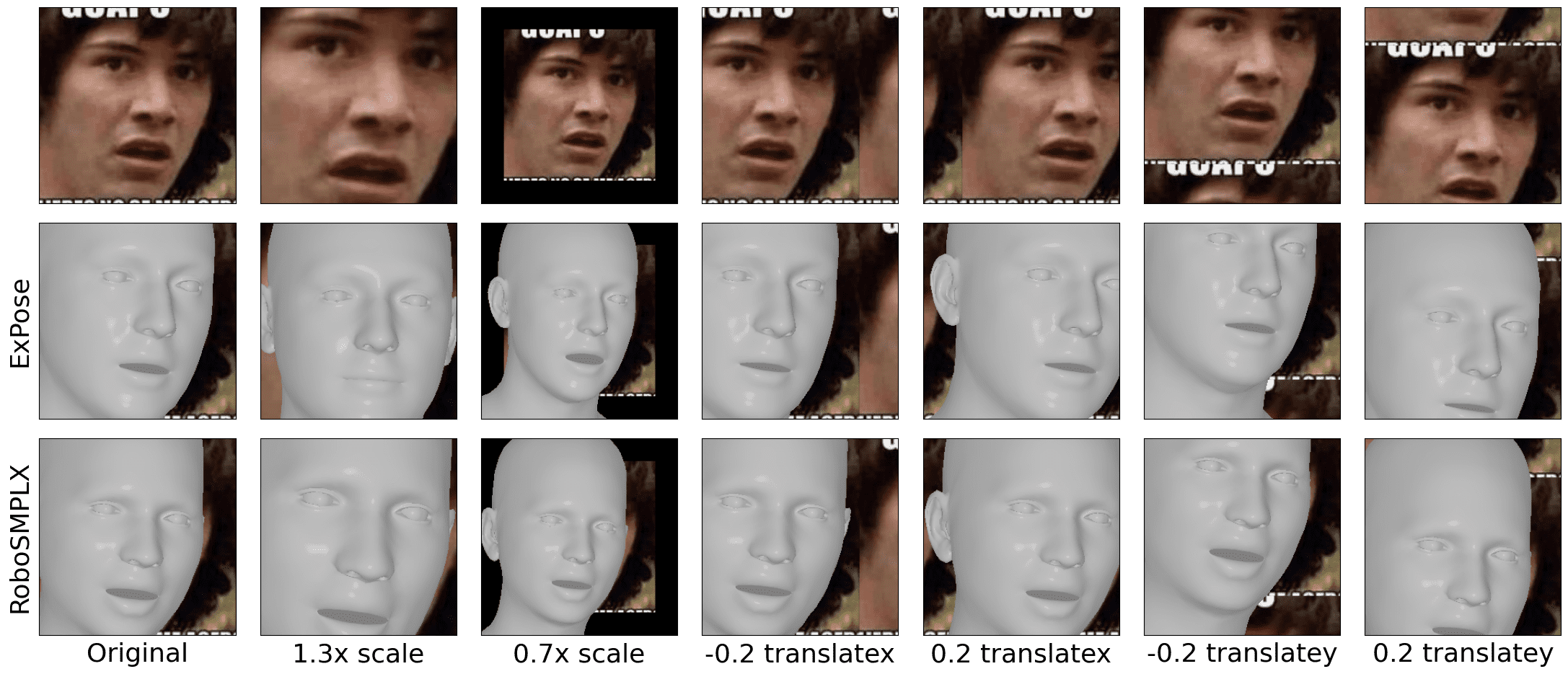}
    }
    \subfigure{
    \includegraphics[width=0.9\linewidth ,keepaspectratio]{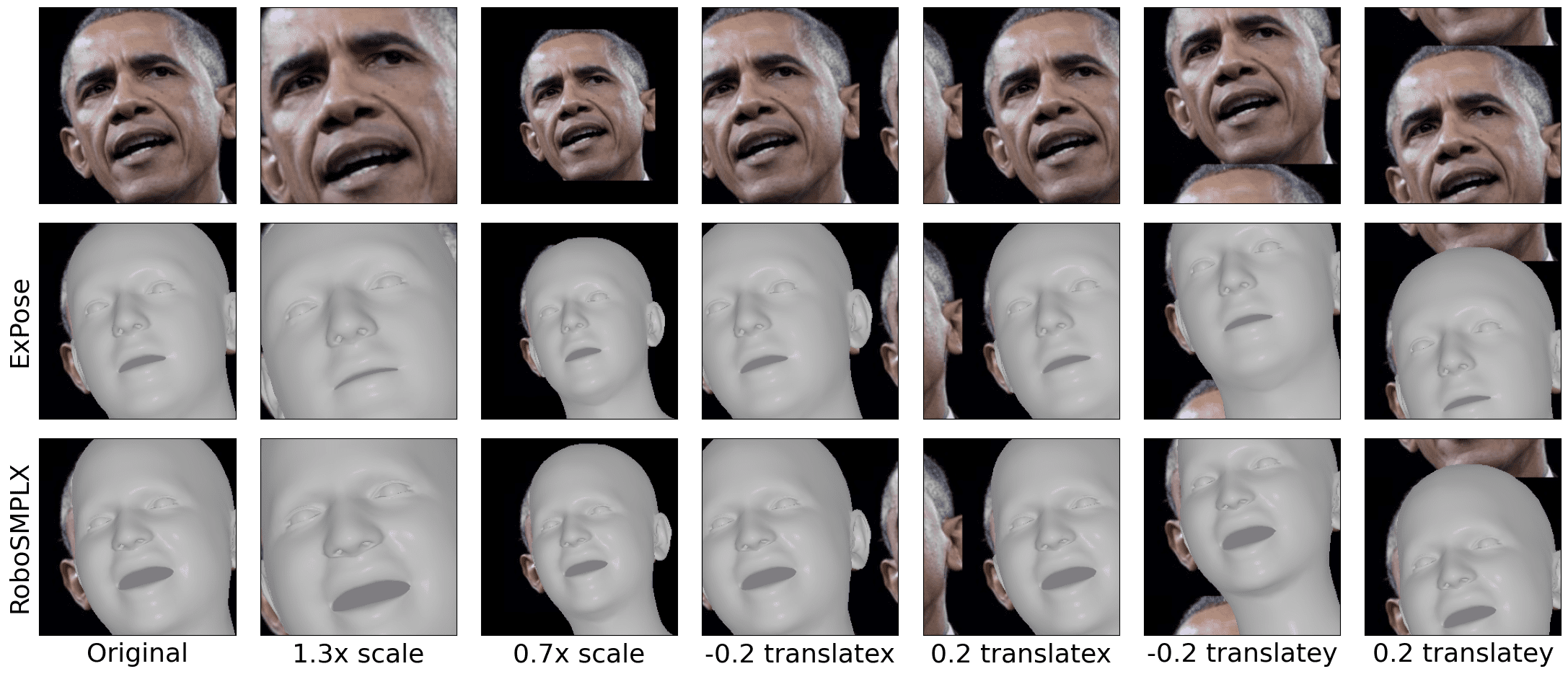}
    }
    \subfigure{
    \includegraphics[width=0.9\linewidth ,keepaspectratio]{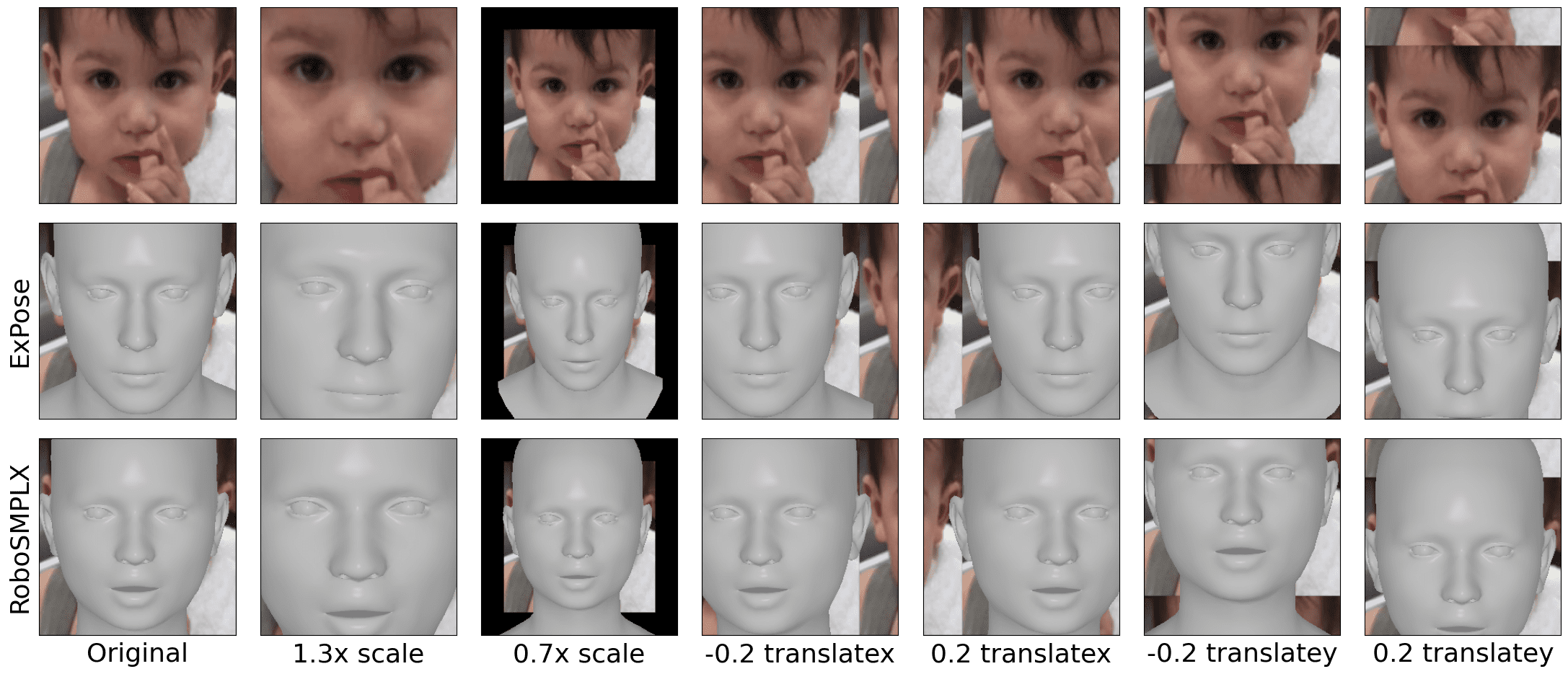}
    }
    \caption{\small \textbf{Comparison of ExPose \cite{Choutas2020} and \Name's Face subnetwork under various augmentations on AffectNet val set.}}
    \label{figure:face_aug_vis}
\end{figure}

We show qualitative comparisons of \Name's Hand (Figure \ref{figure:hand_aug_vis}), Face (Figure \ref{figure:face_aug_vis}) and Body (Figure \ref{figure:body_aug_vis}) subnetwork to existing models under different positional augmentations. 

In general, \Name s' subnetworks demonstrate better pixel alignment and are less sensitive to changes in scale and alignment.


\begin{figure}[H]
    \subfigure{
    \includegraphics[width=0.9\linewidth ,keepaspectratio]{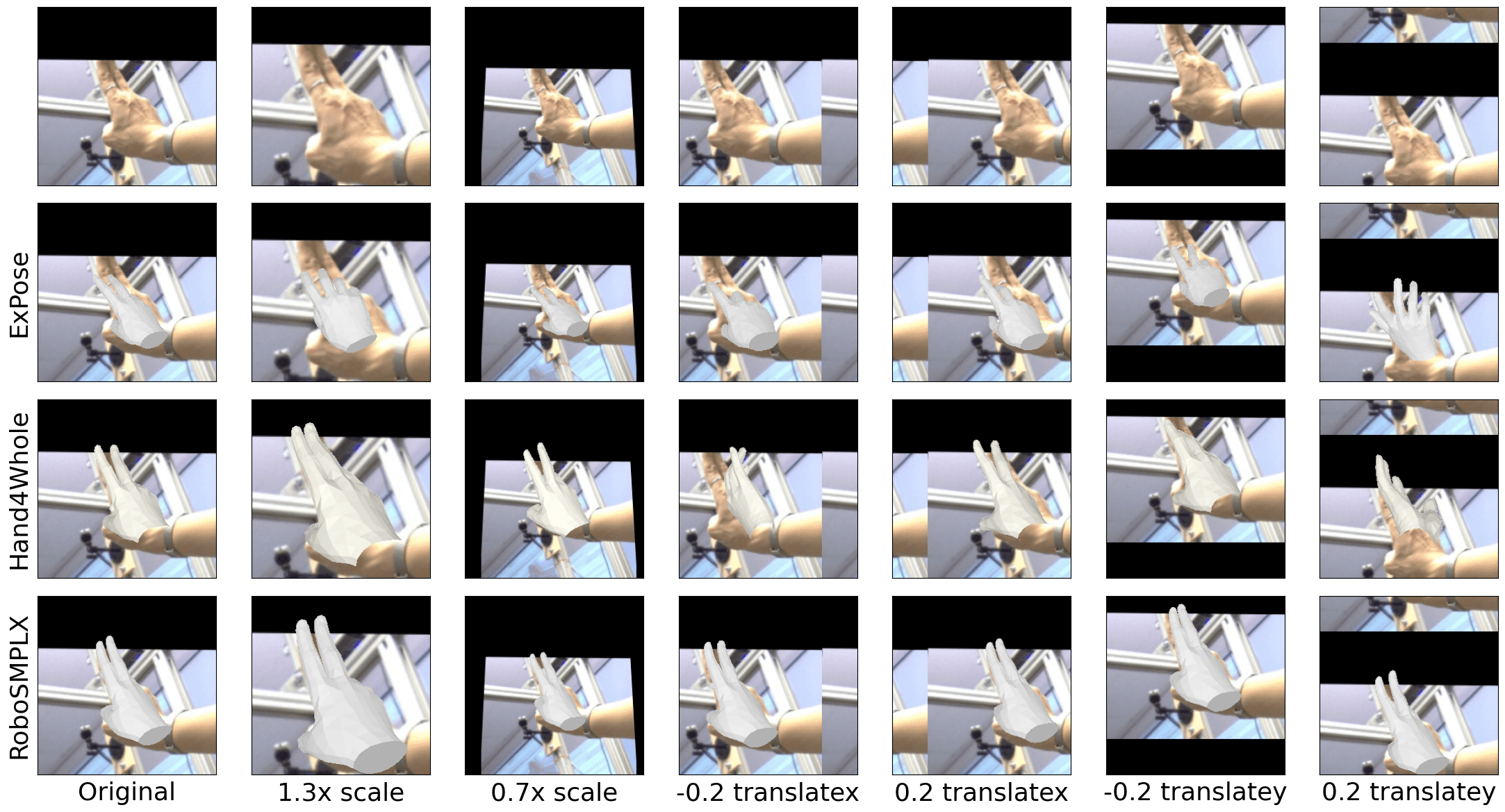}
    }
    \subfigure{
    \includegraphics[width=0.9\linewidth ,keepaspectratio]{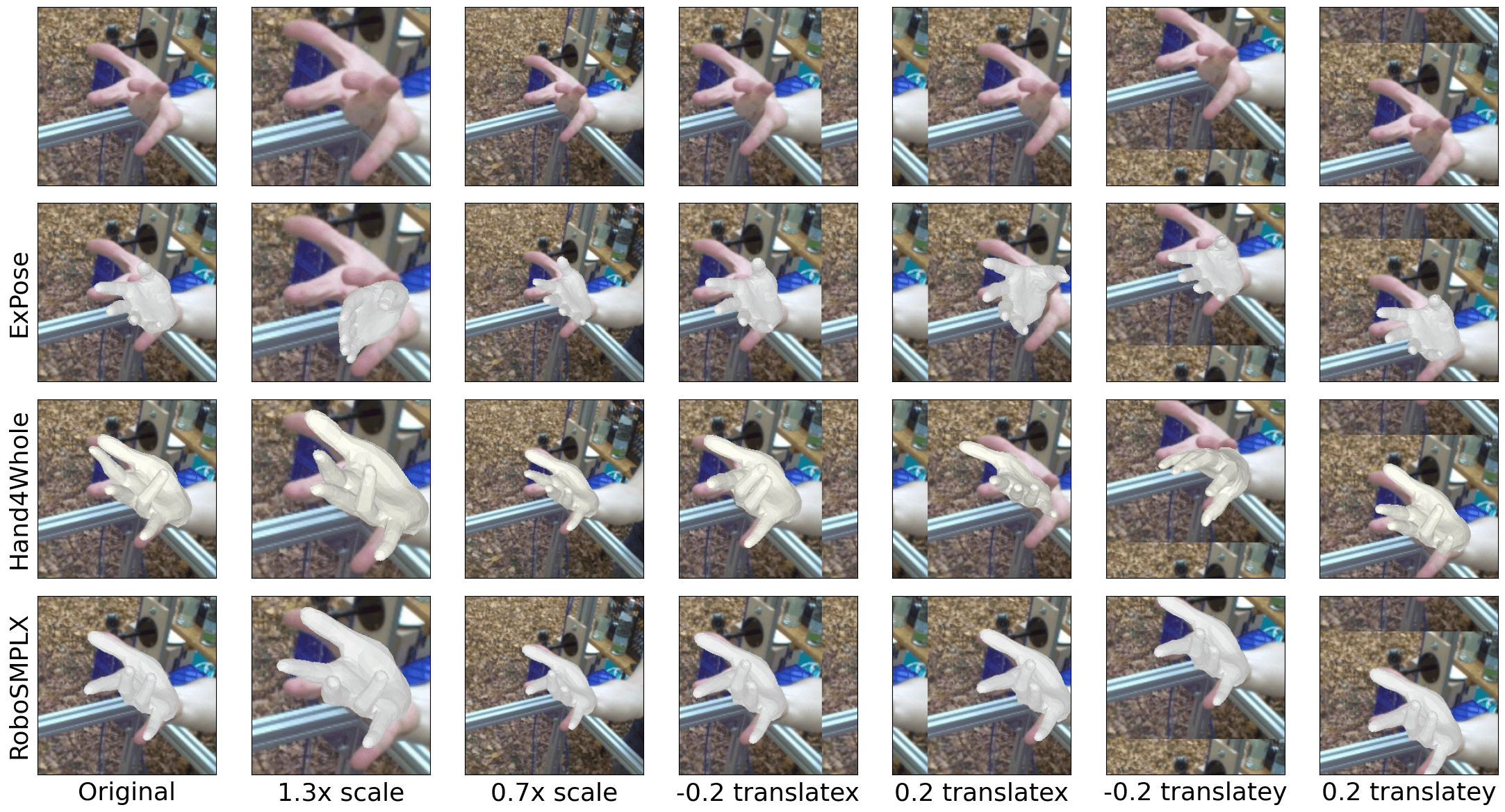}
    }
    \caption{\small \textbf{Comparison of ExPose \cite{Choutas2020}, Hand4Whole \cite{Moon2022} and \Name's Hand subnetwork under various augmentations on FreiHAND test set.}}
\end{figure}%
\begin{figure}[H]\ContinuedFloat
    \subfigure{
    \includegraphics[width=0.9\linewidth ,keepaspectratio]{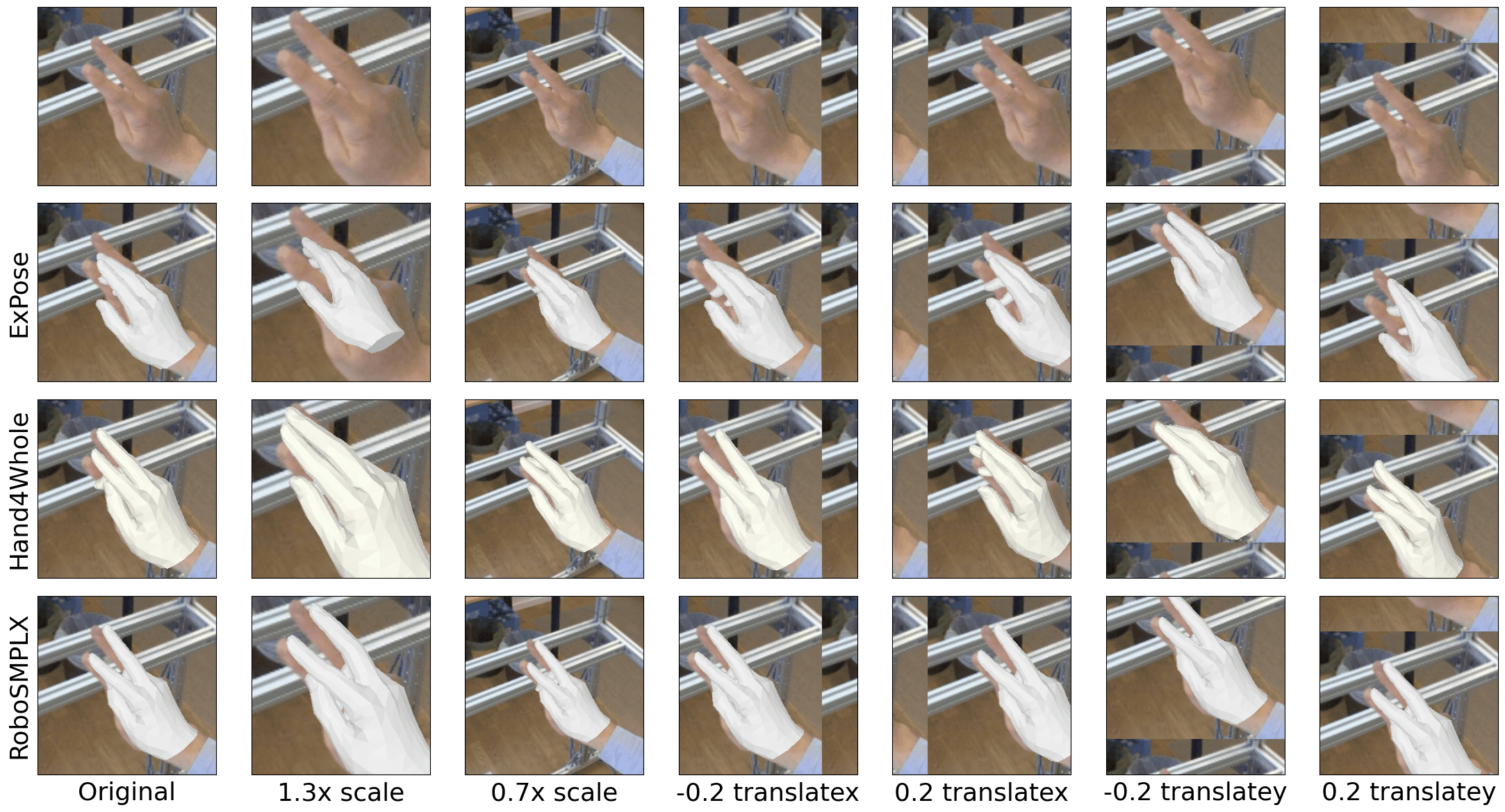}
    }
    \subfigure{
    \includegraphics[width=0.9\linewidth ,keepaspectratio]{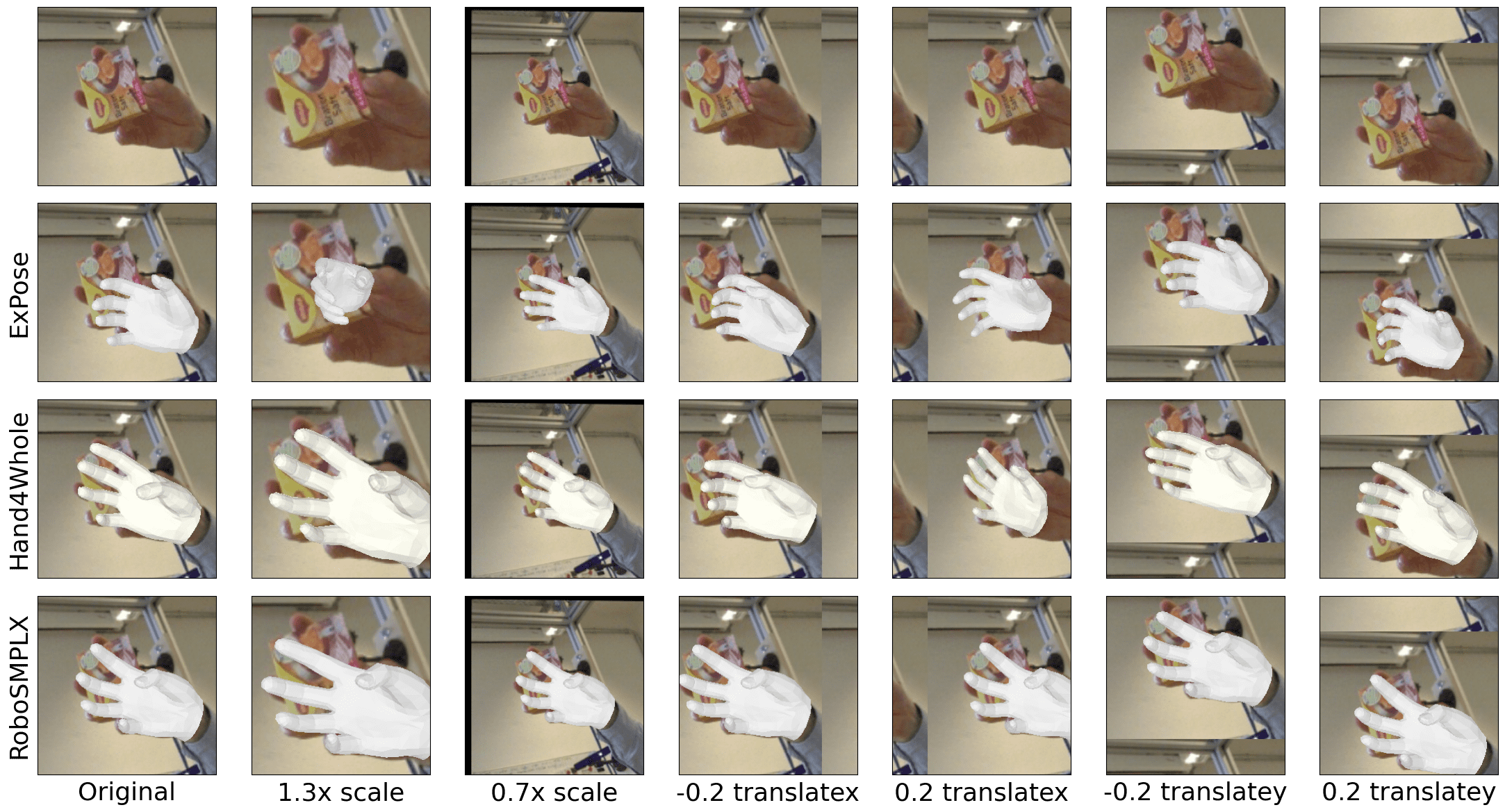}
    }
    \subfigure{
    \includegraphics[width=0.9\linewidth ,keepaspectratio]{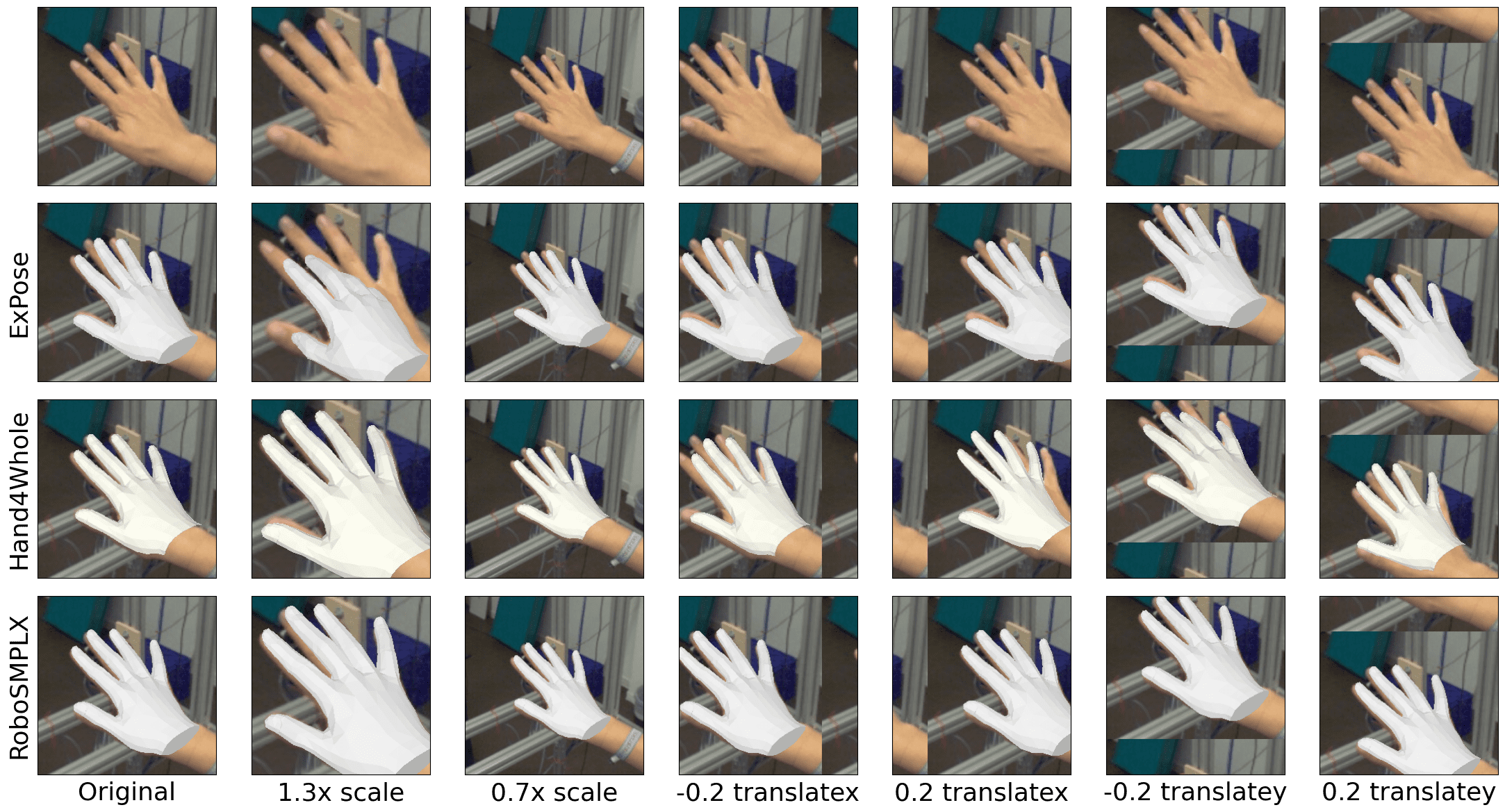}
    }
    \caption{\small \textbf{Comparison of of ExPose \cite{Choutas2020}, Hand4Whole \cite{Moon2022} and \Name's Hand subnetwork under various augmentations on FreiHAND test set (cont.)}}
    \label{figure:hand_aug_vis}
\end{figure}

\addtocounter{figure}{-1}
\begin{figure}[H]
    \subfigure{
    \includegraphics[width=0.97\linewidth ,keepaspectratio]{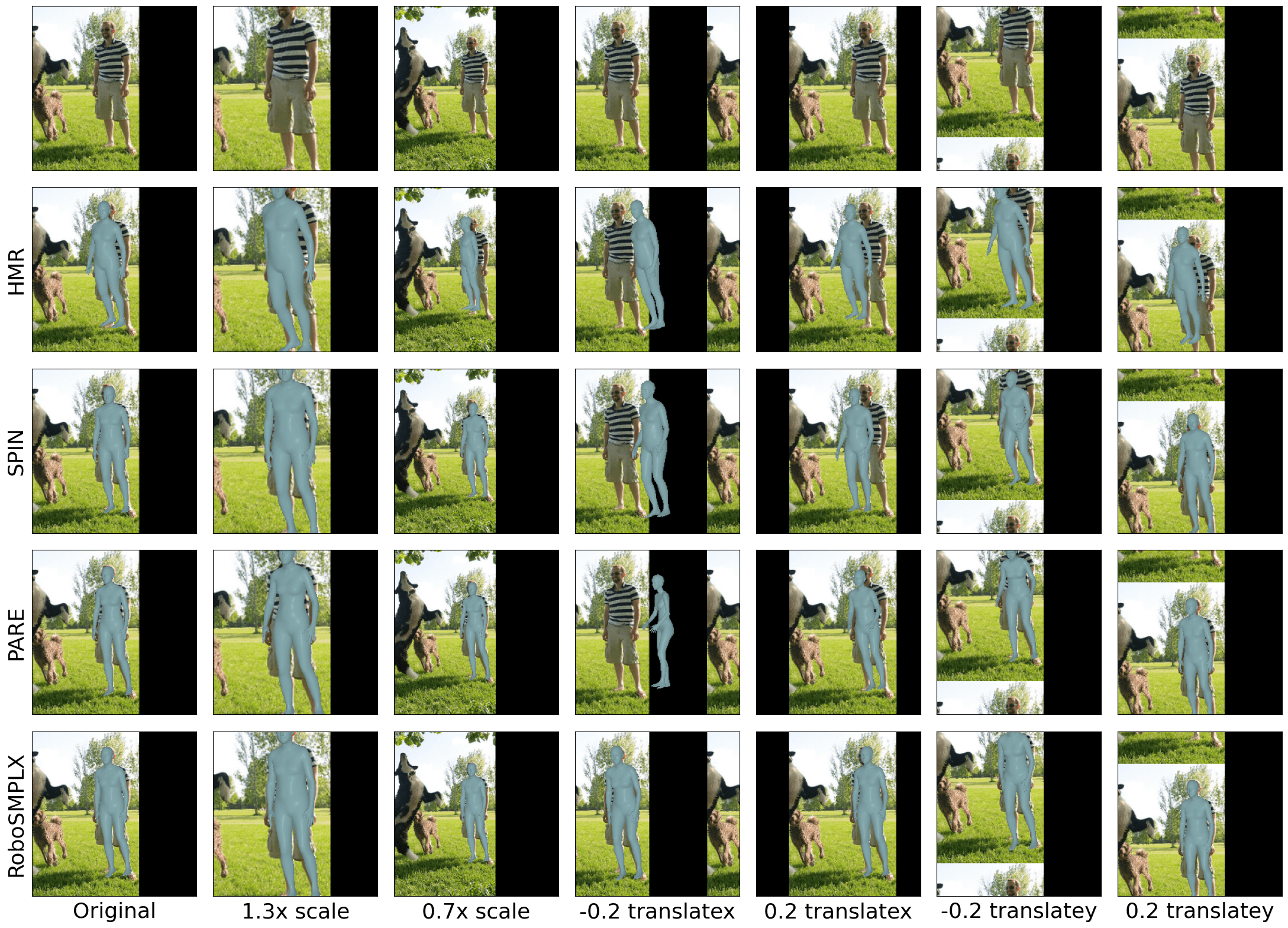}
    }
    \subfigure{
    \includegraphics[width=0.97\linewidth ,keepaspectratio]{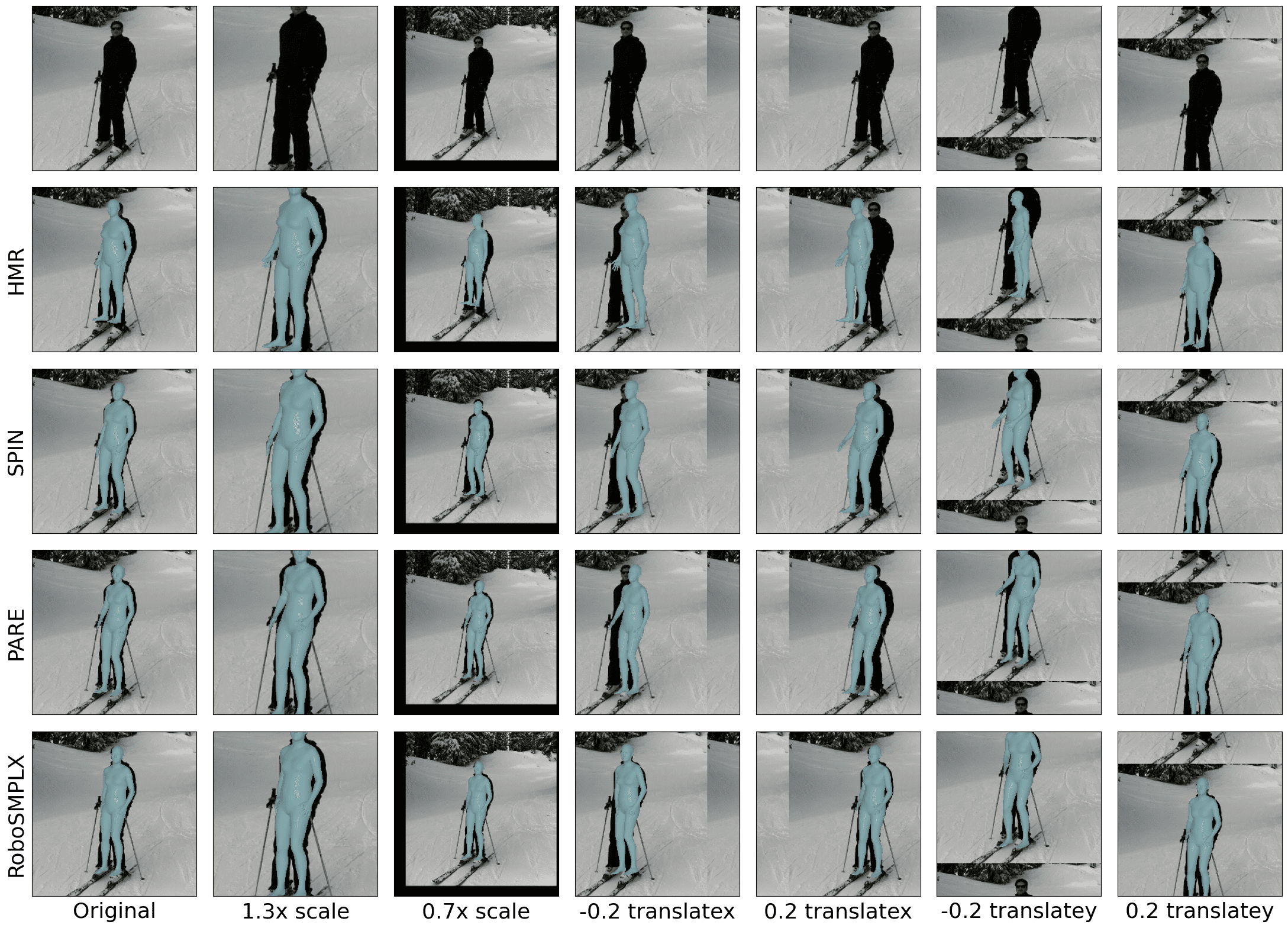}
    }

    \caption{\small \textbf{Comparison of HMR \cite{Kanazawa2017}, SPIN \cite{Kolotouros2019}, PARE\cite{Kocabas2021} and \Name's Body subnetwork under various augmentations on COCO validation set.}}
    \label{figure:body_aug_vis}
\end{figure}%
\begin{figure}[H]\ContinuedFloat
    \subfigure{
    \includegraphics[width=0.97\linewidth ,keepaspectratio]{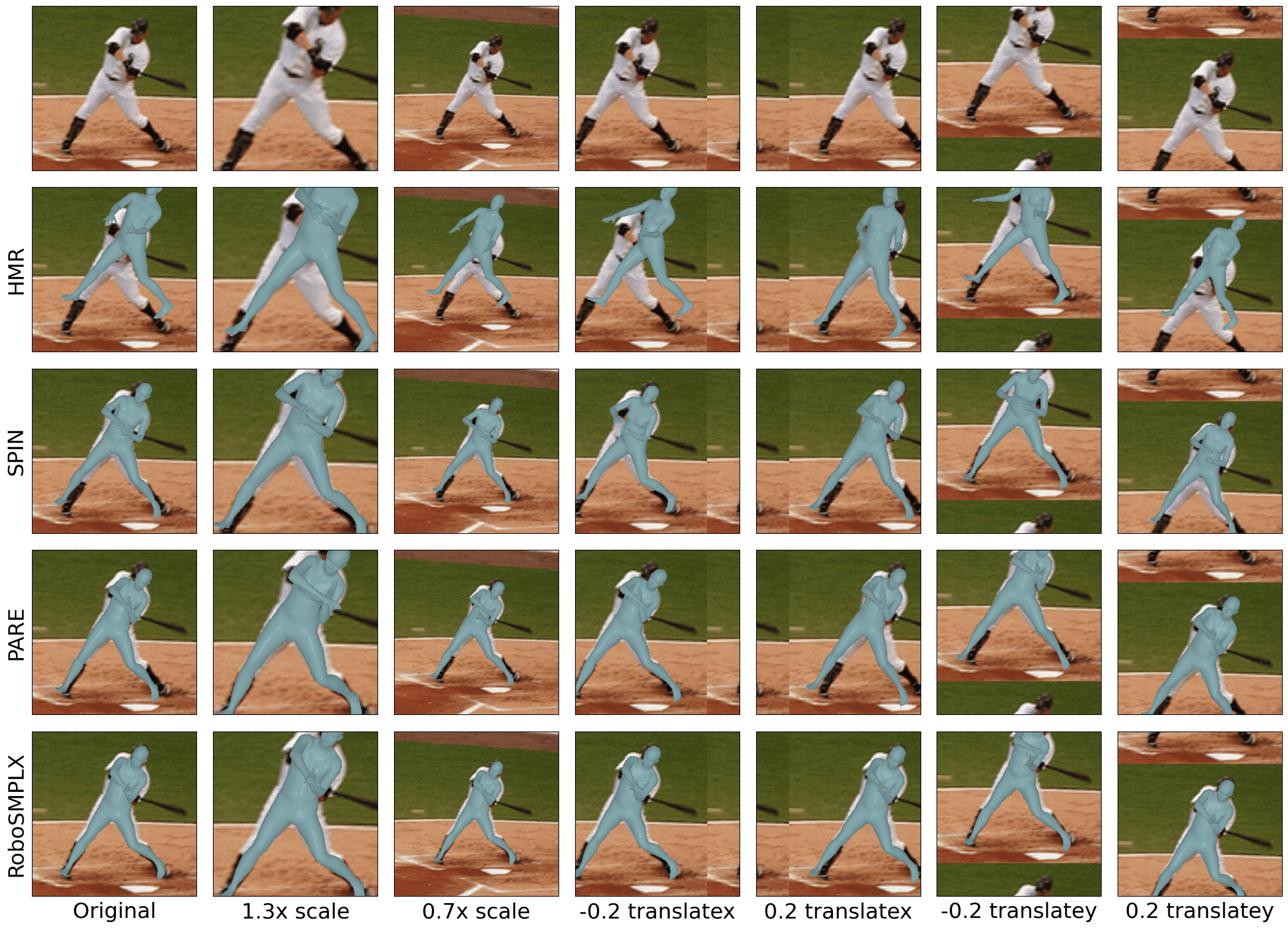}
    }
    \subfigure{
    \includegraphics[width=0.97\linewidth ,keepaspectratio]{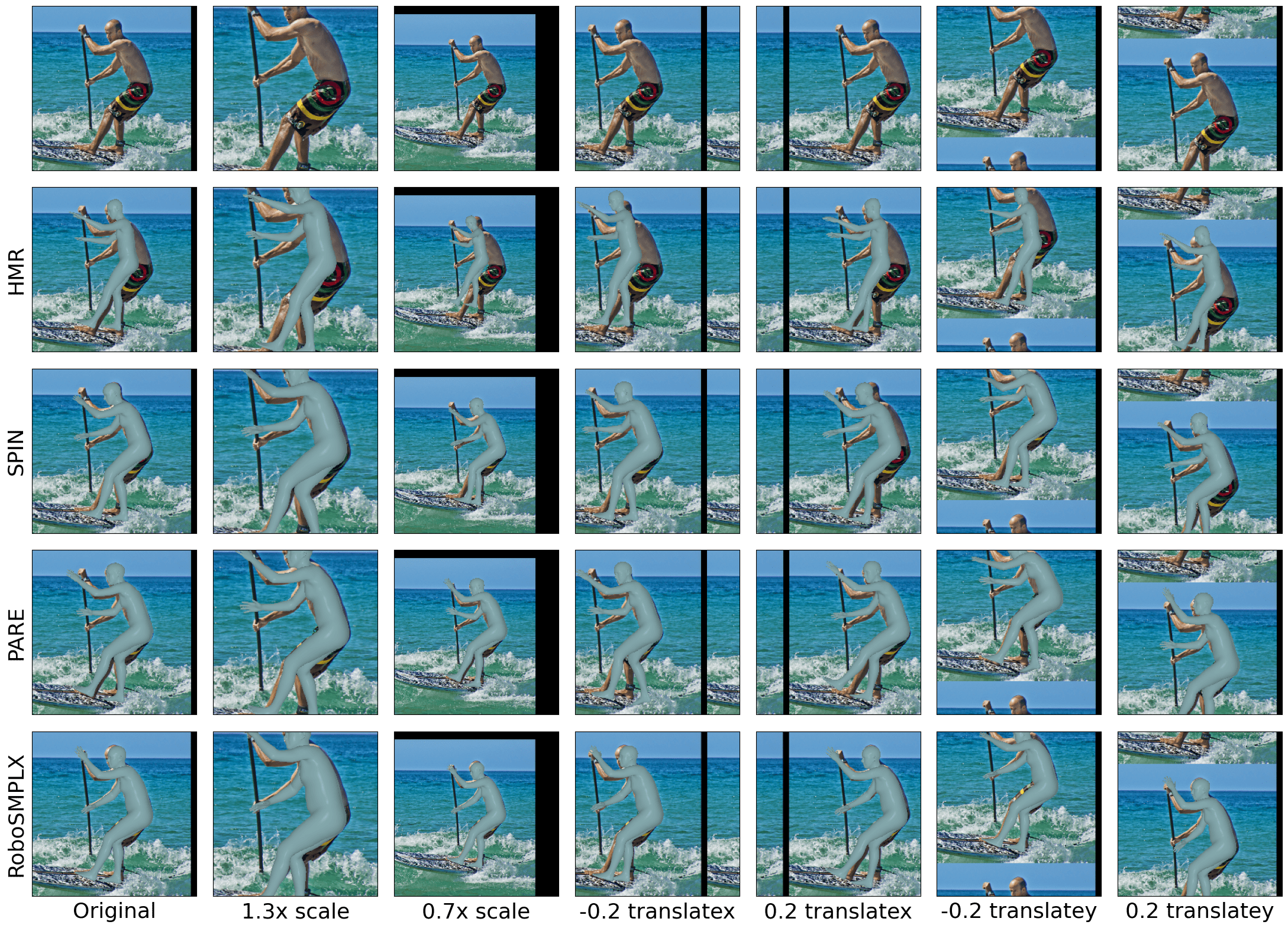}
    }
    \caption{\small \textbf{Comparison of HMR \cite{Kanazawa2017}, SPIN \cite{Kolotouros2019}, PARE\cite{Kocabas2021} and \Name's Body subnetwork under various augmentations on COCO validation set (cont.)}}
\end{figure}%

\section{Quantitative and qualitative comparisons for wholebody models}  
\label{sec:evaluation_wholebody}

\begin{figure}[H]
    \subfigure{
    \includegraphics[width=0.98\linewidth ,keepaspectratio]{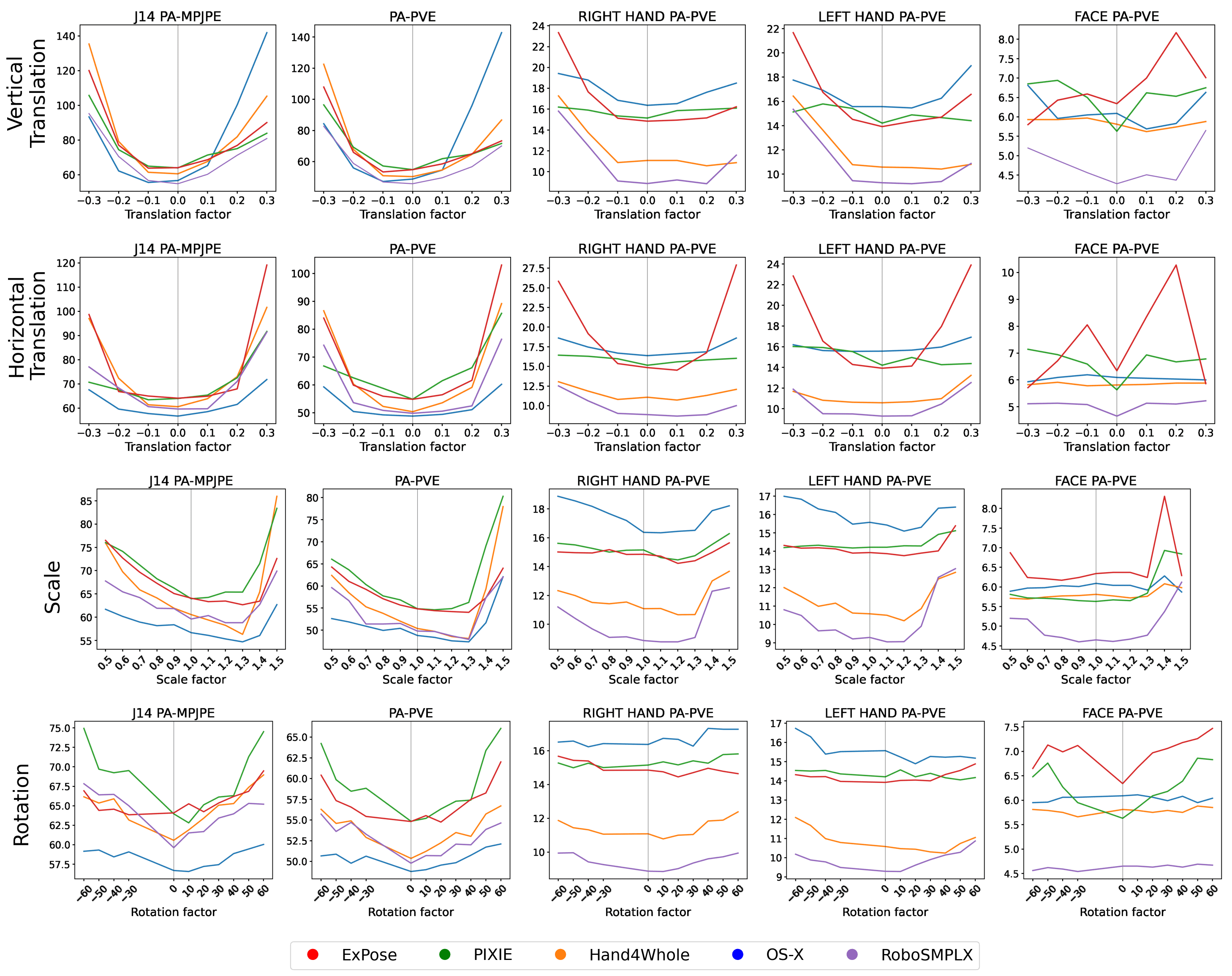}
    }
    \caption{\small \textbf{Wholebody errors under different amounts of augmentation on EHF test set. The gray line indicates baseline performance without augmentation.}}
    \label{figure:wholebody_errors_compile1}
\end{figure}

\begin{figure}[H]
    \subfigure{
    \includegraphics[width=0.98\linewidth ,keepaspectratio]{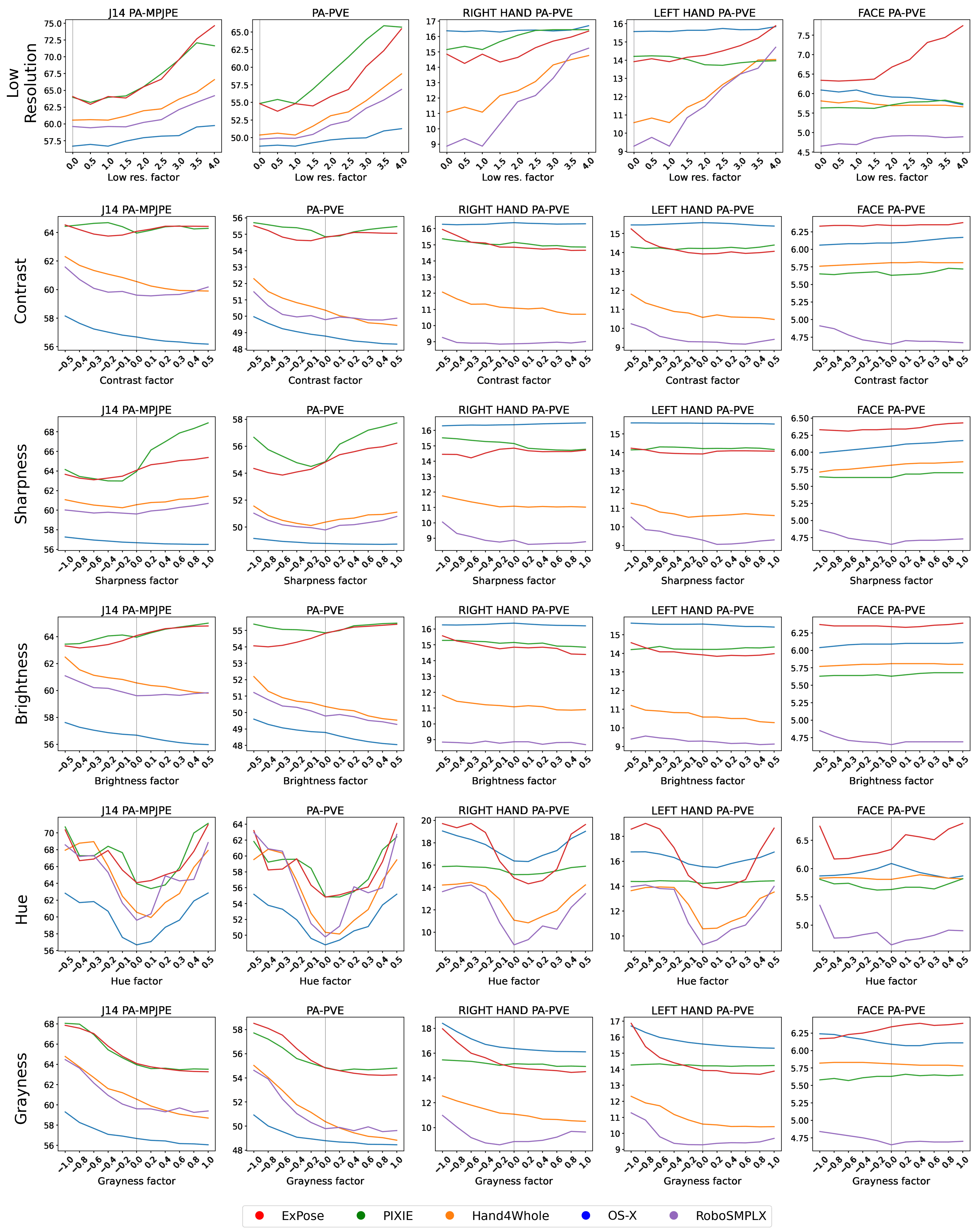}
    }
    \caption{\small \textbf{Wholebody errors under different amounts of augmentation on EHF test set (cont.) The gray line indicates baseline performance without augmentation.}}
    \label{figure:wholebody_errors_compile2}
\end{figure}

\begin{figure}[H]
    \centering
    \subfigure{
    \includegraphics[width=0.98\linewidth,keepaspectratio]{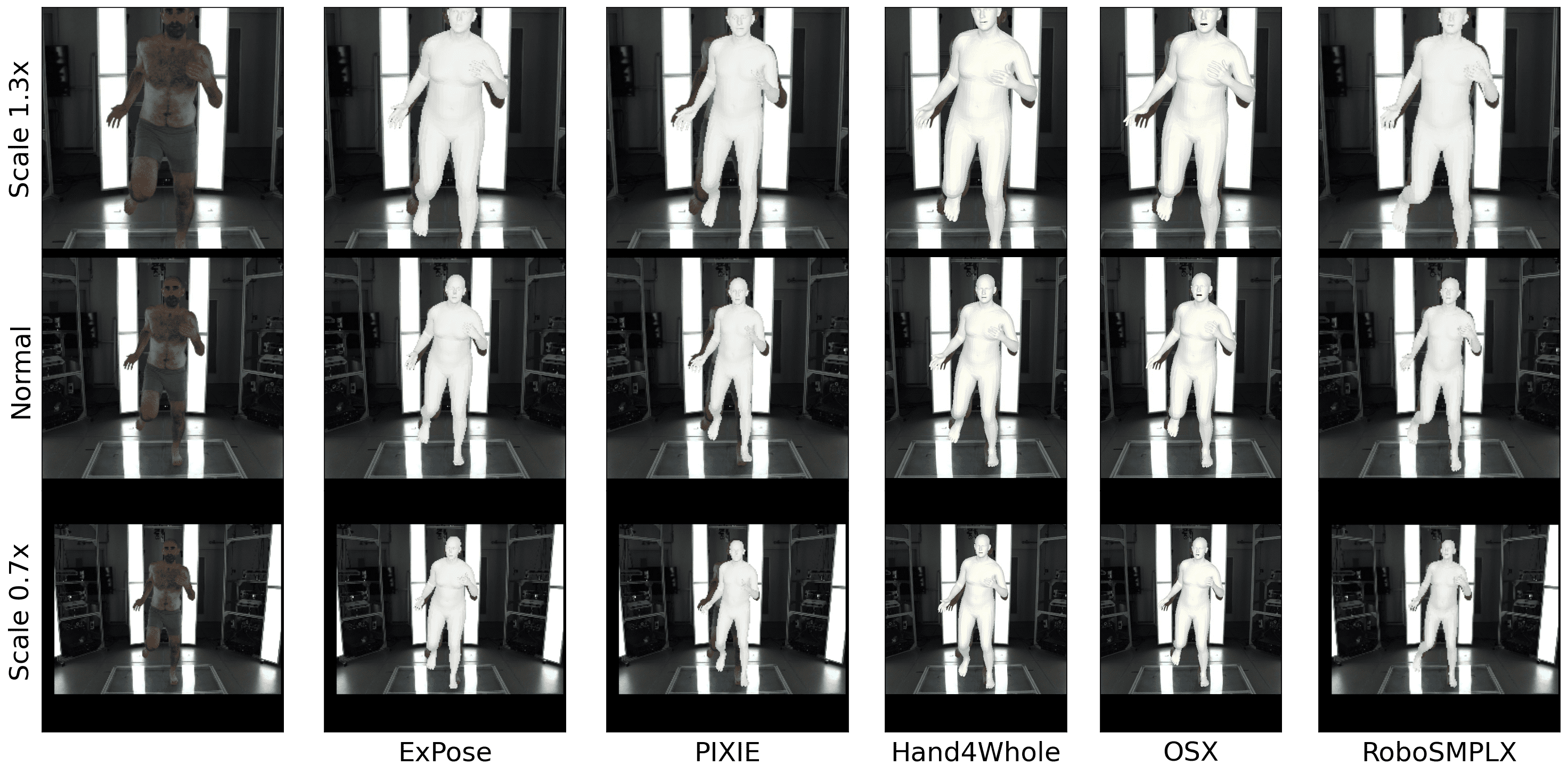}
    }
    \subfigure{
    \includegraphics[width=0.98\linewidth,keepaspectratio]{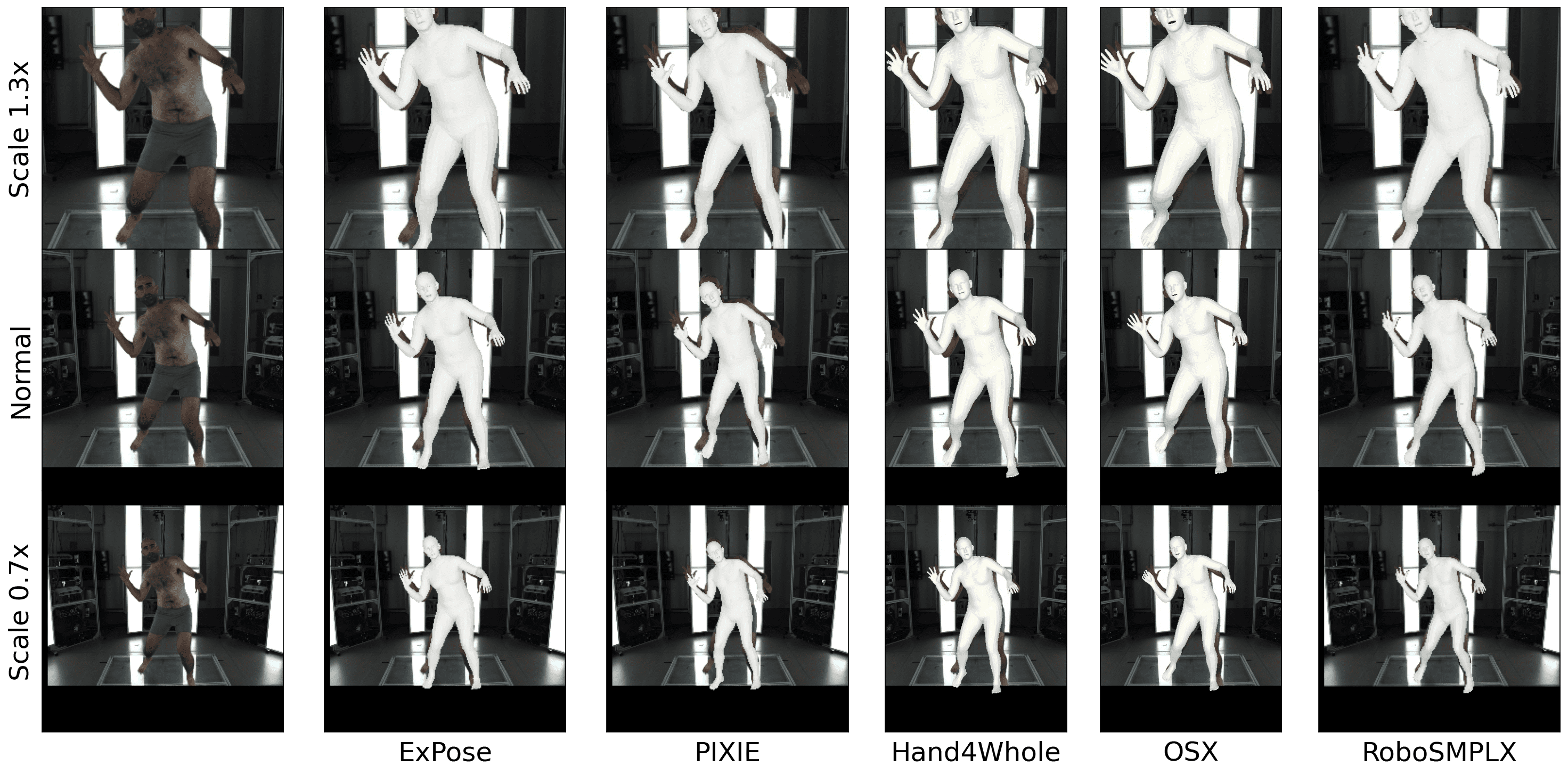}
    }
    \caption{\small \textbf{Visualisation of Expose \cite{Pavlakos2019}, PIXIE \cite{Feng2019}, Hand4Whole \cite{Moon2022}, OS-X \cite{Lin2023} and \Name under different scales on EHF test set.}}
    \label{figure:wb_vis_scale}
\end{figure}

\begin{figure}[H]
    \centering
    \subfigure{
    \includegraphics[width=0.98\linewidth,keepaspectratio]{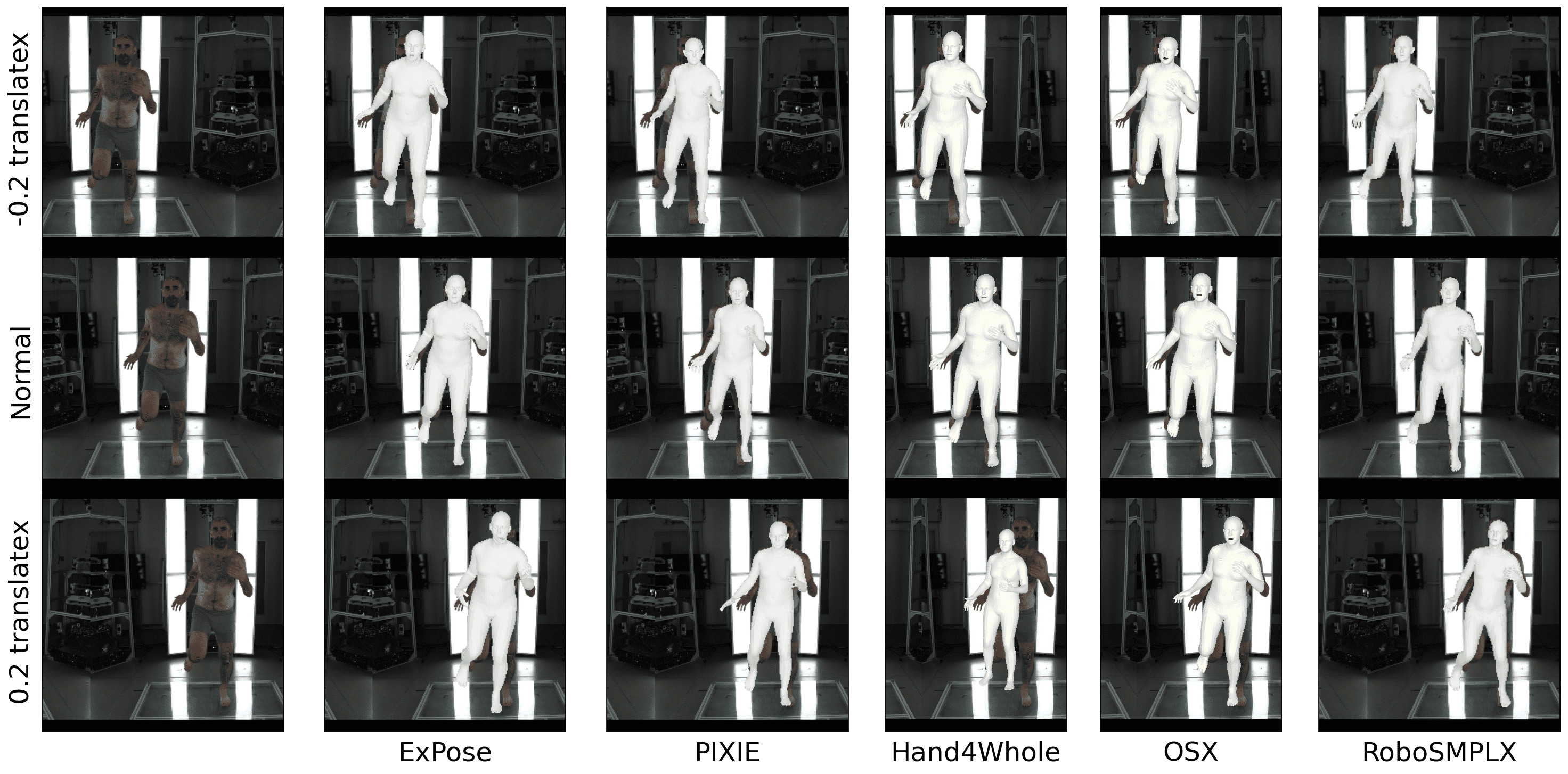}
    }
    \subfigure{
    \includegraphics[width=0.98\linewidth,keepaspectratio]{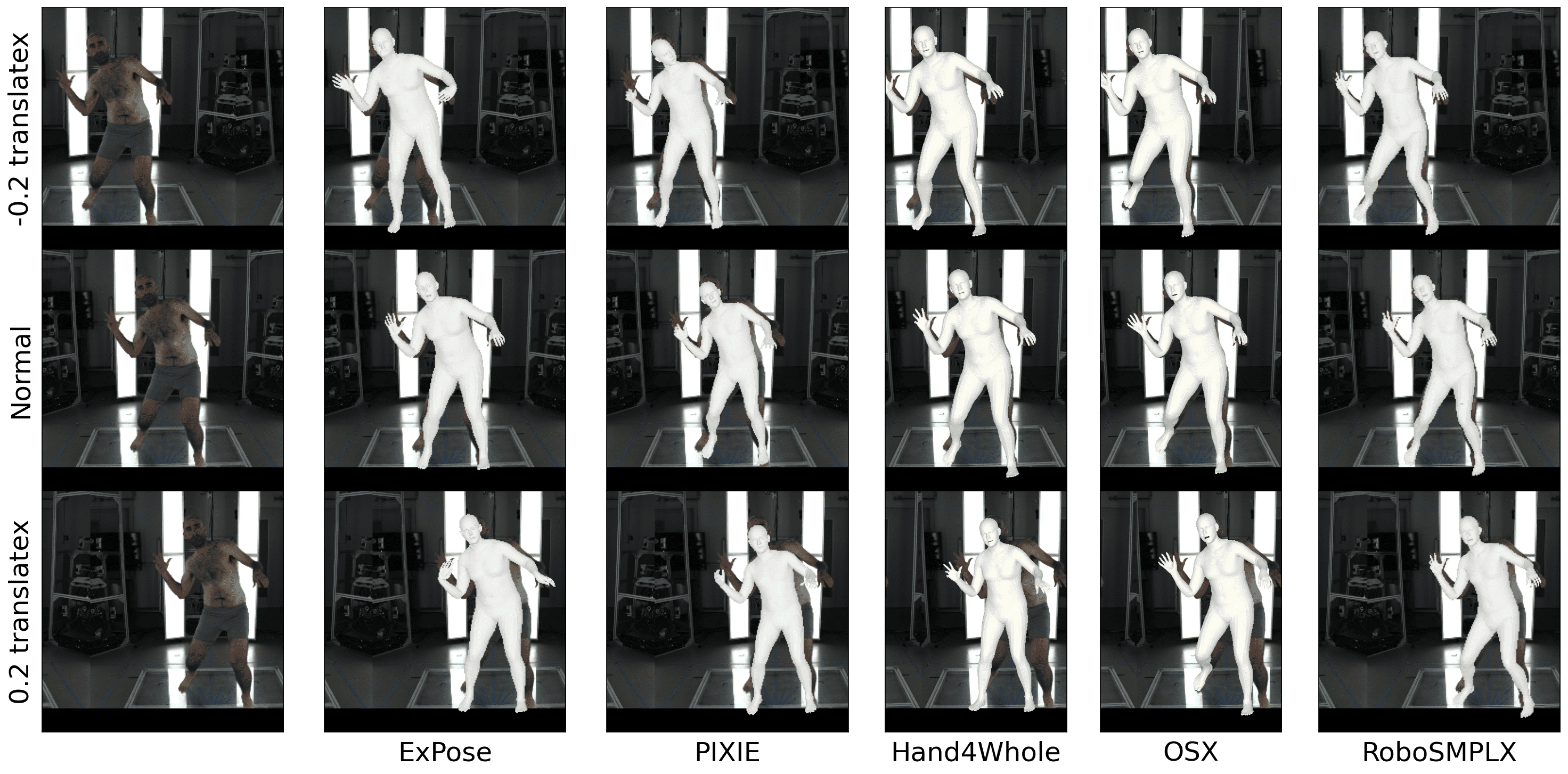}
    }
    \caption{\small \textbf{Visualisation of Expose \cite{Pavlakos2019}, PIXIE \cite{Feng2019}, Hand4Whole \cite{Moon2022}, OS-X \cite{Lin2023} and \Name under different levels of horizontal translation on EHF test set.}}
    \label{figure:wb_vis_translatex}
\end{figure}

\begin{figure}[H]
    \centering
    \subfigure{
    \includegraphics[width=0.98\linewidth,keepaspectratio]{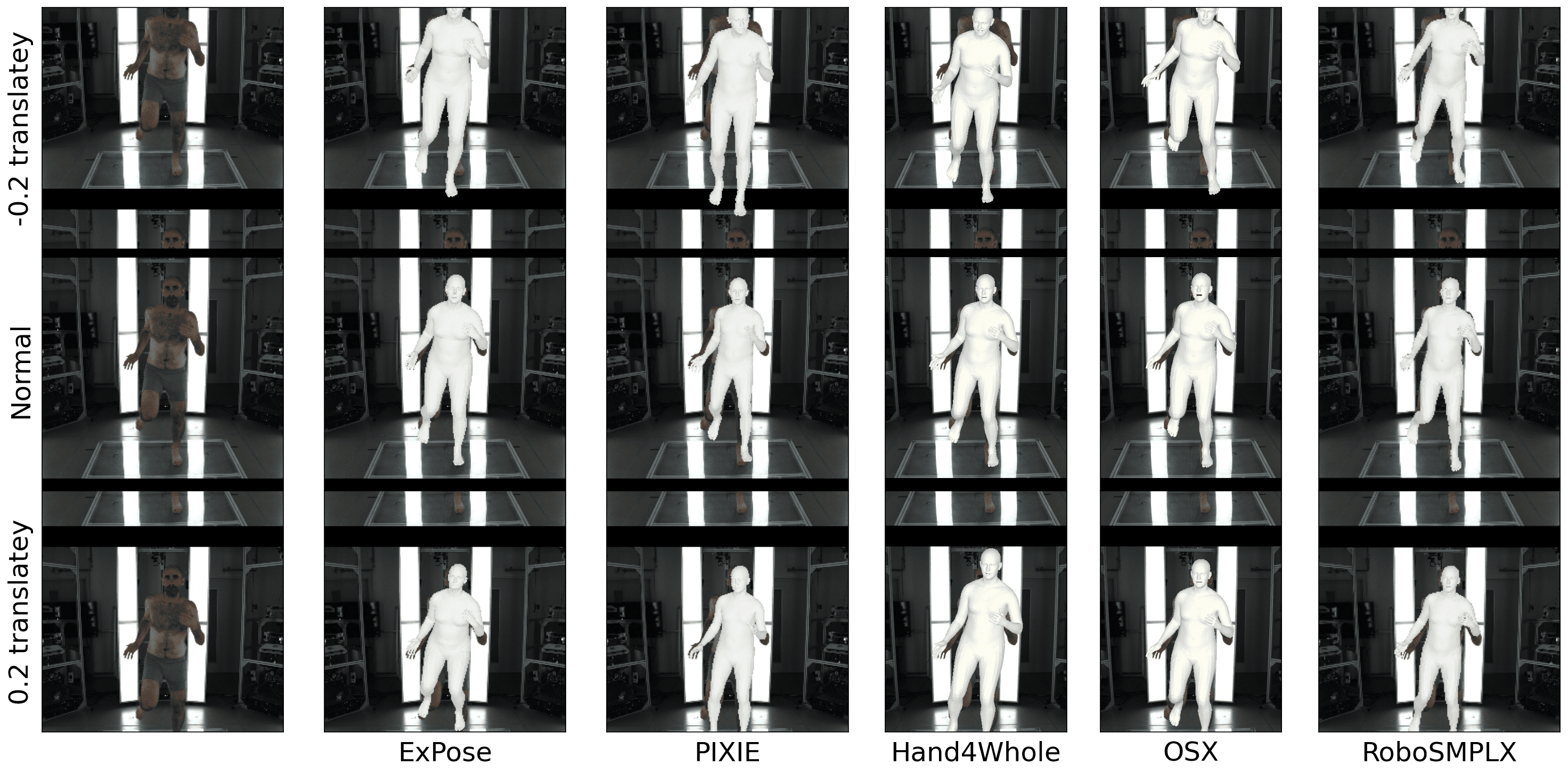}
    }
    \subfigure{
    \includegraphics[width=0.98\linewidth,keepaspectratio]{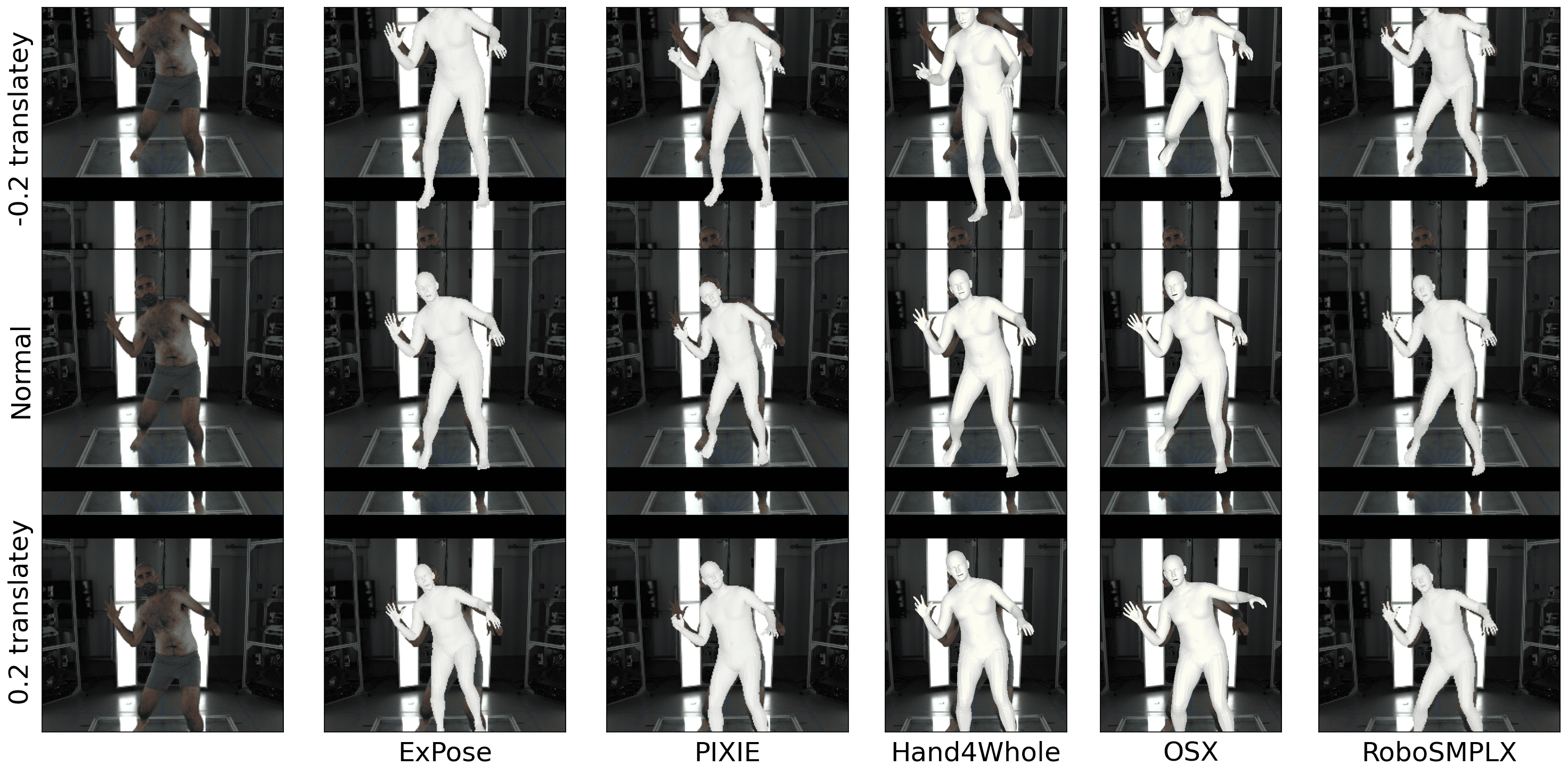}
    }
    \caption{\small \textbf{Visualisation of Expose \cite{Pavlakos2019}, PIXIE \cite{Feng2019}, Hand4Whole \cite{Moon2022}, OS-X \cite{Lin2023} and \Name under different levels of vertical translation on EHF test set.}}
    \label{figure:wb_vis_translatey}
\end{figure}

\begin{figure}[H]
    \centering
    \subfigure{
    \includegraphics[width=0.9\linewidth,keepaspectratio]{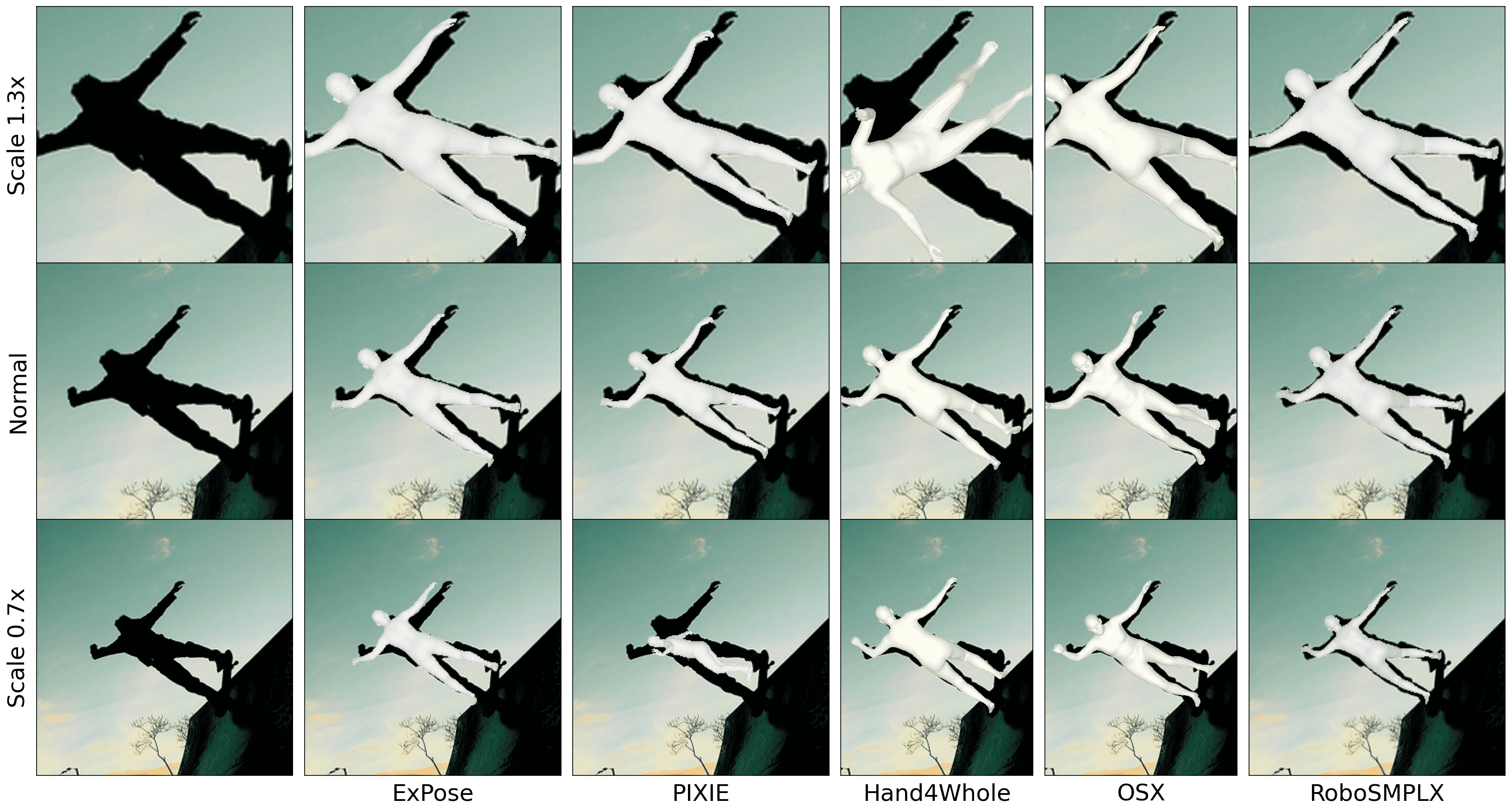}
    }
    \subfigure{
    \includegraphics[width=0.9\linewidth,keepaspectratio]{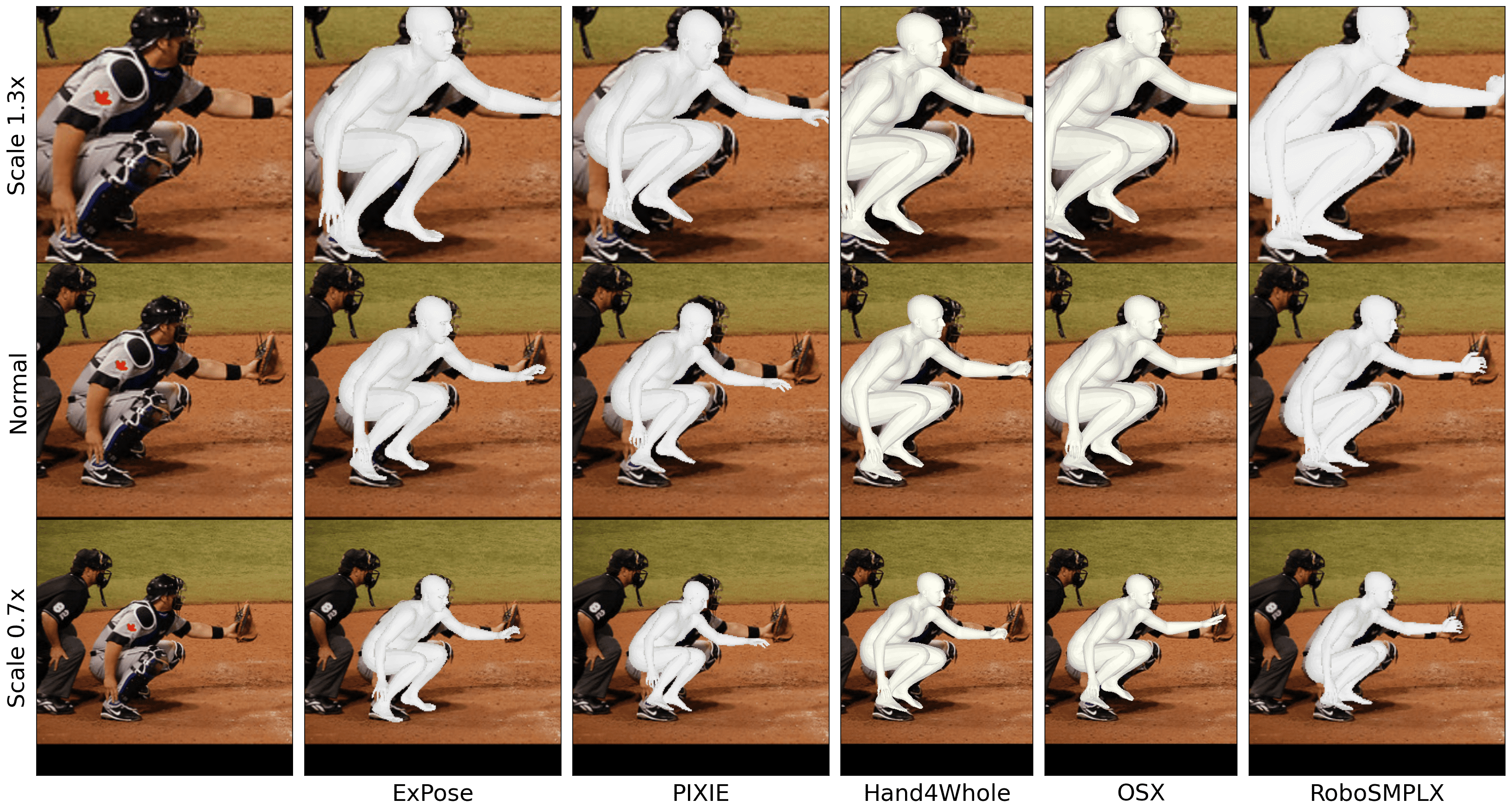}
    }
    \subfigure{
    \includegraphics[width=0.9\linewidth,keepaspectratio]{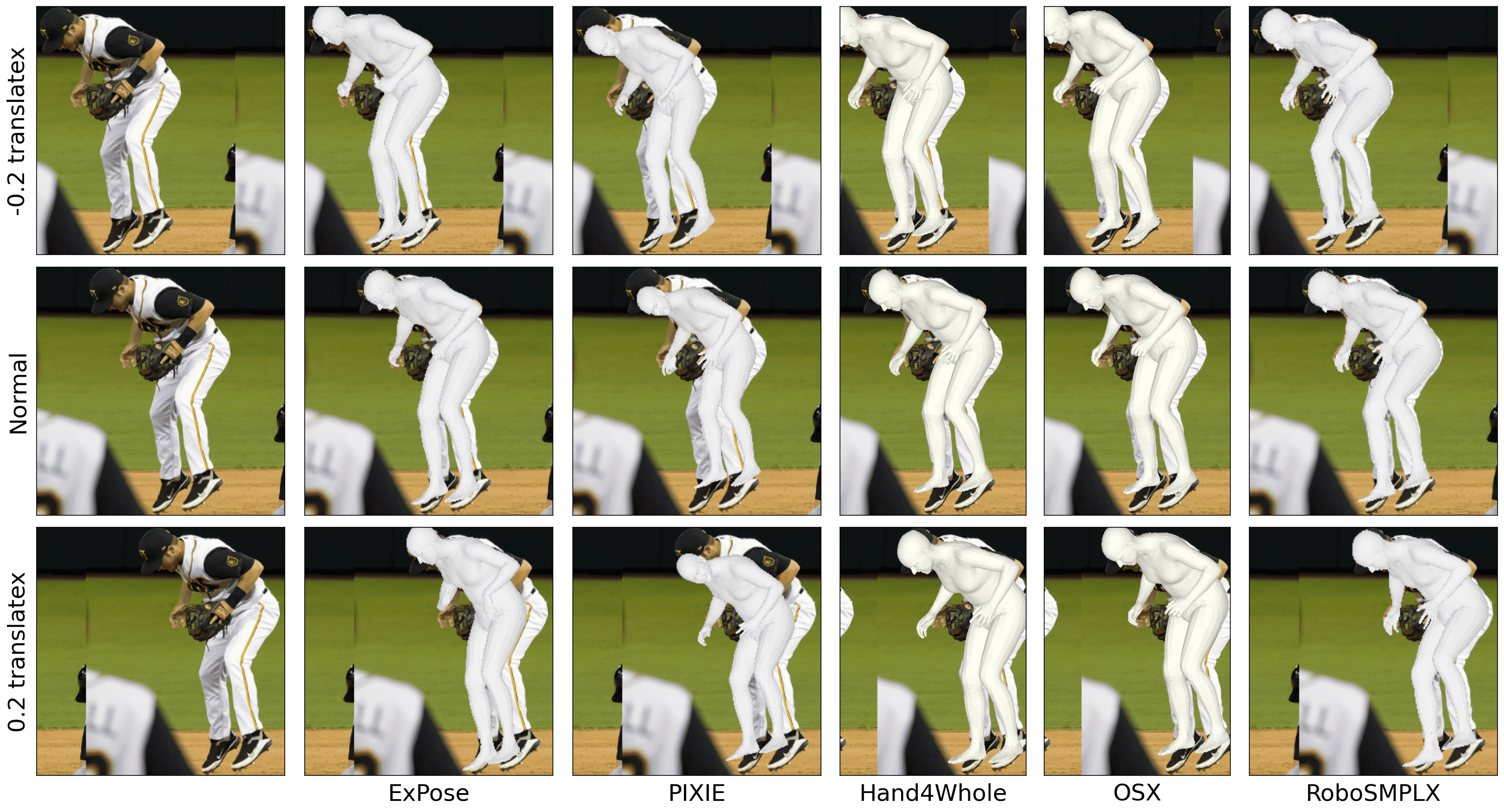}
    }
    \caption{\small \textbf{Visualisation of Expose \cite{Pavlakos2019}, PIXIE \cite{Feng2019}, Hand4Whole \cite{Moon2022}, OS-X \cite{Lin2023} and \Name under different scales and alignment on COCO validation set.}}
    \label{figure:wholebody_itw_aug_vis1}
\end{figure}

\begin{figure}[H]
    \subfigure{
    \includegraphics[width=0.9\linewidth ,keepaspectratio]{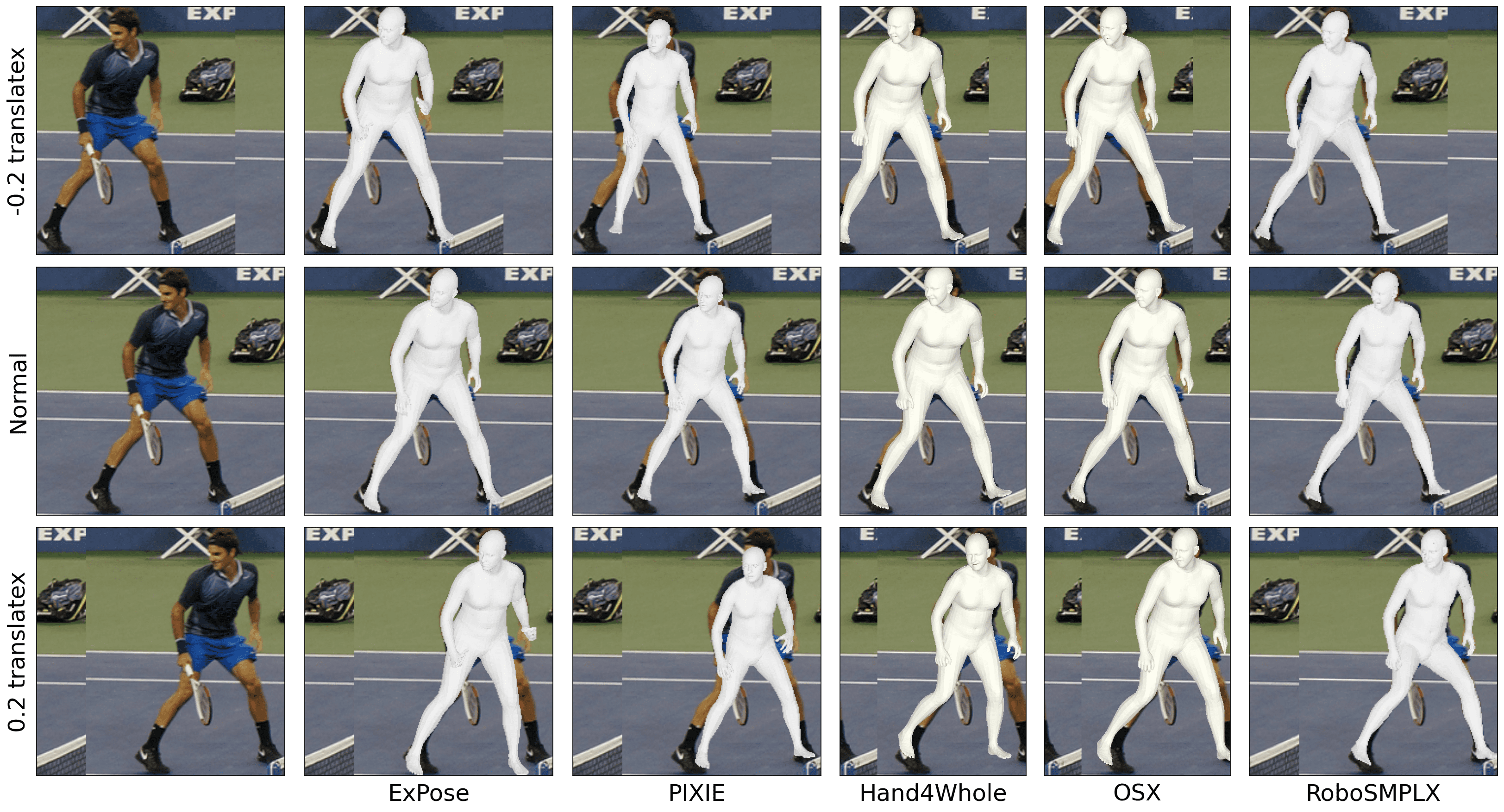}
    }
    \subfigure{
    \includegraphics[width=0.9\linewidth ,keepaspectratio]{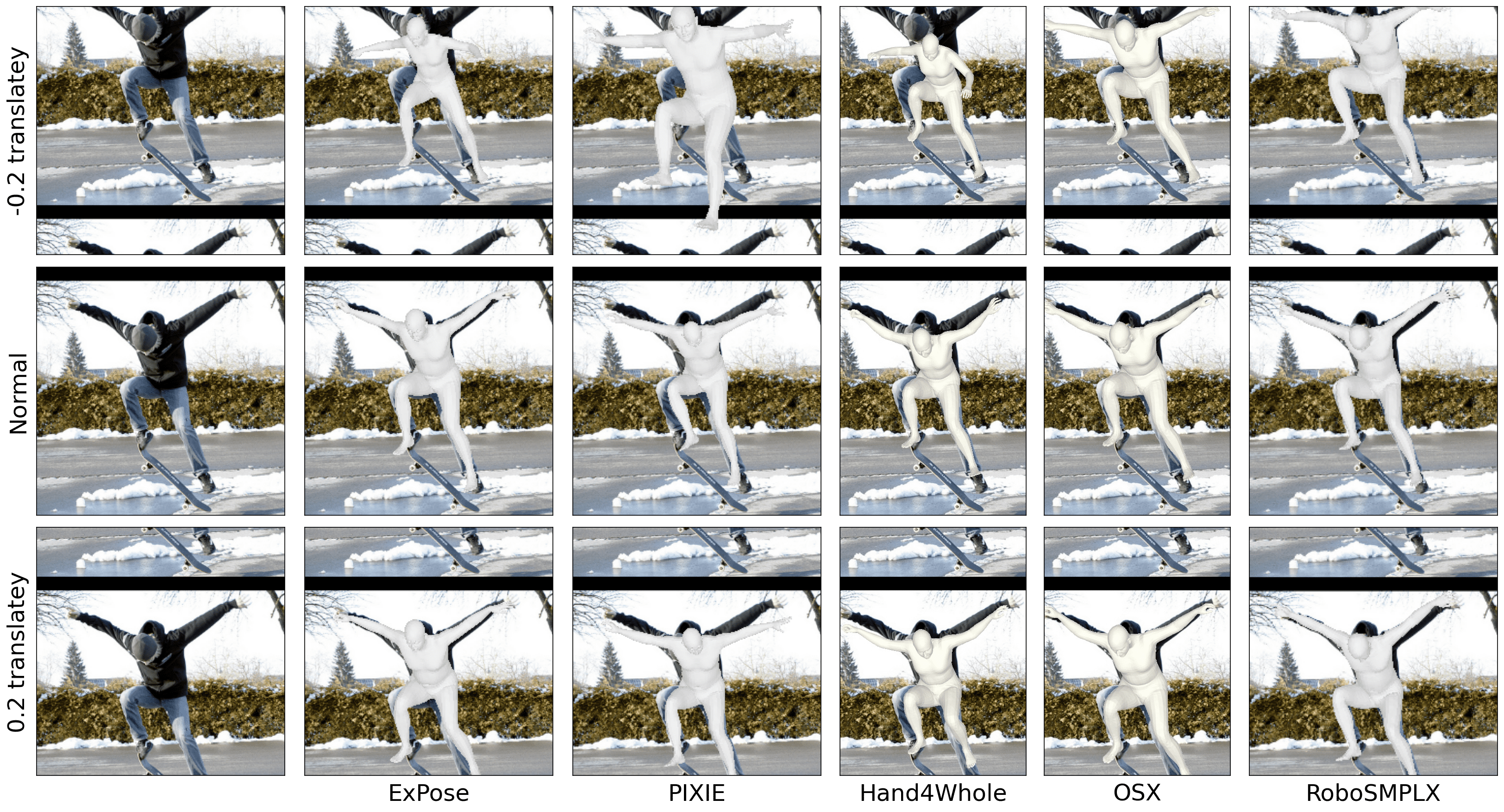}
    }
    \subfigure{
    \includegraphics[width=0.9\linewidth ,keepaspectratio]{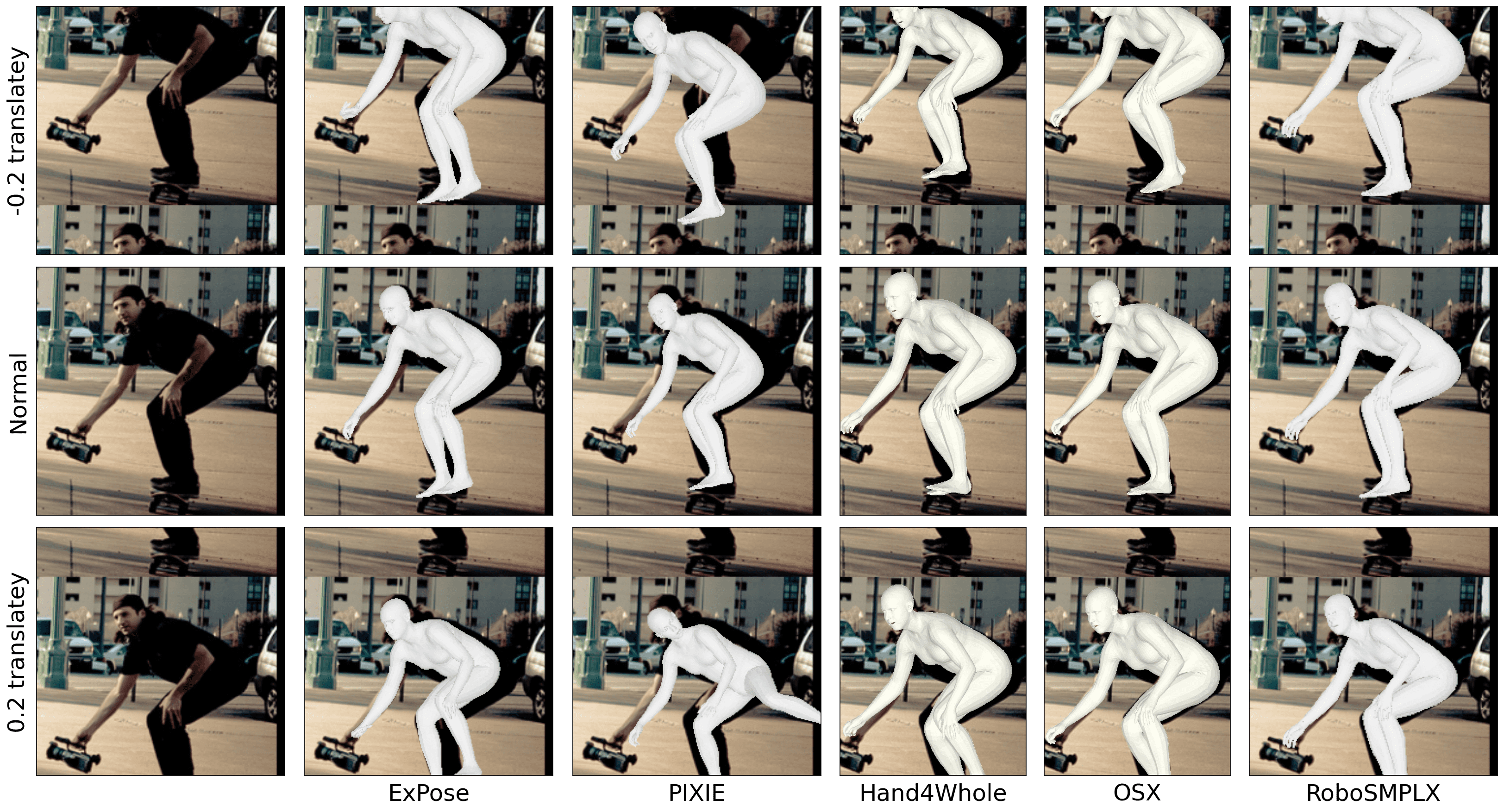}
    }
    \caption{\small \textbf{Visualisation of Expose \cite{Pavlakos2019}, PIXIE \cite{Feng2019}, Hand4Whole \cite{Moon2022}, OS-X \cite{Lin2023} and \Name under different scales and alignment on COCO validation set.}}
    \label{figure:wholebody_itw_aug_vis2}
\end{figure}

We provide quantitative comparisons of wholebody models under different augmentations on EHF test set in Figures \ref{figure:wholebody_errors_compile1} and \ref{figure:wholebody_errors_compile2}. We also added qualitative comparisons under different scale and alignment on EHF test set in Figures \ref{figure:wb_vis_scale} to \ref{figure:wb_vis_translatey}. We demonstrate that \Name produces better pixel alignment of the body, and more accurate hand and face predictions. In addition, we inference on in-the-wild examples on COCO-validation set in Figures \ref{figure:wholebody_itw_aug_vis1} and \ref{figure:wholebody_itw_aug_vis2}.

\begin{figure*}[!ht]
    \centering
    \subfigure{
    \includegraphics[width=0.9\linewidth ,keepaspectratio]{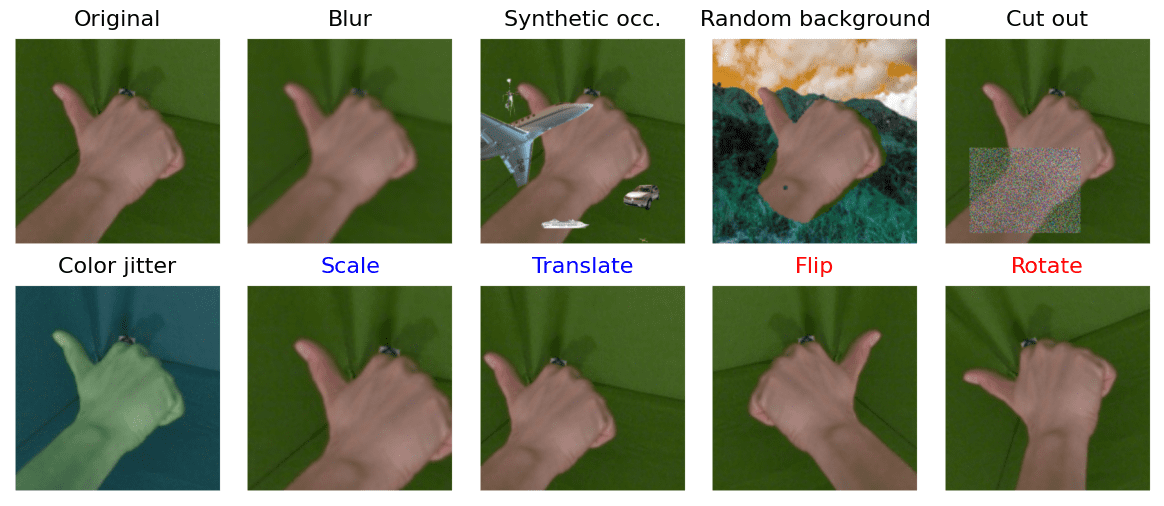}
    }
    \caption{\small \textbf{Augmentations for Hand sub-networks. Blue and red labels represent location-variant and pose-variant augmentations respectively.}}
    \label{figure:simclr_hand_aug}
\end{figure*}




\end{document}